\documentclass[11pt]{article}
\usepackage{xcolor}

\usepackage[T1]{fontenc}
\usepackage[utf8]{inputenc}
\usepackage[english]{babel}
\usepackage[margin=.89in]{geometry} 
\usepackage{comment}
\usepackage{hyperref}

\usepackage{amsmath, amssymb, mathabx, mathrsfs, mathtools, nccmath, amsthm}
\usepackage{hyperref}
\usepackage{graphicx}
\usepackage{color}
\usepackage{tcolorbox}

\usepackage{epsfig}
\usepackage{tikz, pgfplots}
\usepackage{subcaption}
\usepackage{float}

\usepackage{comment}
\usepackage[multiple]{footmisc}
\usepackage{bigfoot}
\usepackage{multicol}
\usepackage{ulem}
\usepackage{stackengine}
\usepackage{algorithm}
\usepackage{algpseudocode}
\usepackage{url}

\usepackage[toc,page]{appendix}

\theoremstyle{plain}
\newtheorem{thm}{Theorem}[section]

\newtheorem{lem}[thm]{Lemma}

\newtheorem{defn}[thm]{Definition}

\newtheorem{ex}[thm]{Example}
\newtheorem{aplem}[thm]{Approximation Lemma}

\theoremstyle{definition}
\theoremstyle{remark}
\newtheorem{remark}{Remark}[section]

\newcommand{\R}{\mathbb{R}}
\newcommand{\lzz}[1]{{\color{red} #1}} 

\title{Enhancing Accuracy in Deep Learning Using Random Matrix Theory}
\author{
    Leonid Berlyand \\
    {\small Department of Mathematics} \\
    {\small Pennsylvania State University} \\
    {\small University Park, PA 16802, USA} \\
    \and
    Etienne Sandier \\
    {\small LAMA-CNRS UMR 8050} \\
    {\small Université Paris-Est Créteil} \\
    {\small 61, Avenue du Général de Gaulle} \\
    {\small 94010, Créteil, France} \\
    \and
    Yitzchak Shmalo \\
    {\small Department of Mathematics} \\
    {\small Pennsylvania State University} \\
    {\small University Park, PA 16802, USA} \\
    \and
    Lei Zhang \\
    {\small Institute of Natural Sciences} \\
    {\small Shanghai Jiao Tong University} \\
    {\small Shanghai, 200240, P. R. China}
}
\date{} 

\begin{document}

\maketitle

\begin{abstract}
    
We explore the applications of random matrix theory (RMT) in the training of deep neural networks (DNNs), focusing on layer pruning that reduces the number of DNN parameters (weights). Our numerical results show that this pruning leads to a drastic reduction of parameters while not reducing the accuracy of DNNs and Convolutional Neural Network (CNNs) \footnote{Code for this work can be found at \url{https://github.com/yspennstate/RMT_pruning_2/blob/main/rmt_pruning_2.ipynb}}.  Moreover,  pruning the fully connected DNNs actually increases the accuracy and decreases the variance for random initializations. Our numerics indicate that this enhancement in accuracy is due to the simplification of the loss landscape. We next provide rigorous mathematical underpinning of these numerical results by proving the RMT-based 
 Pruning Theorem. Our results offer valuable insights into the practical application of RMT for the creation of more efficient and accurate deep-learning models.



\end{abstract}

\tableofcontents

\markboth{Leonid Berlyand, Etienne Sandier, Yitzchak Shmalo, Lei Zhang}{Application of RMT to Deep Learning}

\section{Introduction}
Deep neural networks (DNNs) have become a dominant tool for tackling classification tasks, where objects within a set $S \subset \mathbb{R}^n$ are categorized. DNNs are trained on labeled datasets $T \subset \mathbb{R}^n$ by optimizing a loss function such as a cross-entropy loss function in \eqref{loss_function} to maximize classification accuracy. Through this training process, DNNs have achieved state-of-the-art results on many real-world classification challenges, including handwriting recognition \cite{LBD}, image classification \cite{krizhevsky2017imagenet}, speech recognition \cite{hinton2012deep}, and natural language processing \cite{sutskever2014sequence}.

Overfitting is a common challenge for the training of DNNs, which occurs when the model's complexity results in memorization of the training data rather than generalization to new data. Consequently, despite high training set accuracy, the model's performance on the test set deteriorates. To counteract overfitting,  different regularization techniques such as dropout \cite{srivastava2014dropout}, early stopping \cite{prechelt2012early}, and weight decay regularization \cite{Park2023} have been developed.

Recently, Random Matrix Theory (RMT) has been used in deep learning for addressing overfitting \cite{martin2021implicit,mahoney2019traditional}. Similar works have used RMT to obtain  RMT-based stopping criteria, see \cite{meng2023impact}, and regularization, see \cite{xiao2023heavy}. It has also been shown that RMT can be used to predict DNN performance without access to the test set, see \cite{martin2021predicting,martin2020heavy}, and in general to study the spectrum of weight layers \cite{thamm2022random} and the input-output Jacobina matrix \cite{pastur2020random,pastur2023random}. RMT-based initializations were also studied in \cite{saada2023initialisation}. However, these RMT works in deep learning focused on issues other than utilizing RMT-based pruning in DNNs, which is the focus of our work. 
  Specifically, we study the applications of RMT for pruning DNNs during training. We present numerical simulations on simple DNN models trained on the MNIST, Fashion MNIST, and CIFAR-10 datasets. Generally, these RMT techniques can be extended to other DNN types and any fully connected or convolutional layer of pre-trained DNNs to reduce layer parameters while preserving or enhancing accuracy.

We chose the MNIST, Fashion MNIST, and CIFAR-10 datasets because they balance complexity and efficiency, allowing us to conduct numerous experiments on very large DNNs, which is necessary for assessing the overall behavior of our algorithm within the RMT framework. MNIST is the simplest of these datasets and is easy to train on, while Fashion MNIST and CIFAR-10 are a little more complex. All of these datasets are complex enough to demonstrate the effectiveness of our MP-based pruning method yet simple enough to enable quick and extensive experimentation.
  
Our RMT pruning approach simplifies DNNs, enabling them to find deeper minima on the loss landscape of the training set. As a result, DNNs can achieve higher accuracy directly on the training set. DNNs, during their training, navigate a complex, multi-dimensional loss landscape in search of the global minimum - the optimal solution. However, the nature of these landscapes can often be rugged, filled with numerous sub-optimal local minima that trap the learning process. By implementing RMT pruning, the landscape becomes smoother, less prone to local minima, and more navigable for the learning algorithm. This makes the optimization process more efficient and enables the DNN to find deeper minima for the loss of the training set.

The works \cite{ yang2020learning,xue2013restructuring,cai2014fast,anhao2016svd,xu2019trained} utilized Singular Value Decomposition (SVD) to eliminate small singular values from DNN weight matrices. This pruning of singular values was used to prune the parameters in the weight layer matrices of the DNNs, similar to our work. This pruning was based on techniques such as energy ratio thresholds and monitoring the error of a validation set. However, this energy threshold comes from empirical observations. In contrast, the Marchenko-Pastur (MP) threshold used for pruning in our work is justified theoretically by RMT and applied to fully connected networks and simple Convolutional Neural Network (CNNs) to establish an agreement between theory and numerics. 

Other pruning methods can be found in \cite{vadera2022methods}, in which the authors categorize over 150 studies into three pruning categories: methods that use magnitude-based pruning, methods that utilize clustering to identify redundancy, and methods that use sensitivity analysis to assess the effect of pruning. Our work is mostly related to the first method of pruning; we use  MP-based pruning to prune small singular values, together with sparsification to prune all weights of the DNN bigger than some threshold (and set them to 0), see Subsections \ref{MP+sp_1}, \ref{MP+sp_2} and \ref{MP+sp_3}. To the best of our knowledge, we are the first to use the MP distribution as a threshold for pruning. Furthermore, the Pruning Theorem \ref{the_main_result_RMT} provides mathematical justification for the MP-based pruning approach, while Lemma \ref{main_result_remove _R} provides mathematical justification for the sparsification approach. Other pruning methods, such as \textit{pruning at initialization}, have also been used; see \cite{price2021dense}.

In \cite{staats2022boundary}, the MP distribution was used to decrease the size of large singular values, which allows for the extraction of the denoised matrix from the original (noisy) one.  Then, a validation set was used to determine the SVD pruning threshold for filtering noisy data in DNN classification. However, there are important distinctions with our work. First, we use an MP threshold for directly pruning weights without access to a validation set. Second, the pruning in \cite{staats2022boundary} is done after training, whereas pruning in our work is done during training, corresponding to a different improvement mechanism of accuracy improvement via simplification of loss landscape during training. Thirdly, the authors of \cite{staats2022boundary} focused on how pruning small singular values can improve the accuracy of DNNs trained on noisy data. On the other hand, we study how the pruning of small singular values can guide the pruning of weights of the DNN that are random due to initialization. Moreover, our theoretical approach applies to both sources of randomness: initialization of weights and noise in the data. 


\textit{In contrast with the above numerical works, our work also provides rigorous theoretical underpinning on the relation between RMT-based pruning of DNN weight matrices and accuracy.} 
To this end, we establish rigorous mathematical results (the Pruning Theorem), which explain the effectiveness of our RMT-based algorithm. The theoretical results will help elucidate the underlying mechanisms of the numerical algorithm, demonstrating why it successfully reduces the number of parameters in a DNN without reducing accuracy. These theoretical results will allow for the development of RMT-based pruning for state-of-the-art DNNs such as ResNets and VITs. 

The remainder of this paper is organized as follows:

In Section 3, we present an overview of DNN training.
In Section 4,  we present the numerical results of this paper. In Section 5, we present the Pruning Theorem.  

\section{Acknowledgements}

The work of LB was partially supported by NSF grant DMS-2005262 and NSF grant IMPRESS-U 2401227. LB and ES are grateful to the Labex Bézout Foundation for supporting the stay of LB while visiting Université Paris-Est, which helped facilitate the collaboration between ES, LB, and YS on this work. The authors thank J. Tanner for bringing several relevant publications to our attention. 

\section{Background on Deep Learning}
\label{backround_deep_learning}

DNNs have become a widely-used method for addressing classification problems, in which a collection of objects $S \subset \R^n$ is assigned to one of $K$ classes. The objective is to approximate an exact classifier $\psi$, which maps an element $s \in S \subset \R^n$ to a probability vector $(p_1(s), \dots, p_K(s))$. In this vector, $p_{i(s)}=1$ and $p_{j}=0$ for $j \neq i(s)$, where $i(s)$ denotes the correct class for $s$. The exact classifier $\psi$ is known only for a training set $T$, and DNNs are trained to approximate $\psi$ by constructing a parameterized classifier $\phi(s,\alpha)$ with the aim of extending $\psi$ from $T$ to all of $S$ via $\phi(s, \alpha)$.

This is accomplished by finding parameters $\alpha$ that allow $\phi(s,\alpha)$ to map $s \in T$ to the same class as $\psi$ while maintaining the classifier's ability to generalize to elements $s \in S$. The parameters $\alpha$ are optimized by minimizing a loss function, aiming to enhance the accuracy as the loss declines.

In this study, a DNN is represented as a composition of two functions: the softmax function $\rho$ and an intermediate function $X(\cdot,\alpha)$. The function $X(\cdot,\alpha)$ is defined as a composition of affine transformations and nonlinear activations, as follows:
\begin{itemize}
\item $M_l(\cdot,\alpha_l)$ is an affine function that maps $\mathbb R^{N_{l-1}}$ to $\mathbb R^{N_{l}}$, and depends on a parameter matrix $W_l$ of size $N_{l} \times N_{l-1}$ and a bias vector $\beta_l$.
\item $\lambda: \R^m \mapsto \R^m$ is a nonlinear activation function.
\item $X(\cdot,\alpha)=\lambda \circ M_k\cdots \lambda \circ M_1$, where $k$ is the number of layers in the DNN. Note that each $\lambda$ here might be different from the others, given that the domains of each differ. 
\end{itemize}

Lastly, $\rho$ is the softmax function, which normalizes the output of $X(\cdot,\alpha)$ into probabilities. The components of $\rho$ are calculated as:
\begin{equation}
\label{soft_max_new}
\rho_i(s,\alpha)=\frac{\exp(X_i(s,\alpha))}{\sum_{i=1}^{K}\exp(X_i(s,\alpha))}.
\end{equation}

The DNN's output, $\phi$, is a vector representing the probabilities of an object $s \in S$ belonging to a particular class $i$. $\phi=\phi(s,\alpha)$, where $\alpha \in \R^\nu$ is the DNN's parameter space and $\nu \gg 1$ is the dimension of the parameter space. The goal is to train the DNN $\phi$ to approximate the exact classifier by minimizing a loss function, such as the cross-entropy loss function
\begin{equation}
\bar L(\alpha)=-\frac{1}{|T|}\sum_{s\in T} \log\left(p_{i(s)}(s,\alpha)\right).
\label{loss_function}
\end{equation}

Training a DNN essentially involves traversing a high-dimensional, non-convex loss landscape to locate the global minimum. But the complexity of these landscapes frequently leads to local minima, saddle points, or flat regions, all of which trap the learning process, impeding it from reaching an optimal solution \cite{dauphin2014identifying}. These issues amplify as the dimensionality (and thus the complexity) of the DNN increases \cite{choromanska2015loss}.
``
In local minima, the gradient of the loss function equals zero, but it is not the global minimum, thus, the algorithm incorrectly assumes it has found the best possible solution. Saddle points, on the other hand, are points where the gradient is zero, but they are neither a global nor a local minimum. They are particularly problematic in high-dimensional spaces, a common feature in deep learning.

To overcome these obstacles, various sophisticated optimization techniques are employed. For example, optimization algorithms such as Momentum, RMSProp, or Adam are designed to prevent getting stuck by adding additional components to the update rule, which can help in navigating the complex optimization landscape. These methods imbue the optimization process with a form of `memory' of previous gradients, enabling it to continue its search even in flat regions, hence helping to escape local minima and saddle points.

DNNs, with their intricate and numerous parameters, offer formidable modeling capabilities \cite{goodfellow2016deep}. However, the same attribute that enables their power can also serve as a curse during the training process. Theoretically, DNNs, due to their extensive parameterization, should reach high levels of accuracy on their training sets \cite{zhang2021understanding}. But in practice, the accuracy on the training set can often plateau, suggesting the DNN is getting stuck in a local minimum or saddle point of the loss function \cite{ge2015escaping}. This forms a critical impediment, limiting the achievable accuracy on both the training and test sets \cite{hochreiter1997flat}.

Thus, despite the vast number of parameters, DNNs can often find themselves stranded in areas of poor performance. This seemingly paradoxical occurrence is attributable to the interplay between the DNN's architecture, the data it is training on, and the optimization process being employed \cite{goodfellow2016deep}. Factors like poor initialization, inappropriate learning rates, or the vanishing/exploding gradients problem can cause the DNN to settle in sub-optimal regions of the loss landscape \cite{glorot2010understanding}.

Techniques like gradient clipping \cite{pascanu2013difficulty} can aid in overcoming these issues. Regularization techniques, which either penalize complex models or enforce sparsity in the weight matrix, can also assist in avoiding local minima \cite{goodfellow2016deep}. Nonetheless, these methods do not alter the fundamental structure of the loss landscape, indicating that the problem of local minima remains \cite{ge2015escaping}.

The potential of the RMT approach stands out in this context.  We show that utilizing RMT in pruning the DNN's weight layers simplifies the loss landscape. This simplification reduces the incidence of local minima and saddle points, aiding the optimization process in its quest for a global minimum. In doing so, the DNN might attain higher levels of accuracy on the training set directly without reaching a plateau. This results in an overall enhancement in model performance, as higher training set accuracy generally translates to improved performance on the test set, assuming overfitting does not occur.

\section{Numerical Algorithm and Experiments}
\label{num_results}

In this section, we focus on the training of two DNNs: the normal DNN, which keeps all of its singular values, and a pruned DNN based on algorithm \ref{algo_1}, see Subsection \ref{overallpicture}. Each DNN is trained for a predetermined number of epochs, with the number of epochs varying per example. The DNNs are also trained for multiple seeds to ensure the reproducibility of the simulations.  

The performance of the DNNs is evaluated by plotting the average accuracy and variance of accuracy for the different seeds. This allows us to visually compare the performance of both the normal and pruned DNNs. For more numerical results using a slightly different RMT training approach, see \cite{shmalo2023deep}.

\subsection{Numerical Algorithm}

\subsubsection{An overview of the Marchenko-Pastur (MP) distribution and its applications in machine learning}

We start with the MP distribution from RMT. This distribution is of fundamental importance in RMT and has numerous applications, such as signal processing, wireless communications, and machine learning, as described in \cite{vershynin2018high,ge2021large,serdobolskii2000multivariate,couillet2011random}. The MP distribution characterizes the limiting spectral density of large random matrices and conveys information about the asymptotic distribution of eigenvalues in a random matrix, predicting the behavior of random matrices under various conditions. Additionally, the MP distribution is utilized in principal component analysis (PCA) and other dimension reduction techniques, see \cite{abdi2010principal,bro2014principal,ringner2008principal}.

To begin, we introduce the empirical spectral distribution (ESD) of an $N \times M$ matrix $G$ as follows:
\begin{defn}\label{ESD_Definition}

The ESD of an $N \times M$ matrix $G$ is given by:

\begin{equation}
\mu_{G_M} = \frac{1}{M} \sum_{i=1}^M \delta_{\sigma_i},
\end{equation}
where $\sigma_i$ denotes the $i$th  non-zero singular values of $G$, and $\delta$ represents the Dirac measure.
\end{defn}

\begin{thm}[Marchenko and Pastur (1967) \cite{marchenko1967distribution}]
\label{RMT_MP_theorem}
Let $W$ be an $N\times M$ random matrix with $M \leq N$. The entries $W_{i,j}$ are independent and identically distributed random variables with mean $0$ and variance $\sigma^2<\infty$. Define $X = \frac{1}{N} W^T W$. Assuming that $N \to \infty$ and $\frac{M}{N} \to c \in (0,+\infty)$, the ESD of $X$, denoted by $\mu_{X_M}$, converges weakly in distribution to the Marchenko-Pastur probability distribution:

\begin{equation}
\label{MP_distribution}
\frac{1}{2\pi\sigma^2} \frac{\sqrt{(\lambda_+ - x)(x - \lambda_-)}}{cx} \mathbf{1_{[\lambda_-, \lambda_+]}} dx
\end{equation}

with

\begin{equation}
\label{lambda_parameters}
\lambda_\pm = \sigma^2(1\pm\sqrt{c})^2.
\end{equation}
\end{thm}

This theorem asserts that the eigenvalue distribution of a random matrix converges to the Marchenko-Pastur distribution as its dimensions increase. The MP distribution is a deterministic distribution, dependent on two parameters: the variance of the random variables in the initial matrix, $\sigma^2$, and the ratio of the number of columns to the number of rows, $c$.

\subsubsection{Using MP for pruning DNN weights}
\label{overallpicture}
  As stated in Section \ref{backround_deep_learning}, a DNN is a composition of affine functions $M_l$ and non-linear activation functions. The affine functions $M_l$ can be thought of as a $N \times M$ matrix $W_l$ of parameters and a bias vector $\beta_l$. In this work, we only focus on the matrix $W_l$ of parameters. It has been shown that $W_l$ can be studied using the spiked model approach in random matrices, with the ESD of $X_l=\frac{1}{N}W_l^T W_l$ having some eigenvalues which are bigger than $\lambda_+$ and some eigenvalues which are smaller than $\lambda_+$, see \cite{martin2021implicit,staats2022boundary}. In this paper, we focus on weight layer matrices $W_l$ which were initialized in such a way that $\sqrt{N}W_l(0)$ satisfy the assumptions of $W$ given in Theorem \ref{RMT_MP_theorem}. Then, we look at the ESD of $B_l=W_l^TW_l$, without normalizing by $\frac{1}{N}$. This setting is more applicable for the situation in which the components of $R$ are i.i.ds taken from $N(0,\frac{1}{N})$.  

Thus, we take $B_l(t)=W_l(t)^T W_l(t)$, with $W_l(t)$ a $N \times M$ weight of the $l$th layer matrix at time $t$ of DNN training. We use RMT to study the \textit{deformed} matrix $W_l(t)=R_l(t)+S_l(t)$, with $R_l(t)$ random and $S_l(t)$ a deterministic matrix. One can assume that during training we go from $W_l(0)=R_l$ (i.e. $W_l$ is random) to $W_l(t_\text{final})=R_l(t_\text{final})+S_l(t_\text{final})$, with $\|S_l(t_\text{final})\| \neq 0$ and $t_\text{final}$ the final training time.
Meaning that as $t \to t_\text{final}$, $\|S_l(t)\|$ grows and so $W_l(t)$ becomes less random. 

\textbf{An important question is:} Why does training reduce randomness in weight matrices?

\begin{itemize}
    \item Suppose a DNN has only one weight layer matrix $W$. Before training starts, the matrix $W(0)$ is initialized with DNN weights. Entries of $W$ arranged into a vector $\alpha(0)$ are chosen randomly, meaning $W(0)$ is fully random. A gradient descent step can be written as:
    \begin{equation}
        \alpha(n+1) = \alpha(n) - \tau\nabla L(\alpha(n)).
    \end{equation}
    The loss gradient $\nabla L(\alpha(n))$ is determined by the training data $T$, which is mostly deterministic. That is, when we take a step from $n=0$ to $n=1$, $\alpha(0)$, the random DNN parameters are gradually replaced with deterministic parameters, and this process continues throughout training.
    
    \item \textbf{However,} $T$ is only \emph{mostly} deterministic. Each object $s \in T$ is sampled from a probability distribution and is a random variable, so $T$ contains some randomness.
    
    \item In practice, the randomness of $\alpha(n)$ decreases as $n \to \infty$ ($\|S(n)\|$ increases), but some randomness due to the data remains.

\end{itemize}

 We make the following observations based on the singular values of the weight layer matrix $W_l$:

\begin{itemize}
    \item \textbf{Observation 1:} Singular values $\sigma_i$ of $W_l$ that are smaller than a threshold $\sqrt{\lambda_+}$ are likely to be singular values of $R_l$, where $\lambda_+$ is the upper bound of the MP distribution of $R^TR$. For more on this observation, see \cite{thamm2022random,staats2022boundary}.
    \item \textbf{Observation 2:} $R_l$ does not enhance the accuracy of a DNN. In other words, the random components of the weight layers do not contain any valuable information and, therefore do not improve accuracy, see Lemma \ref{main_result_remove _R}.
\end{itemize}

Based on these observations, the main idea is to remove some randomness from the DNN by eliminating some singular values of $W_l$ smaller than the threshold $\sqrt{\lambda_+}$.  Algorithm \ref{algo_1} describes this procedure.

\begin{algorithm}
\caption{Optimized DNN Training and Pruning for Parameter Efficiency}
\label{algo_1}
\begin{algorithmic}[1]
\Require $\ell$, a predetermined number of epochs; $\tau$, a threshold for the MP fit criteria in Subsection \ref{Conformance_Assessment}; $f(\text{epoch})$, a monotonically decreasing function from $1$ to $0$ (i.e. \eqref{eqn:f}) and for each $1\leq l \leq L$ and weight layer matrix $W_l$ \textbf{state} $split_{l}=false$.
\State Initialize: Train the DNN for $\ell$ epochs. Take $\text{epoch}:=\ell$.

\While{a predefined training condition is met (i.e. $\text{epoch}\leq 100)$}
    \For{each $l$, if $split_{l}=false$ then for weight matrix $W_l$ in the DNN $\phi$}
        \State Perform SVD on $W_l$ to obtain $W_l=U_l\Sigma_lV_l^T$.
        \State Calculate eigenvalues of $ W_l^T W_l$.
        \State Apply BEMA algorithm (see Subsection \ref{finding_lambda}) to find the best fit MP distribution for ESD of $X =  W_l^T W_l$ and corresponding $\lambda_+$.
        \State Check if ESD of $X$ fits the MP distribution using MP fit criteria from Subsection \ref{Conformance_Assessment} and  threshold $\tau$.
        \If{ESD fits the MP distribution}
            \State Eliminate the portion ($1-f(\text{epoch})$) of singular values smaller than $\sqrt{\lambda_+}$ to obtain $\Sigma'$ and form $W'_l=U_l\Sigma'_lV_l^T$.
            \State Use $\Sigma'$ to create $W'_{1,l}=U_l\sqrt{\Sigma'_l}$ and $W'_{2,l}=\sqrt{\Sigma'_l}V_l^T$.
            \If{$W'_{1,l}$ and $W'_{2,l}$ together have fewer parameters than $W'_l$}
            \State Replace $W_l$ in the DNN $\phi$ with  $W'_{1,l}W'_{2,l}$, change $split_l=true$.
            \Else
              \State
                Replace $W_l$ in the DNN $\phi$ with  $W'_l$.
            \EndIf
        \Else
            \State Don't replace $W_l$.
        \EndIf
    \EndFor
    \State Train the DNN for $\ell$ epochs. Take $\text{epoch}:=\text{epoch}+\ell$.
    \For{each $l$, if $split_{l}=true$}
  
    \If{for $W_l:=W'_{1,l}W'_{2,l}$ the ESD of $X_l$ fits the MP distribution with thresholds $\tau$ and $\lambda_+$ \textbf{and} if, we (hypothetically) applied steps 4-12 to $W_l$, the number of parameters in the DNN $\phi$ would decrease}
    \State replace $W'_{1,l}W'_{2,l}$ with $W_l$ and $split_l=false$.
    \Else
    \State Don't change anything.
    \EndIf 
    \EndFor
\EndWhile
\end{algorithmic}
\end{algorithm}

\subsubsection{MP and Tracy Widom distribution for DNN training}
\label{RMT_DNN_Method}

We use the bulk eigenvalue matching analysis (BEMA) algorithm (see Subsection \ref{finding_lambda}) to find the MP distribution that best fits the ESD of $X_l=W^T_l W_l$, with $W_l$ a weight layer matrix. We then use the Tracy Widom distribution (see \cite{ke2021estimation}) to find a confidence interval for the $\lambda_+$ of the ESD of $X_l$ and then prune the small singular values of $W_l$ based on the MP-based threshold $\sqrt{\lambda_+}$, see Subsection \ref{finding_lambda} for more details on the Tracy Widom distribution and the BEMA algorithm. The steps of this procedure are shown in Algorithm \ref{algo_1}.

In step $9$ of Algorithm \ref{algo_1}, we ensure not to eliminate all of the small singular values (i.e., singular values whose corresponding eigenvalues fall within the MP distribution). Striking a balance between removing the smaller singular values and retaining some is found to be essential. Removing all of the smaller singular values might result in the underfitting of the DNN, thereby inhibiting its learning capability. Conversely, retaining some of the smaller singular values adds a degree of randomness in the weight layer matrix $W_l$, which is found to impact the DNN's performance positively.

\begin{remark}
    
Note that it is possible to continue splitting the matrices $W_{\ell}$, $W'_{\ell}$, and so on. This also improves accuracy for fully connected layers. However, we found that recombining and splitting the original matrix works better.
\end{remark}

\subsection{Numerical Experiments}

Numerical simulations presented in this paper show that MP-based pruning enhances the accuracy of DNNs while reducing the number of DNN weights (parameters). The first set of numerical simulations employs fully connected DNNs trained on the MNIST and Fashion MNIST datasets, revealing that MP-based pruning during training improves accuracy by 20-30\% while reducing the parameter count by 30-50\%. These findings are consistent across various architectures and weight initializations, underscoring the consistency of the MP-based pruning approach. Further, the combination of this approach and sparsification (eliminating parameters below a certain threshold, see \cite{vadera2022methods}) leads to even more significant reductions in parameters (up to 99.8\%) while increasing accuracy (by 20-30\%). This reduction in parameters is greater than what is achievable through sparsification alone (99.5\%); see Subsection \ref{MP+sp_1}.

Unless stated otherwise, in all numerical simulations, the parameter matrices were initialized from $N(0,1/N)$, with $N$ the number of input features, while the bias vectors were initialized to $0$. This initialization is closest to the theoretical work in this paper, which is why we use it. The ReLU activation function was applied after every layer, including the final layer. While it might be easier to train DNNs with other initializations and architectures, we found that we obtained the highest accuracies when training with the affirmation structure while using MP-based pruning. For example, using a fully connected DNN, we obtained a $91.27\%$ accuracy on the Fashion MNIST test set, which is the highest accuracy we observed on the data set using a fully connected DNN (see Subsection \ref{fully_connected_MNIST}). MP-based pruning also increases the accuracy of fully connected DNNs that do not have an activation function on the final layer; for example, see Subsection \ref{MP_reg}.

For simplicity of presentation, we choose to demonstrate the MP-based pruning for fully connected DNNs. In short, the idea is as follows. First, we observe that weight layer matrices $W_l$ have singular values of two types: those that contain information and the ones that don't and, therefore, can be removed (pruned). This separation is done via MP threshold $\sqrt{\lambda_+}$. Furthermore, we demonstrate that pruning based on this MP threshold preserves DNN accuracy. These numerical findings are supported by rigorous mathematical results (Theorem \ref{Cor_1}). In fact, for the case of fully connected layers we show numerically that MP-based pruning simplifies the loss landscape, leading to a significant increase in DNN accuracy (by 20-30\%). Finally, we show that a combination of MP-based pruning with sparsification preserves or even increases accuracy while reducing parameters by $99.8\%$ vs. MP-based pruning alone, with a reduction of 30-50\%, or sparsification alone, with a reduction of 99.5\%. Our theoretical results also explain why sparsification does not reduce accuracy; see Lemma \ref{main_result_remove _R} and Remark \ref{sparsification}.      


Our numerics explores the application of MP-based pruning on DNNs that already achieve relatively high accuracy on MNIST (Section \ref{fully_connected_MNIST}), Fashion MNIST (Section \ref{full_Fash_MNIST}), and CIFAR10 datasets (Section \ref{CIFAR-10}), including those using Convolutional Neural Networks (CNNs) and sparsification techniques (Section \ref{MP+sp_1},\ref{MNIST_Fashion_Con_Red},\ref{MP+sp_2} \ref{MP+sp_3}). Our results show a substantial reduction in parameters (over 95\%) while preserving accuracy through a combination of MP-based pruning during training and post-training sparsification, surpassing the efficiency of using sparsification alone (80-90\% reduction in parameters). These extensive simulations for various architectures and initializations demonstrate the consistency and wide applicability of MP-based pruning in optimizing DNN performance. 

To see how MP-based pruning performed on a simple regression problem, see Subsection \ref{MP_reg}.

\subsubsection*{Training and testing procedure}

The training and testing procedure for each network consists of the following steps:

\begin{enumerate}
\item We follow the standard partition for MNIST and Fashion MNIST, with 60,000 images for training and 10,000 images for testing. 
\item We train the network for a certain number of epochs.
\item We test the network after each epoch and store the accuracy for later comparison.
\item For the pruned DNN, we apply algorithm \ref{algo_1} after a set number of epochs (defined by the split frequency).
\end{enumerate}

\subsubsection{Training of  fully connected DNNs on MNIST: simplifying the loss landscape}
\label{fully_connected_MNIST}

The results of the simulations are presented in the examples below. These examples show figures and tables which compare the average accuracy and variance of accuracy for both the normally trained and pruned DNNs, as well as the average loss and number of parameters in both DNNs. Detailed discussion and analysis of these results are presented in the examples.

\textbf{Training hyperparameters:}

\begin{itemize}
\item Split frequency ($\ell$) (every how many epochs we split the pruned DNN and remove small singular values): 7
\item goodness of fit (GoF or $\tau$) = $.7$
\end{itemize}

See Subsection \ref{hyperparam_1} for the other hyperparameters in these simulations.


\begin{remark}
    The algorithm for finding the GoF parameter is given in Subsection \ref{Conformance_Assessment}. It is used to determine if the assumption given in Theorem \ref{the_main_result_RMT}, that $W_l=R_l+S_l$, is reasonable and that the weight layers can reasonably be modeled as a spiked model (that is $W_l$ is a deformed matrix).  
\end{remark}

\begin{ex}

We conducted several numerical simulations to compare the performance of the normal DNNs, trained using conventional methods, and pruned DNNs, trained using our RMT approach.

In all simulations, the networks start with different initial topologies and are trained over a course of 40 epochs. The portion of singular values smaller than $\sqrt{\lambda_{+}}$ that we retain (see step 3 in algorithm \ref{algo_1}) is given by the linear function:

\begin{equation}
    f(\text{{epoch}}) = \max\left(0, -\frac{1}{30} \cdot \text{{epoch}} + 1\right).
    \label{eqn:f}
\end{equation}

The topologies and the results of the simulations are summarized in Table 1 and Fig. \ref{Comparison}. The results indicate a consistent trend across different topologies: the pruned DNNs outperform the normal DNNs in terms of accuracy on the test set while also displaying smaller variance across multiple runs. Furthermore, the pruned DNNs consistently achieve a significant reduction in parameters by the end of the training, see Remark \ref{reduction_paramters_1}.

\begin{table}[h!]
\centering
\small 
\begin{tabular}{|c|c|c|}
\hline
\textbf{Initial Topology} & \textbf{Unpruned DNN Accuracy} & \textbf{Pruned DNN Accuracy}\\
\hline
$[784, 3000,3000,2000, 500, 10]$ & $\sim$85\% & $\sim$98.5\%\\
\hline
$[784, 1000,1000,1000, 500, 10]$ & $\sim$70\% & $\sim$98.5\%\\
\hline
$[784, 2000,2000,1000, 500, 10]$ & $\sim$70\% & $\sim$98.5\%\\
\hline
$[784, 1500,3000,1500,500, 10]$ & $\sim$70\% & $\sim$98.5\%\\
\hline
$[784, 1000,1000,1000, 500, 10]$ (GoF 1) & $\sim$70\% & $>$98.5\% \\
\hline
\end{tabular}
\caption{Performance of normal and pruned DNNs for different initial topologies.}
\label{Table:1}
\end{table}

\begin{figure}[h!]
\centering
\begin{subfigure}[b]{0.274\textwidth}
\includegraphics[width=\textwidth]{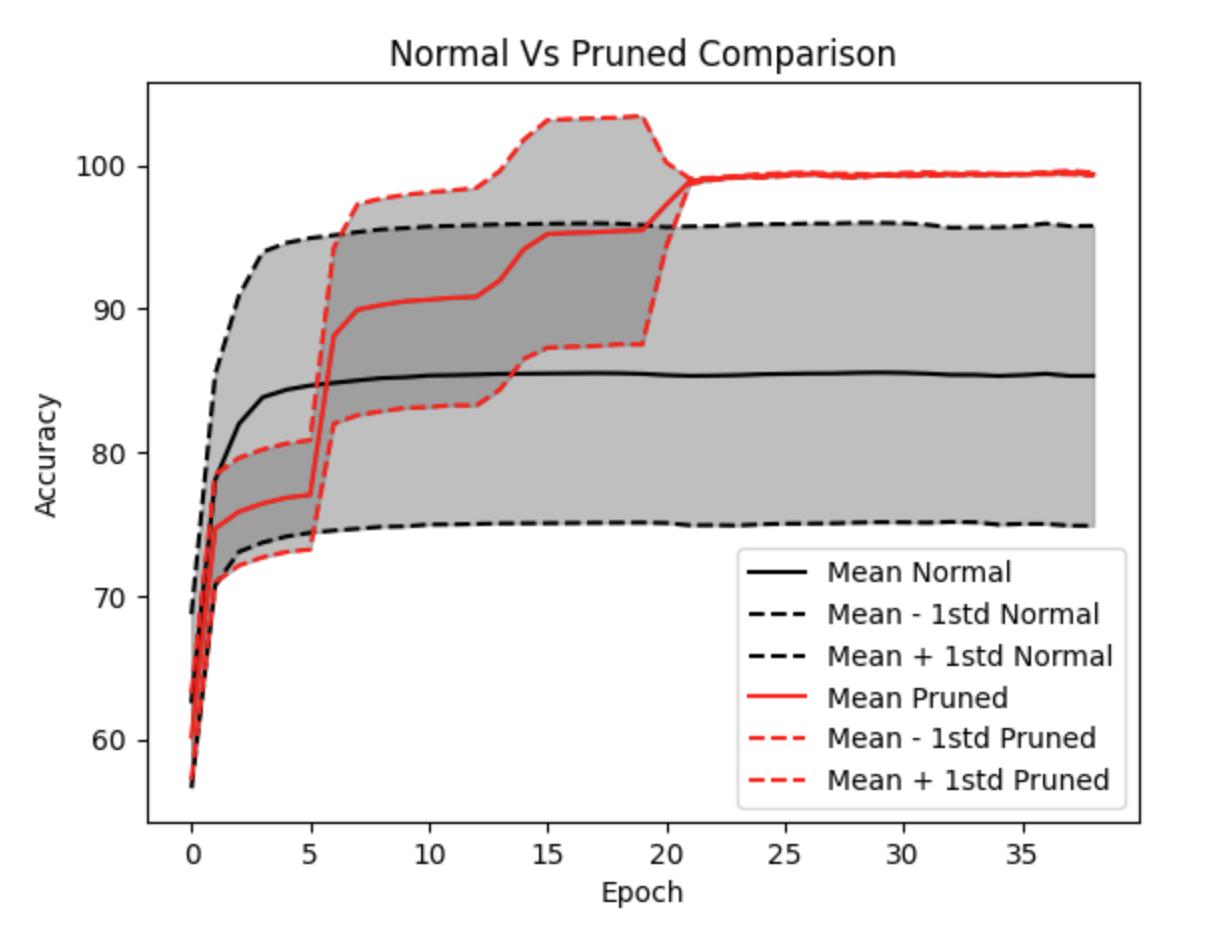}
\caption{}
\end{subfigure}
\begin{subfigure}[b]{0.3\textwidth}
\includegraphics[width=\textwidth]{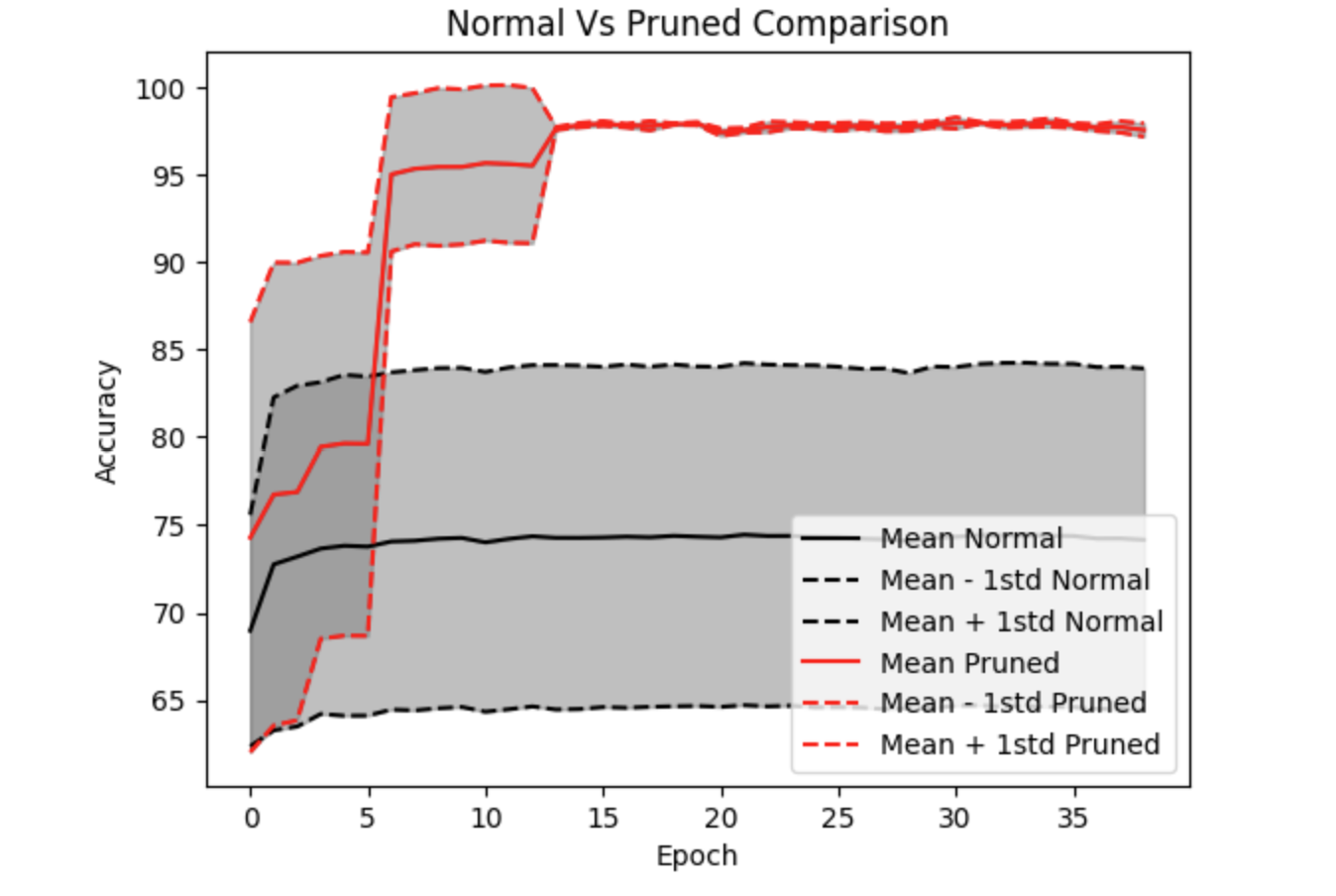}
\caption{}
\end{subfigure}
\begin{subfigure}[b]{0.3\textwidth}
\includegraphics[width=\textwidth]{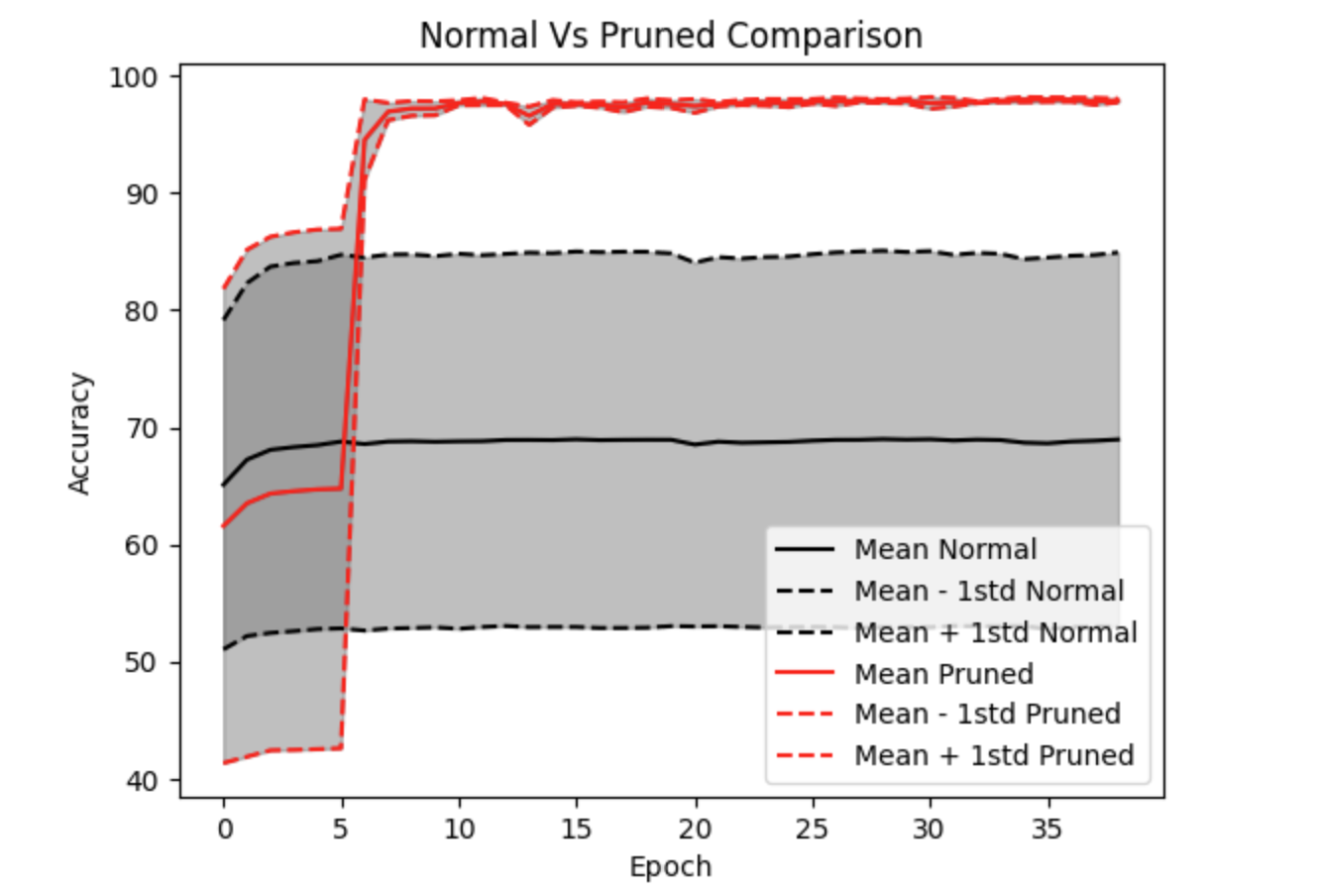}
\caption{}
\label{fig:ex3}
\end{subfigure}
\begin{subfigure}[b]{0.33\textwidth}
\includegraphics[width=\textwidth]{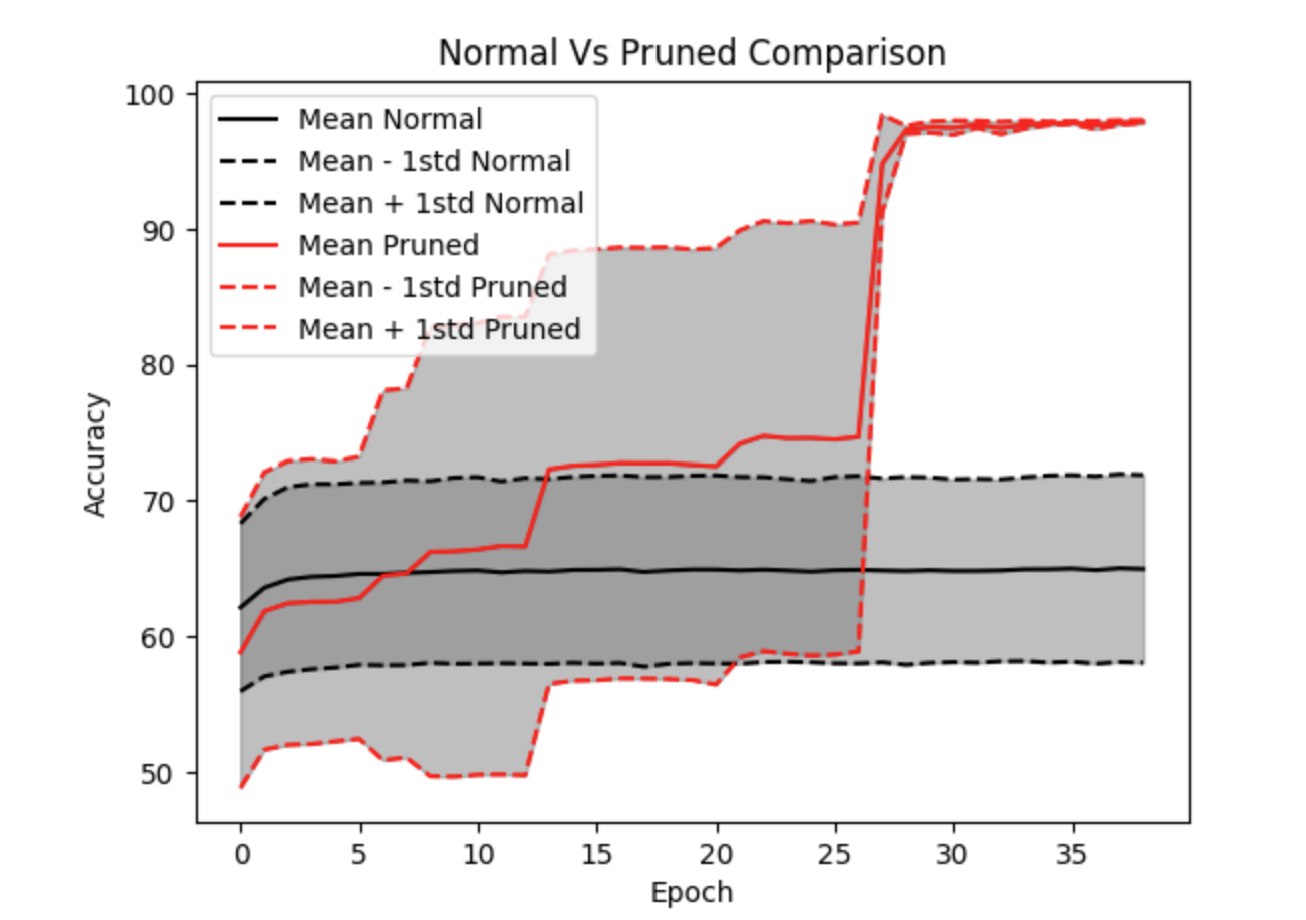}
\caption{}
\label{fig:ex4}
\end{subfigure}
\begin{subfigure}[b]{0.3\textwidth}
\includegraphics[width=\textwidth]{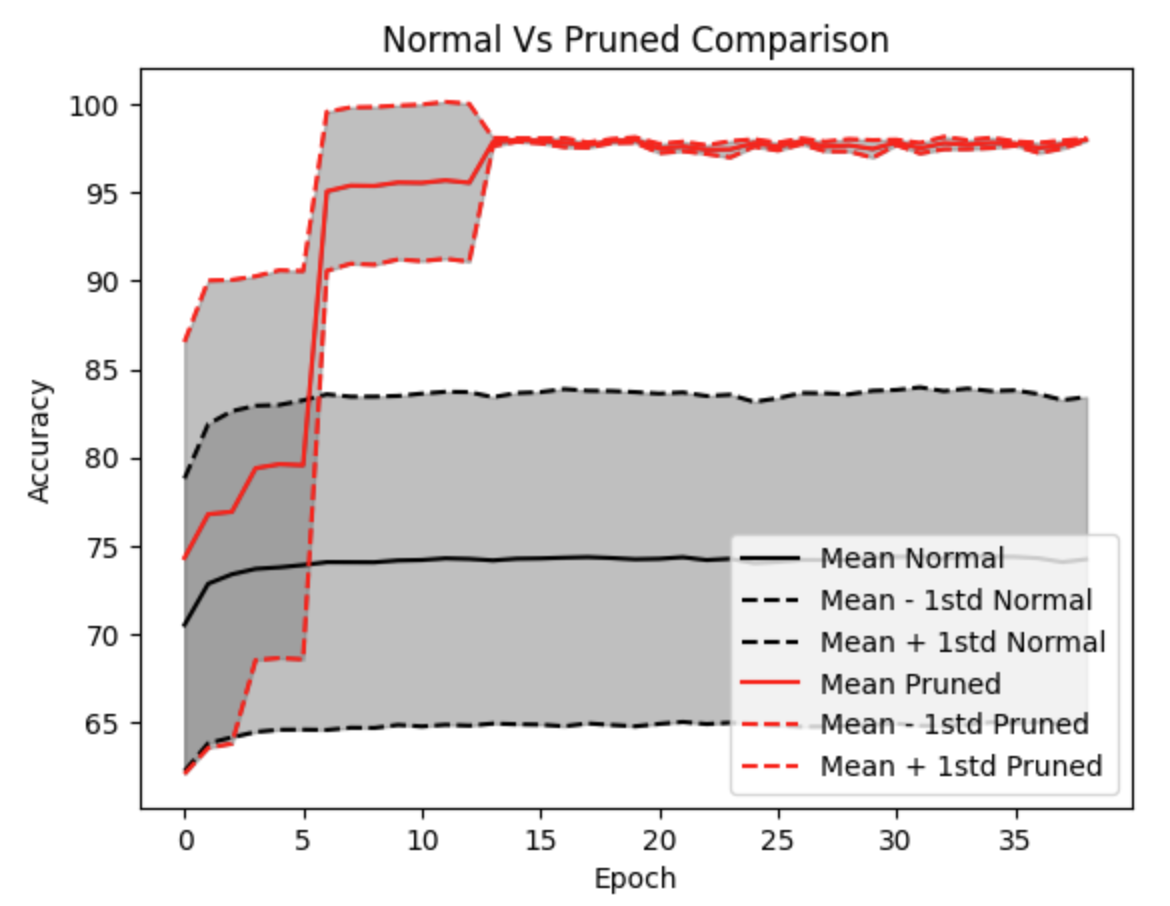}
\caption{}
\label{fig:ex5}
\end{subfigure}
\caption{Comparison of Normal DNN, trained normally, and pruned DNN, trained using the RMT approach on the test set. The sub-figures correspond to the different initial topologies: (a) $[784, 3000,3000,2000, 500, 10]$, (b) $[784, 1000,1000,1000, 500, 10]$, (c) $[784, 2000,2000,1000, 500, 10]$, (d) $[784, 1500,3000,1500,500, 10]$, and (e) $[784, 1000,1000,1000, 500, 10]$ with a larger goodness-of-fit parameter of 1.}
\label{Comparison}
\end{figure}
\end{ex}

\begin{remark}
    In these examples, the goodness of fit parameter can be very large (even 1) and does not change the accuracy of the DNN. This is not always the case, especially for state-of-the-art pre-trained DNNs, as we will show in future works. See also Subsection \ref{MNIST_Fashion_Con_Red} for an example of when GoF must be smaller.   
\end{remark}

\begin{remark}
\label{reduction_paramters_1}

In Table \ref{tab:my_label}, we observe the effect of our RMT training approach on the number of parameters in our DNN with different topologies. Each topology started with a fixed number of parameters, and by the end of training, we see a significant reduction in the number of parameters across all topologies for the pruned DNN. For each topology and across all seeds, the reduction in the number of parameters was consistent, indicating the robustness of our training process in pruning the network while maintaining performance. 

\begin{table}[h]
\centering
\begin{tabular}{|c|c|c|c|}
\hline
\textbf{Topology} & \textbf{Initial Parameters} & \textbf{Final Parameters} & \textbf{Percentage Reduction} \\ 
\hline
$[784, 3000, 3000, 2000, 500, 10]$ & 18,365,510 & 10,471,510 & 42.98\% \\
$[784, 1000, 1000, 1000, 500, 10]$ & 3,292,510 & 1,678,234 & 49.03\% \\
$[784, 2000, 2000, 1000, 500, 10]$ & 8,078,510 & 4,619,376 & 42.82\% \\
$[784, 1500, 3000, 1500, 500, 10]$ & 10,937,510 & 5,554,966 & 49.21\% \\
\hline
\end{tabular}
\caption{DNN topology, initial and final parameters for the pruned DNNs, with percentage reductions.}
\label{tab:my_label}
\end{table}
    
\end{remark}

\begin{figure}[h!]
\centering
\begin{subfigure}[b]{0.3\textwidth}
\includegraphics[width=\textwidth]{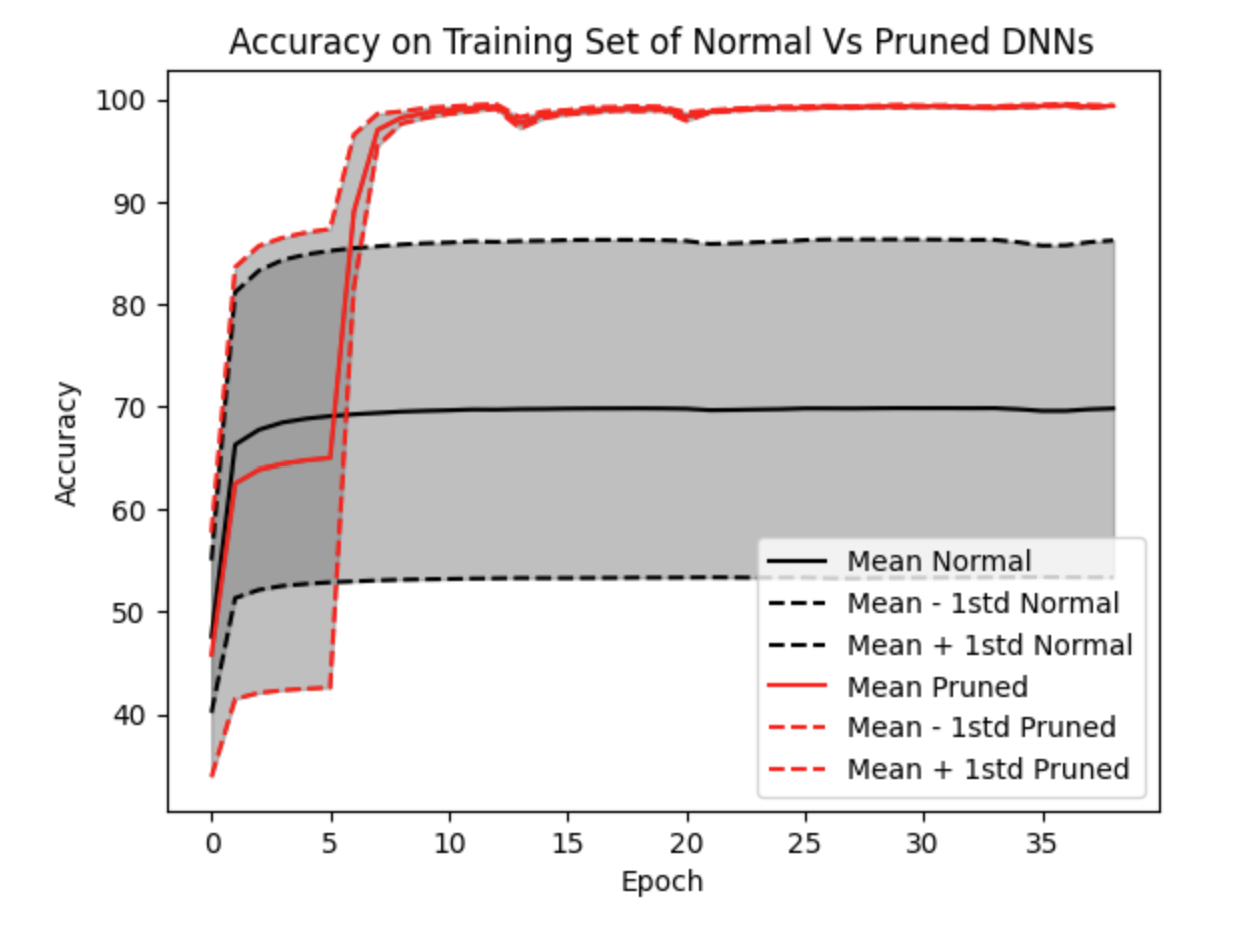}
\caption{}
\label{fig:ex1_train}
\end{subfigure}
\begin{subfigure}[b]{0.3\textwidth}
\includegraphics[width=\textwidth]{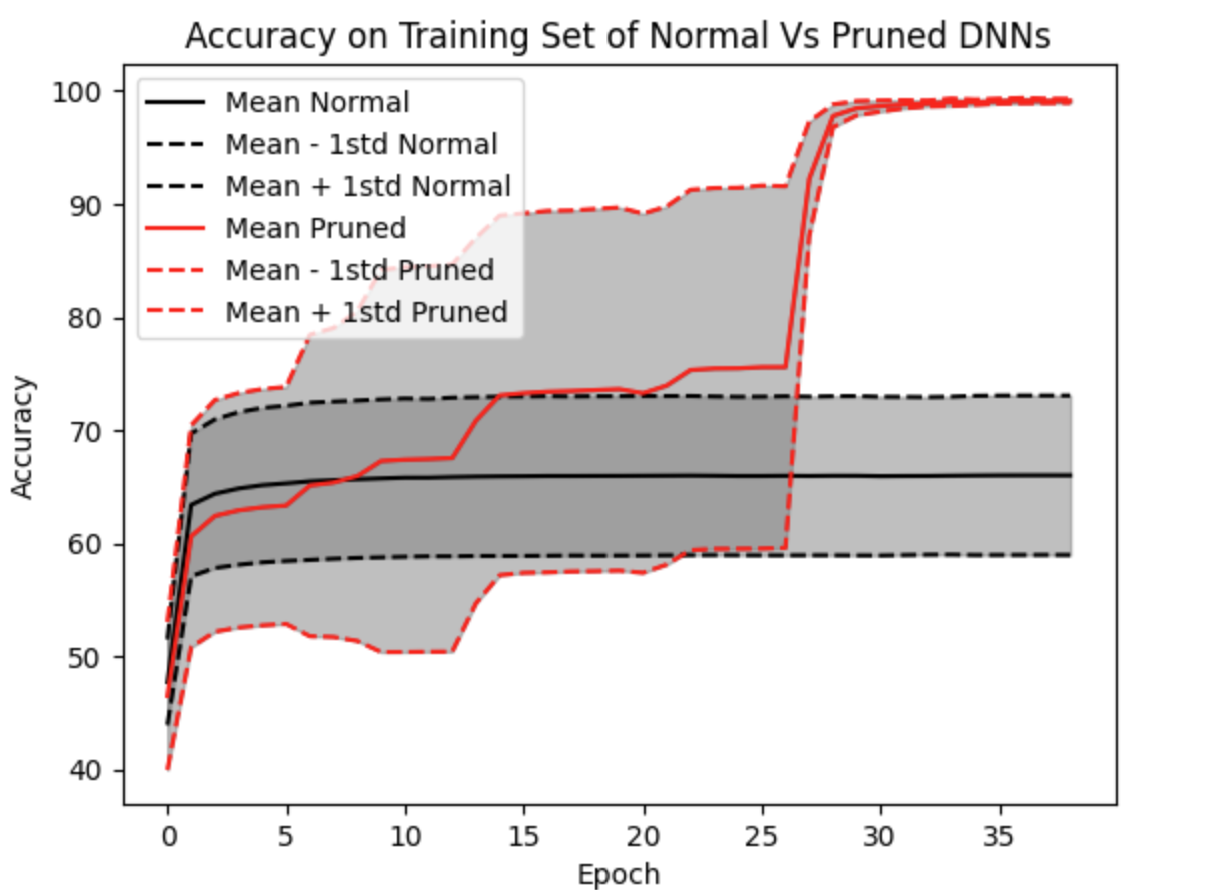}
\caption{}
\label{fig:ex2_train}
\end{subfigure}
\caption{Comparison of Normal DNN, trained normally, and pruned DNN, trained using the RMT approach on the training set. The sub-figures correspond to the different initial topologies: (a) $[784, 2000,2000,1000, 500, 10]$, (b) $[784, 1500,3000,1500,500, 10]$. The other examples in Fig. \ref{Comparison} have similar-looking accuracies on their training set.}
\label{training_set_example}
\end{figure}

\begin{table}[h!]
\centering
\begin{tabular}{|c|c|c|c|c|}
\hline
\textbf{Topology} & \textbf{Type} & \textbf{Training Loss} & \textbf{Test Loss} \\ 
\hline
$[784, 3000,3000,2000, 500, 10]$ & Normal & 1.324090 & 150.541977 \\
 & Pruned & 0.015084 & 9.211179 \\
\hline
$[784, 1000,1000,1000, 500, 10]$ & Normal & 0.938041 & 127.608980 \\
 & Pruned & 0.001350 & 14.543617\\
\hline
$[784, 2000,2000,1000, 500, 10]$ & Normal & 0.567025 & 70.037265 \\
 & Pruned & 0.019599 & 13.676276 \\
\hline
$[784, 1500,3000,1500,500, 10]$ & Normal & 1.037013 & 146.353504 \\
 & Pruned & 0.009206 & 9.519642 \\
\hline
\end{tabular}
\caption{Comparison of training and test losses between normal and pruned DNNs for different topologies.}
\label{tab:loss_example}
\end{table}

\paragraph{Simplification of loss landscape for more efficient training}
As mentioned, a common challenge with DNNs is the complex and high-dimensional loss landscape due to the large number of parameters. This complexity often leads to local minima or saddle points that hinder optimal training. However, by using this clever RMT pruning approach, we effectively eliminate redundant parameters, thereby simplifying the loss landscape. This simplification allows us to avoid suboptimal local minima and converge more readily to a global minimum.

This improved optimization efficiency is evident when comparing the loss and accuracy of the original and pruned DNNs on both training and test sets; see Table \ref{tab:loss_example} and Fig. \ref{training_set_example}. The pruned DNNs achieve lower loss and higher accuracy on the training set directly, indicating that they are finding deeper minima in the loss landscape and avoid suboptimal local minima.

Thus, the RMT pruning approach not only significantly reduces the complexity of DNNs but also enhances their performance by improving their optimization efficiency. Despite the reduction in parameters, the pruned DNNs still exhibit excellent performance on both training and test sets (even higher accuracy than the normally trained DNNs), demonstrating the effectiveness of this approach.

\paragraph{Simplifying the loss landscape for fully connected DNNs on Fashion MNIST}
\label{full_Fash_MNIST}

 In this section, we trained the normal and pruned DNNs on the data set Fashion MNIST and we look at the performance of both DNNs on the training and test set. Again the pruned DNN obtains higher accuracy and lower loss on both the training and test sets, evidence that pruning the DNN using RMT simplifies the loss landscape and allows the DNN to find a deeper global minimum.   

\textbf{Training hyperparameters:}

\begin{itemize}
    
\item Split frequency (every how many epochs we split the modified DNN and remove small singular values): 7
\item goodness of fit (GoF) = $.7$
\end{itemize}

The other hyperparameters for the simulations in this subsection can be found in Subsection \ref{hyperparam_2}.

In all simulations, the networks start with different initial topologies, are trained over a course of 70 epochs, and the portion of singular values smaller than $\sqrt{\lambda_{+}}$ that we retain is given by the linear function:

\begin{equation}
    f(\text{{epoch}}) = \max\left(0, -\frac{1}{60} \cdot \text{{epoch}} + 1\right)
\end{equation}

\begin{ex}

The topologies and the results of the simulations are summarized in Table \ref{results_Fashioi_MNIST} and Fig. \ref{Comparison_fashion_MNIST}. 

As with MNIST, in the case of training on Fashion MNIST the results indicate a consistent trend across different topologies: the pruned DNNs outperform the normal DNNs in terms of accuracy on the test set while also displaying smaller variance across multiple runs. Furthermore, the pruned DNNs consistently achieve a significant reduction in parameters by the end of the training, see Table \ref{paramters_Fashion_MNIST}.

\begin{table}[h!]
\centering
\small 
\begin{tabular}{|c|c|c|}
\hline
\textbf{Initial Topology} & \textbf{Unpruned DNN Accuracy} & \textbf{Pruned DNN Accuracy}\\
\hline
$[784, 2000,4000,2000,500, 10]$ & $\sim$70\% & $\sim$89\%\\
\hline
$[784, 2000,2000,2000,2000,1000, 500, 10]$ & $\sim$65\% & $\sim$89\%\\
\hline
$[784, 3000,4000,3000,500, 10]$ & $\sim$70\% & $\sim$89\%\\

\hline
\end{tabular}
\caption{Performance of normal and pruned DNNs for different initial topologies.}
\label{results_Fashioi_MNIST}
\end{table}

\begin{figure}[h!]
\centering
\begin{subfigure}[b]{0.3\textwidth}
\includegraphics[width=\textwidth]{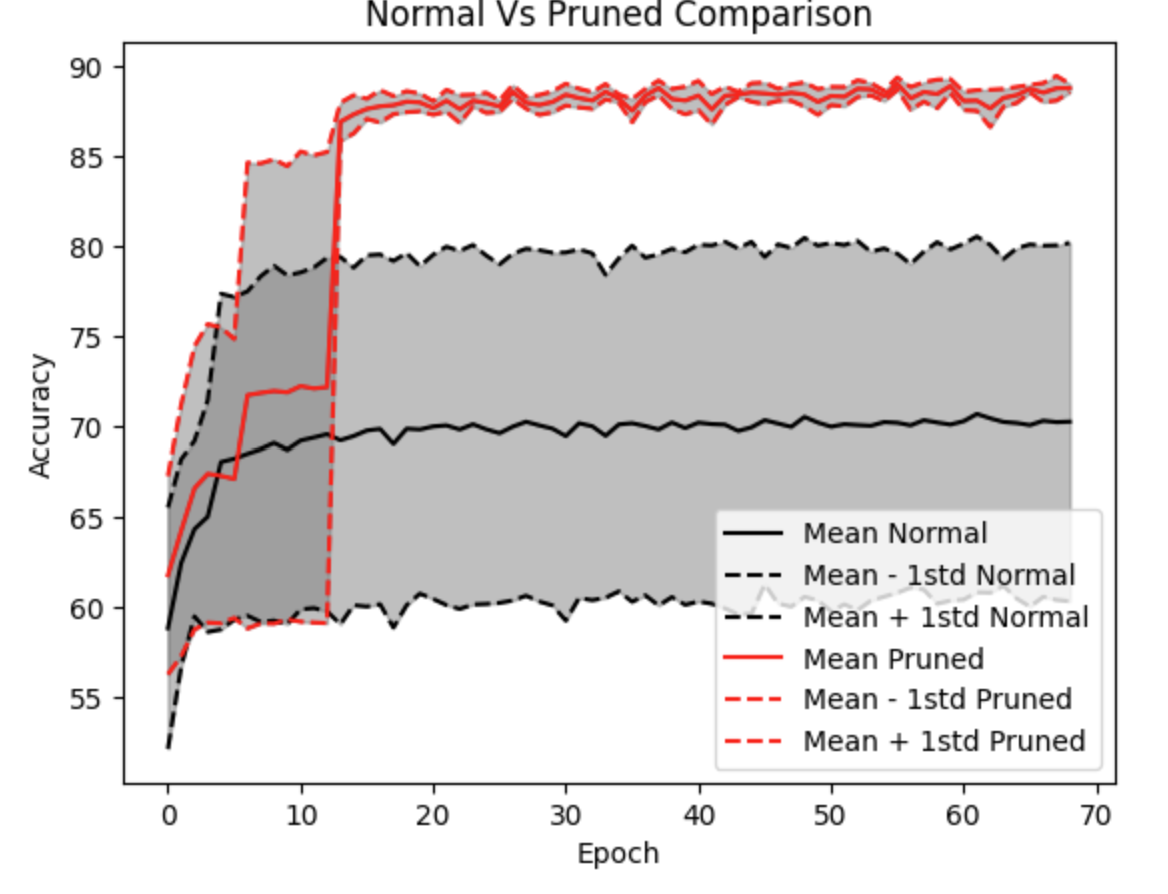}
\caption{}
\end{subfigure}
\begin{subfigure}[b]{0.337\textwidth}
\includegraphics[width=\textwidth]{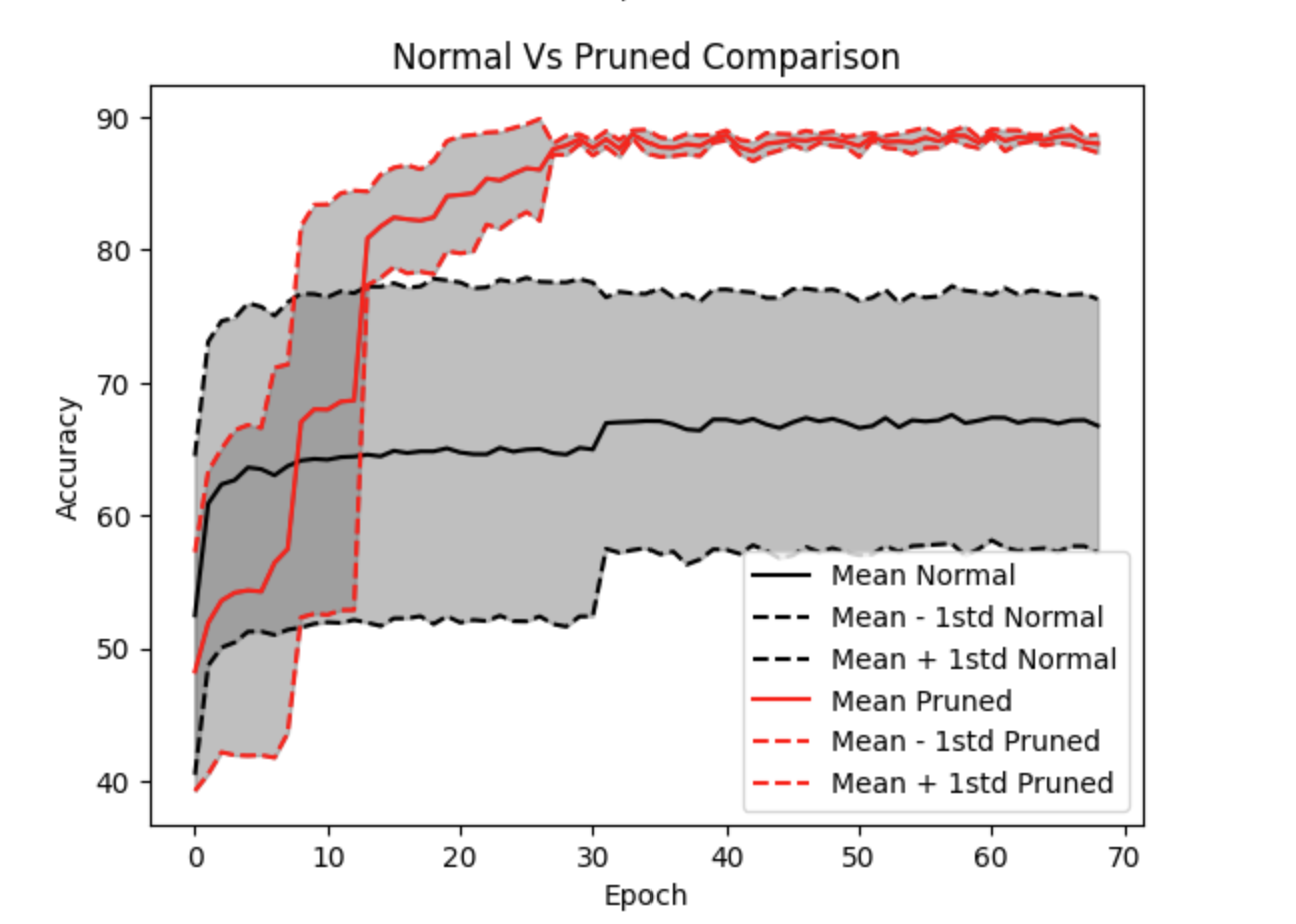}
\caption{}
\end{subfigure}
\begin{subfigure}[b]{0.3\textwidth}
\includegraphics[width=\textwidth]{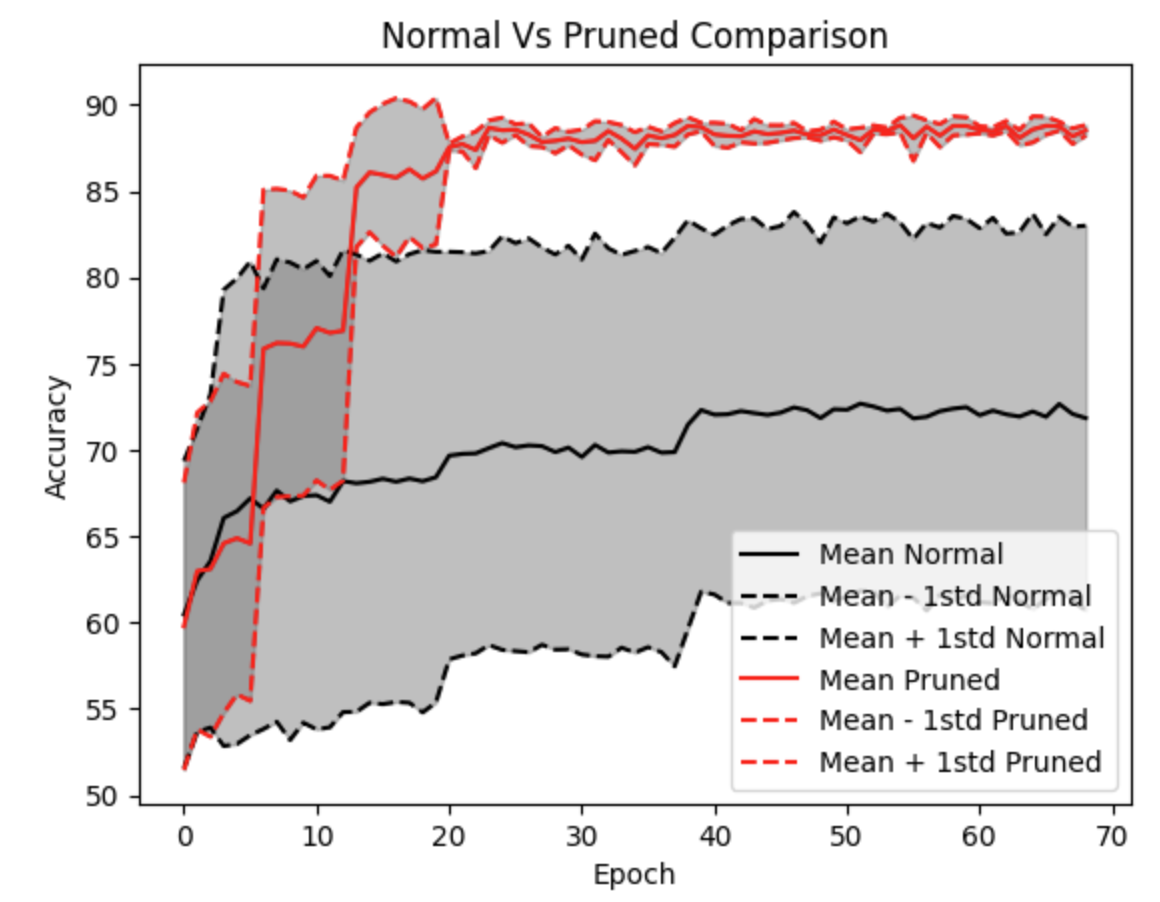}
\caption{}
\label{fig_3}
\end{subfigure}

\caption{Comparison of Normal DNN, trained normally, and pruned DNN, trained using the RMT approach on the test set. The sub-figures correspond to the different initial topologies: (a) $[784, 2000,4000,2000,500, 10]$, (b) $[784, 2000,2000,2000,2000,1000, 500, 10]$, (c) $[784, 3000,4000,3000,500, 10]$.}
\label{Comparison_fashion_MNIST}
\end{figure}

\label{reduction_paramters}

In Table \ref{paramters_Fashion_MNIST}, we observe the effect of our RMT training approach on the number of parameters in our DNN with different topologies. Each topology started with a fixed number of parameters, and by the end of training, we see a significant reduction in the number of parameters across all topologies for the pruned DNN. For each topology and across all seeds, the reduction in the number of parameters was consistent, indicating the robustness of our training process in pruning the network while maintaining performance. 

\begin{table}[h]
\centering
\footnotesize 
\begin{tabular}{|c|p{3.5cm}|p{3.5cm}|c|}
\hline
\textbf{Topology} & \textbf{Initial Parameters} & \textbf{Final Parameters} & \textbf{\% Reduction} \\ 
\hline
$[784, 2000, 4000, 2000, 500, 10]$ & 18,581,510 & 10,471,510 & 43.65\% \\
$[784, 2000, 2000, 2000, 2000, 1000, 500, 10]$ & 16,082,510 & 8,950,860 & 44.34\% \\
$[784, 3000, 4000, 3000, 500, 10]$ & 27,867,510 & 15,599,740 & 44.02\% \\
\hline
\end{tabular}
\caption{DNN topology, initial and final parameters for the pruned DNNs, with percentage reductions.}
\label{paramters_Fashion_MNIST}
\end{table}

\end{ex}

\begin{remark}
Again, we see that the RMT approach helps simplify the loss landscape so that during gradient descent the pruned DNN finds a deeper global minimum than the normal DNN. We can see this by looking at the accuracy of the DNNs on the training set; see Fig. \ref{training_set_example_fashion_MNIST}.

\begin{figure}[h!]
\centering
\begin{subfigure}[b]{0.3\textwidth}
\includegraphics[width=\textwidth]{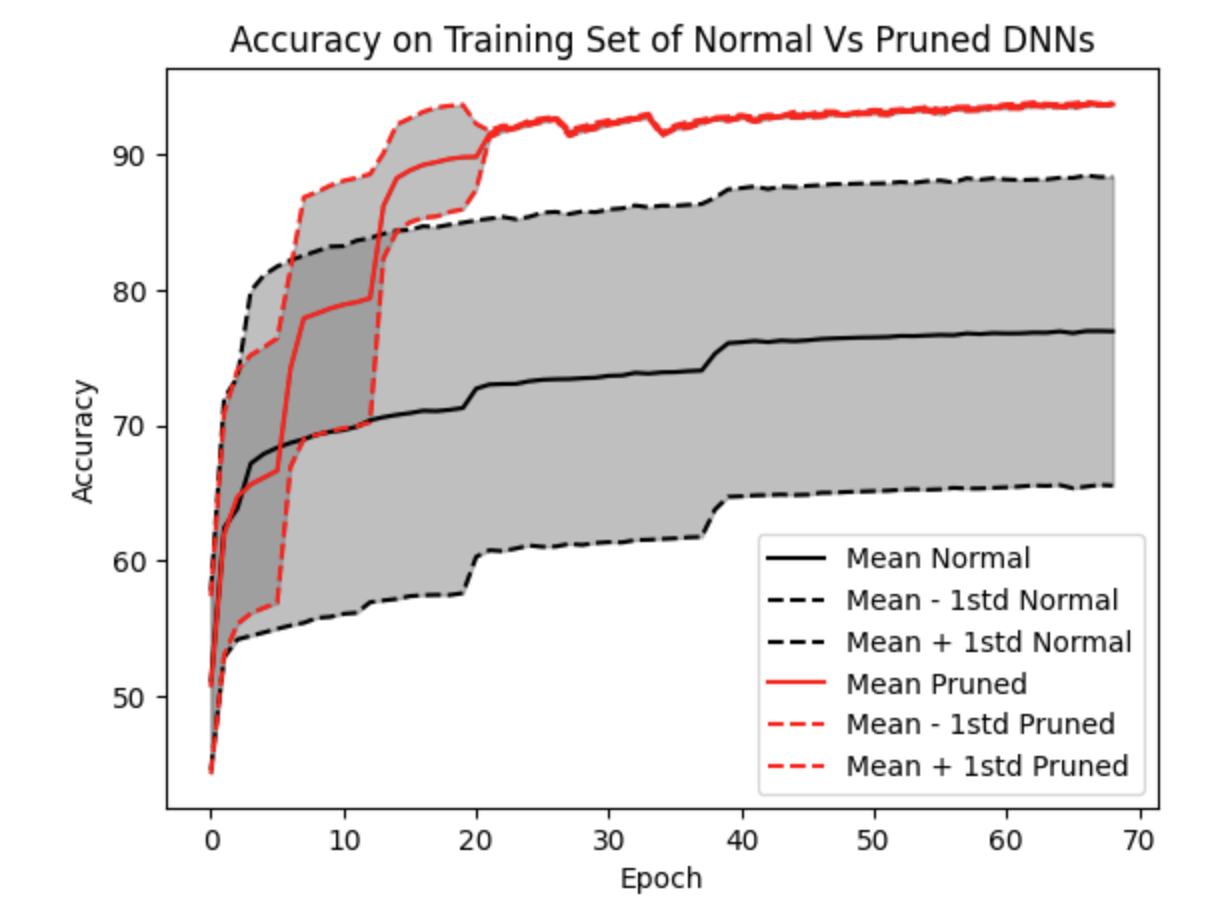}
\caption{}
\end{subfigure}
\begin{subfigure}[b]{0.3\textwidth}
\includegraphics[width=\textwidth]{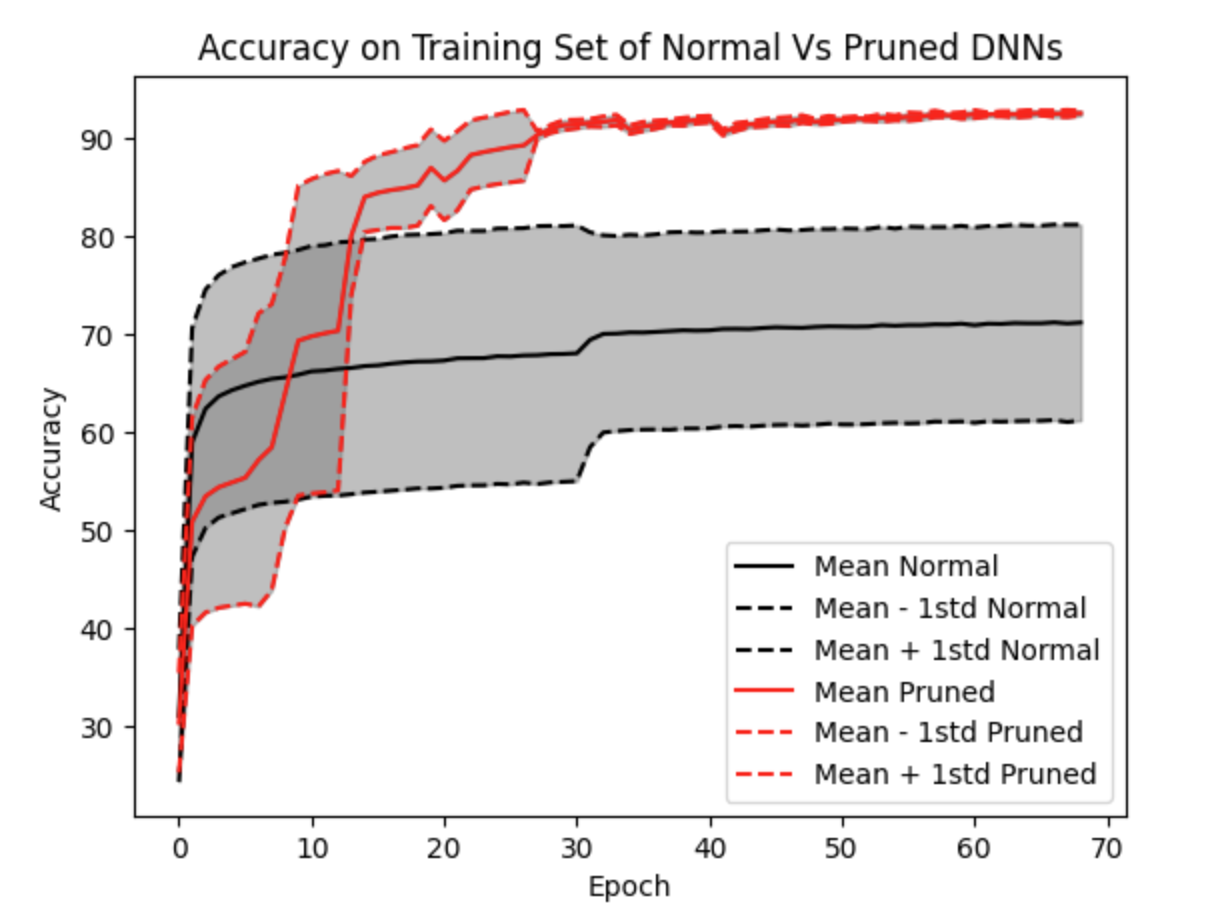}
\caption{}
\label{fig:ex2}
\end{subfigure}
\caption{Comparison of Normal DNN, trained normally, and pruned DNN, trained using the RMT approach on the training set. The sub-figures correspond to the different initial topologies: (a) $[784, 3000,4000,3000,500, 10]$, (b) $[784, 2000,2000,2000,2000,1000, 500, 10]$.}
\label{training_set_example_fashion_MNIST}
\end{figure}

One can also see that the pruned DNN is obtaining a deeper global minimum by looking at Table \ref{tab:loss_example_fashion_MNIST}. 

\begin{table}[h!]
\centering
\begin{tabular}{|c|c|c|c|c|}
\hline
\textbf{Topology} & \textbf{Type} & \textbf{Training Loss} & \textbf{Test Loss} \\ 
\hline
$[784, 2000,4000,2000,500, 10]$ & Normal & 0.460549 & 95.516107 \\
 & Pruned & 0.144334 & 47.063285 \\
\hline
$[784, 3000,4000,3000,500, 10]$ & Normal & 0.880157 & 130.785274 \\
 & Pruned & 0.191681 & 49.781472\\
\hline
$[784, 2000,2000,2000,2000,1000, 500, 10]$ & Normal & 0.790270 & 127.023410 \\
 & Pruned & 0.270197 & 42.925396 \\
\hline

\end{tabular}
\caption{Comparison of training and test losses between normal and modified DNNs for different topologies.}
\label{tab:loss_example_fashion_MNIST}
\end{table}

We present two graphs to analyze the impact of pruning on the performance of the DNN with architecture $[784, 2000, 4000, 2000, 500, 10]$, see Fig. \ref{fig:pruning_does}. Figure \ref{fig:accuracy_vs_epoch} shows the training and testing accuracy of the DNN over epochs. The blue dashed lines indicate the points at which pruning was applied. It is observed that the accuracy does not change significantly after pruning, especially during the initial epochs when the DNN parameters are still random. Figure \ref{fig:pruning_impact} specifically examines the impact of pruning on training accuracy. Red dots represent the accuracy before pruning, and purple dots represent the accuracy after pruning. The graph demonstrates that the training accuracy remains relatively stable before and after pruning, reinforcing the observation that MP-based pruning does not drastically affect performance, particularly in the early stages of training. After the pruning, the training seems to be easier, and the DNN accuracy improves as the training continues.

\begin{figure}[h]
    \centering
    \begin{subfigure}[b]{0.7\textwidth}
        \includegraphics[width=\textwidth]{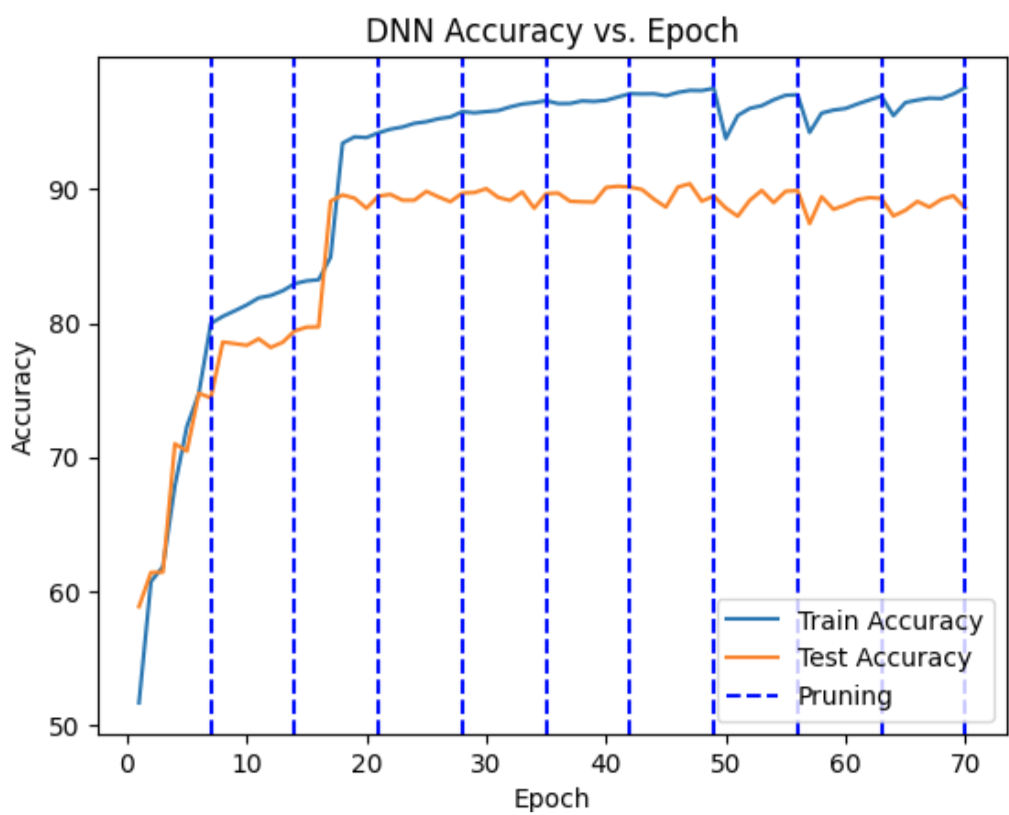}
        \caption{Accuracy vs. Epoch}
        \label{fig:accuracy_vs_epoch}
    \end{subfigure}
    \begin{subfigure}[b]{0.7\textwidth}
        \includegraphics[width=\textwidth]{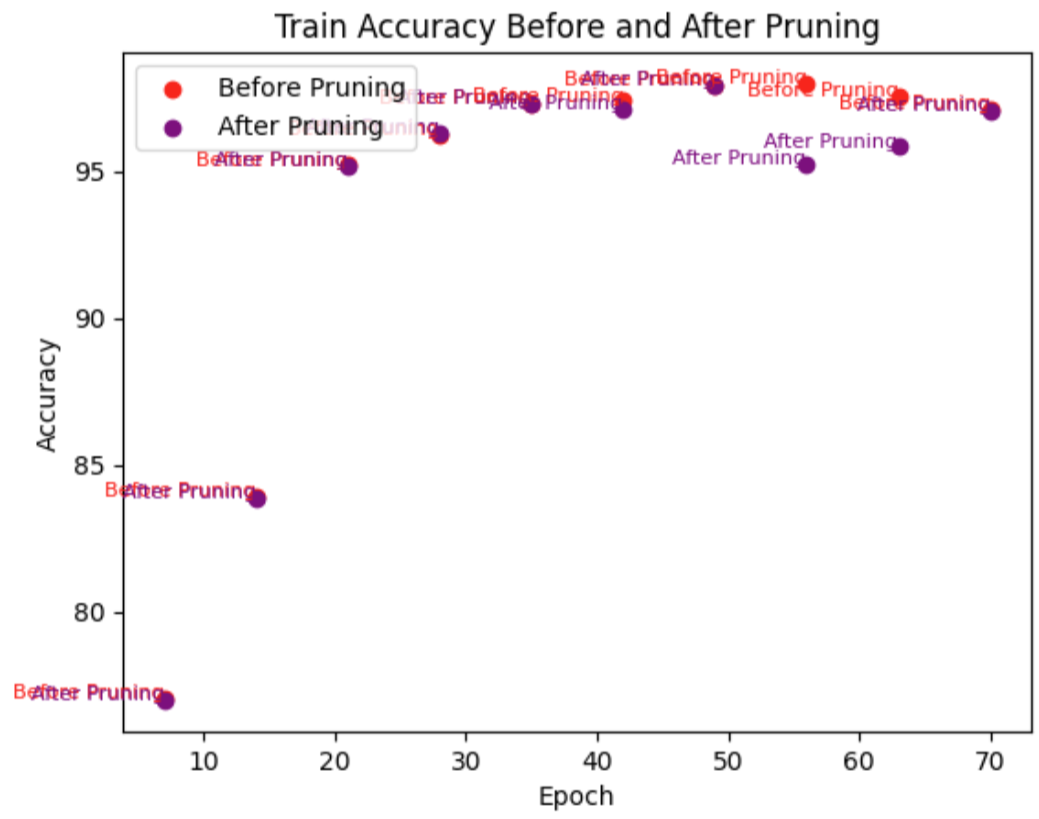}
        \caption{Train Accuracy Before and After Pruning}
        \label{fig:pruning_impact}
    \end{subfigure}
    \caption{Analysis of DNN Training and Pruning}
    \label{fig:pruning_does}
\end{figure}

\end{remark}

We applied the MP-based pruning approach for DNNs with other initializations, such as the He and Xavier initializations, and obtained similar results in improvements of accuracy, see Subsection \ref{Other_initilization}. 

Also, for the DNN with architecture $[784, 2000,4000,2000,500, 10]$, we computed the cost of training the DNN for $70$ epochs on the Intel Xeon CPU with 2 vCPUs (virtual CPUs) and 13GB of RAM chip, with MP-based pruning and a split frequency of $7$. The total cost was $12,215.13$ seconds, and we achieved an accuracy of $\sim 89.7 \%$. Finally, we trained the same DNN for the same amount of CPU time but without MP-based pruning, and the DNN accuracy plateaued at $\sim 84\%$ accuracy. This illustrates that the increase in accuracy provided by MP-based pruning is not obtained because of increases in computational costs alone.   

We applied the MP-based pruning approach on a DNN without adding any regularization while training with GD alone and with a fixed $lr=.01$. The DNN architecture was  $[784, 3000, 3000, 3000,3000, 500, 10]$ and we trained for $70$ epochs with a split frequency of $7$. Without pruning, the accuracy plateaued at $44\%$, while with pruning, it reached $\sim 80\%$. This indicates that other strategies, such as regularization or rate decay, are useful to take advantage of MP-based pruning. However, MP-based pruning improves the accuracy of GD alone.

We trained a DNN with architecture $[784, 3000, 3000, 3000, 500, 10]$ on Fashion MNIST for $300$ epoch, while adding both $L1$ and $L2$ regularization to the loss, see \eqref{loss_noise_det}. The hyperparameters for the regularization were $.0000005$ and $.0000001$, respectively; the split frequency was $13$ and the number of singular values kept was given by $f(\text{{epoch}}) = \max\left(0, -\frac{1}{1000} \cdot \text{{epoch}} + 1\right)$. All other hyperparameters were kept the same as in Example \ref{Comparison_fashion_MNIST}. The DNN achieved a $100\%$ accuracy on the training set and a $91.27\%$ accuracy on the test set, showing that the MP-based pruning algorithm attains higher accuracy when we combine it with regularization. More on the relationship between regularization and MP-based pruning will be discussed in another paper.

\paragraph{MP-based pruning with sparsification for fully connected DNNs on Fashion MNIST}
\label{MP+sp_1}

\begin{figure}
		\centering	\includegraphics[scale=1.05]{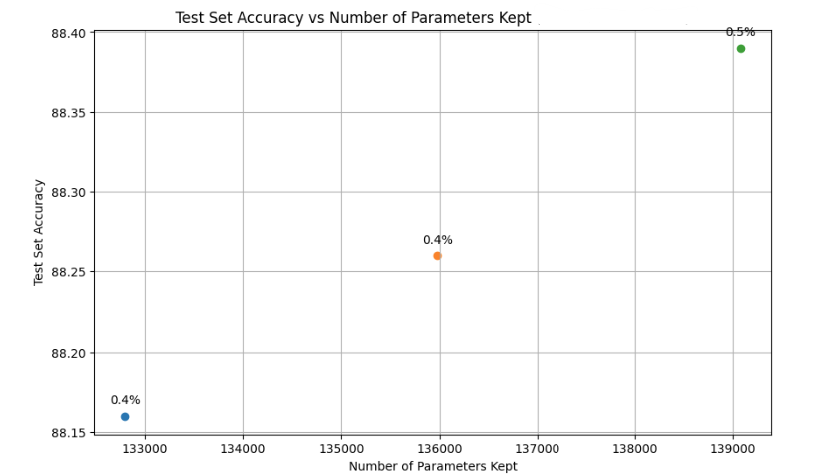}
			\caption{Accuracy vs. number of parameters kept. The percentage of parameters kept is also shown (above each point on the graph).}
			\label{full_fash_MNIST_with_sp}
		\end{figure}

We train a fully connected DNN on Fashion MNIST to achieve $\sim 89\%$ accuracy (on the test set) with the same MP-based pruning approach as in Subsection \ref{full_Fash_MNIST} for a DNN with the topology  [784,  3000, 3000, 3000, 3000, 500, 10]. At the end of the training, we employ the sparsification method by setting to zero weights in the DNN smaller than the \textit{sparsification threshold} $\xi$. Fig. \ref{full_fash_MNIST_with_sp} shows the accuracy of the DNN vs. the number of parameters kept (determined by varying $\xi$). This additional sparsification leads to a large reduction in parameters, by over 99.5\%, without a significant drop in accuracy ($\sim .5\%$ drop). Assuming that the weights of the DNN which are smaller than the threshold $\xi$ are i.i.d.s. from a distribution with zero mean and bounded variance, Lemma \ref{main_result_remove _R} provides an explanation for why removing the small weights (sparsification) does not affect accuracy.

Alternatively, one can prune the weight layers during training by combining the MP-based pruning approach together with sparsification (i.e., removing weights smaller than the threshold $\xi$ every couple of epochs). We performed this training on the above DNN, with $\xi$ depending on the epoch. In our case, we initially took $\xi=0.001$ and set it to grow linearly so by the end of training $\xi= 0.02$. We achieved a $88\%$ accuracy, while the final DNN had $71,331$ parameters (keeping $\sim .2\%$ parameters).  

Finally, we tried the sparsification pruning method during training without MP-based pruning. Again, we initially took $\xi=0.001$ and set it to grow linearly so that by the end of training $\xi= 0.02$. Similar to our observations from subsection \ref{full_Fash_MNIST}, the DNN plateaued at $\sim 70\%$ accuracy while having $128,533$ parameters at the end of training (keeping $\sim .4\%$ parameters). Thus, we see that a combination of MP-based pruning with sparsification is useful for pruning while also increasing DNN accuracy for fully connected DNNs trained on Fashion MNIST.      

\subsubsection{MP-based pruning of CNNs on  MNIST and Fashion MNIST}
\label{MNIST_Fashion_Con_Red}

In our further exploration, we perform numerical simulations on Convolutional Neural Networks (CNNs)  using MNIST and Fashion MNIST. In this simulation, our primary objective is to investigate the effect of pruning the small singular values of the convolutional layers. The overall goal in this example is to reduce the number of parameters in the CNN while at the same time preserving its accuracy. 

Given the multidimensional nature of convolutional layers, the direct application of singular value decomposition is not straightforward. To overcome this challenge, we first transform each convolutional layer into a 2-dimensional matrix. Specifically, for a convolutional layer with dimensions $m \times n \times p \times q$ (where $m$ is the number of output channels, $n$ is the number of input channels, and $p \times q$ is the kernel size), we reshape it into a matrix of size $m \times npq$. 

After this flattening process, we proceed with the pruning as before, employing SVD to remove the smaller singular values. This step essentially compresses the convolutional layer, reducing its complexity while hopefully maintaining its representational capability.

The hyperparameters for the simulations in the next example can be found in Subsection \ref{hyperparam_3}. The other parts of the CNN architecture (which is the same for all of the CNNs in this paper) can be found in Subsection \ref{CNN_arc}. 

The learning rate (lr) is also modified every epoch to be:

\begin{equation}
    lr_n=lr_{n-1}*.96
\end{equation}
where $lr_k$ is the learning rate at epoch $k$. Thus it decays over the learning time, see \cite{goodfellow2016deep} for more information.

\begin{ex}
\label{Example_CNN_F_MNIST}
In this first example, we trained a CNN on MNIST for 30 epochs with a split frequency of 13. The convolutional layers are given by  $[1, 64, 128, 256, 512]$, and the model starts with one input channel, and then each subsequent number represents the number of filters in each subsequent convolutional layer. Therefore, the model has 4 convolutional layers with filter sizes of 64, 128, 256, and 512, respectively. We apply a kernel for each layer of size $3 \times 3$.

The fully connected layers are given by $[41472, 20000, 10000, 5000, 3000, 1400, 10]$, we see that the model has 6 fully connected layers. The GoF parameter for the fully connected layers is $.6$ while the GoF parameter for the convolutional layers is $.05$. 

In these numerical simulations, the portion of singular values smaller than $\sqrt{\lambda_{+}}$ that we retain is given by the linear function:

\begin{equation}
    f(\text{{epoch}}) = \max\left(0, -\frac{1}{20} \cdot \text{{epoch}} + 1\right)
\end{equation}

The accuracy of this DNN on the training and test set is given in Fig. \ref{training_test_set_example_MNIST_con}.

\begin{figure}[h!]
\centering
\begin{subfigure}[b]{0.45\textwidth}
\includegraphics[width=\textwidth]{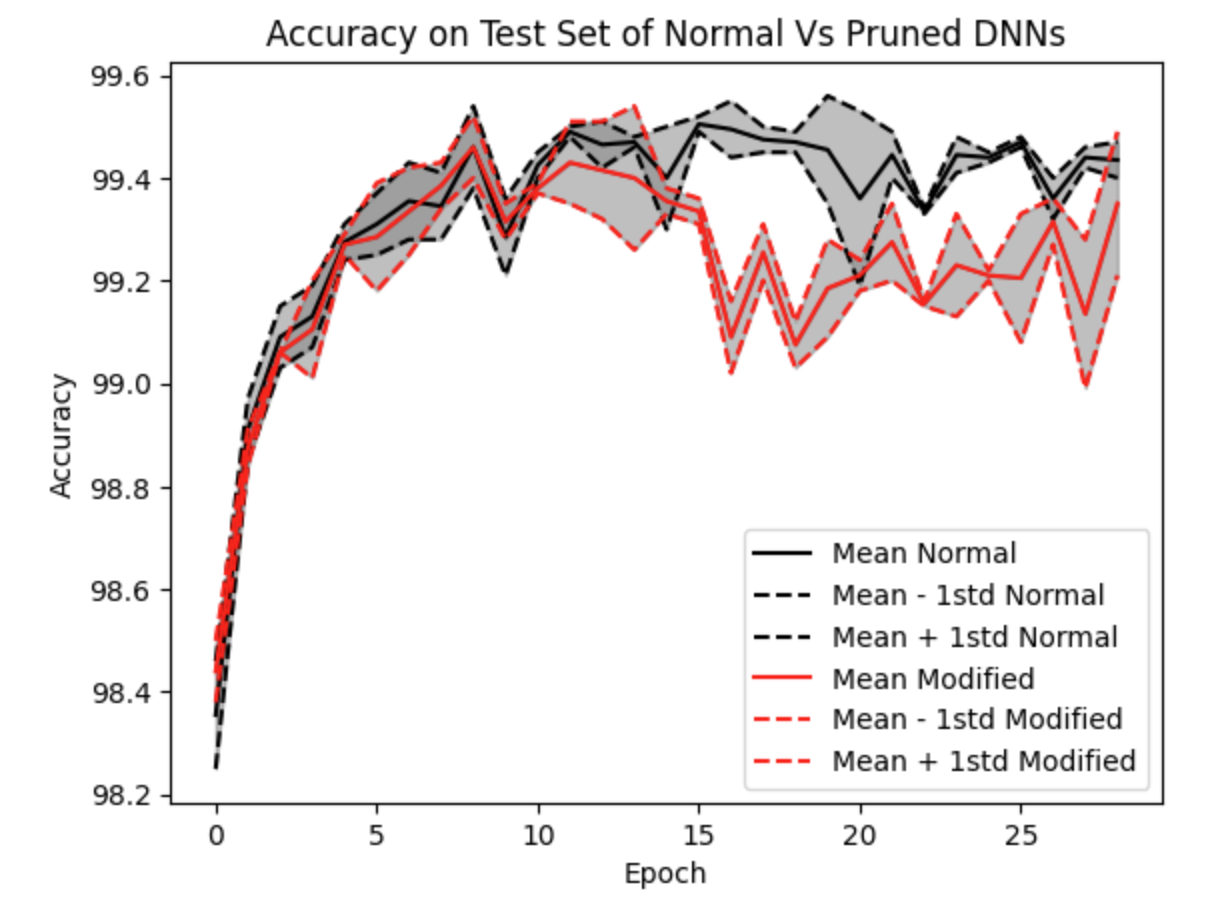}
\caption{}
\label{fig:ex1_con_MNIST_test}
\end{subfigure}
\begin{subfigure}[b]{0.475\textwidth}
\includegraphics[width=\textwidth]{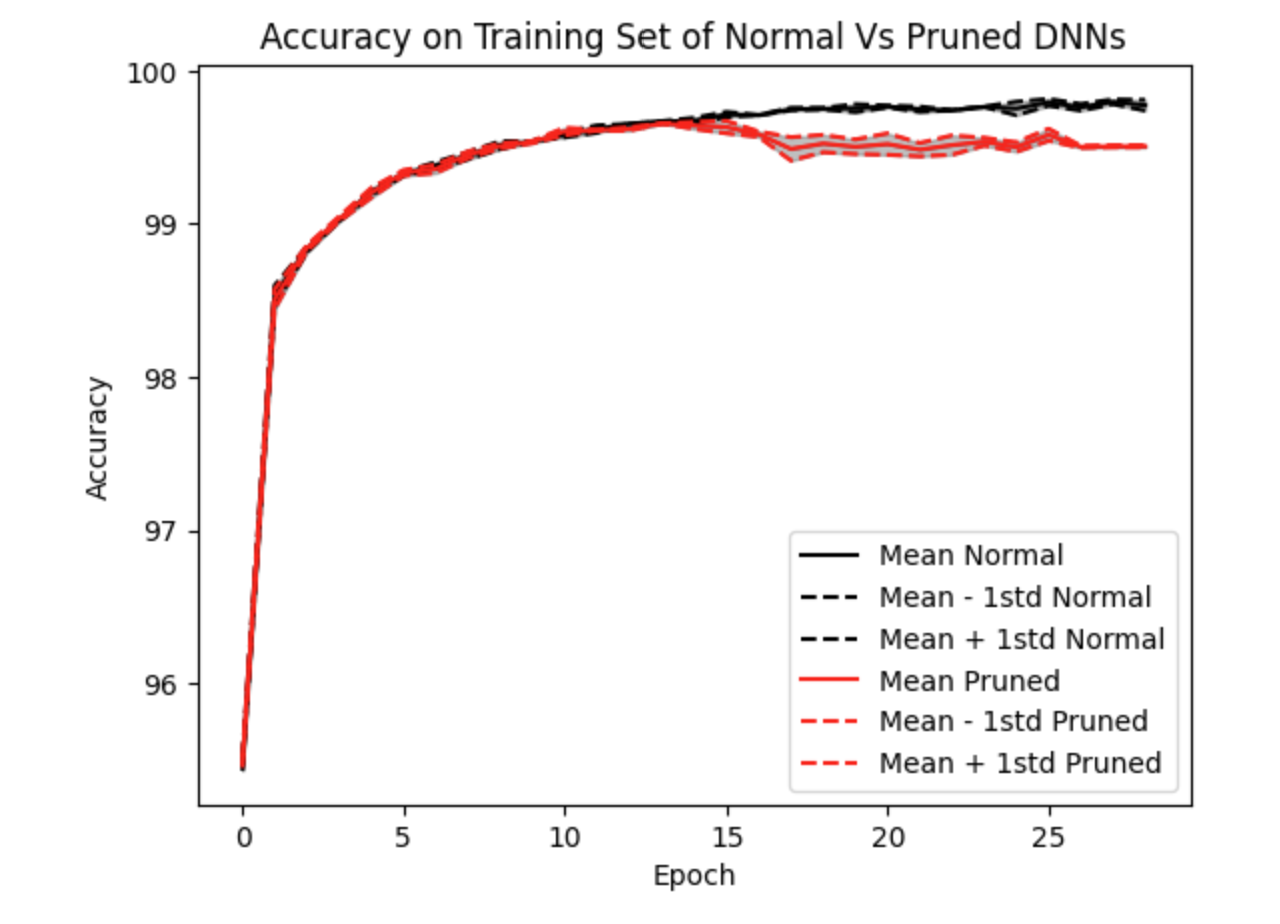}
\caption{}
\end{subfigure}
\caption{Comparison of Normal DNN, trained normally, and pruned DNN, trained using the RMT approach on the test and training sets.}
\label{training_test_set_example_MNIST_con}
\end{figure}

\end{ex}

In Example \ref{Example_CNN_F_MNIST}, it was observed that the pruned  CNN exhibited a slightly lower accuracy in comparison to the normally trained CNN. Remarkably, despite this marginal drop in performance, the pruned CNN managed to maintain this level of accuracy with approximately half of the parameters used by the normally trained CNN. While the normally trained CNN has $1,100,323,974$ parameters (on account of how large the fully connected layers are), the pruned CNN has $583,670,449$ parameters. 

In terms of performance on the training set, the normally trained CNN demonstrated an accuracy of 100\%, an indicator of its potential overfitting to the training data. This is in contrast with the pruned CNN, which displayed a lower accuracy on the training set. The narrower gap between the training set and test set accuracies for the pruned CNN could be interpreted as a sign of reduced variance between the training and test set, suggesting less overfitting in the pruned model. At the same time, the fact that the normal CNN archives have high accuracy on the training set suggests that the loss function in this example is simple- i.e., finding the global max of the loss function is simple. This might be why the pruned CNN does not outperform the unpruned CNN.

In this example, the drop in test set accuracy for the pruned CNN is dependent on the Goodness-of-Fit (GoF) parameter. As the GoF parameter becomes more restrictive, the drop in accuracy becomes less pronounced. However, it is important to note that a more restrictive GoF parameter also leads to a smaller reduction in parameters. These observations suggest a delicate balance between the GoF parameter, model complexity (as indicated by the number of parameters), and model performance.

\begin{ex}
\label{CNN_F_MNIST_example_1}

In this example, we trained a CNN on the fashion MNIST dataset for 70 epochs with a split frequency of 17. The convolutional layers are given by  $[1, 64, 128, 256, 512]$; we again apply a kernel for each layer of size $3 \times 3$. The fully connected layers are given by $[41472, 10000, 5000, 5000,   10]$. The GoF parameter for the fully connected layers is $.7$, while the GoF parameter for the convolutional layers is $.15$. 

In this numerical simulation, the portion of singular values smaller than $\sqrt{\lambda_{+}}$ that we retain is given by the linear function:

\begin{equation}
    f(\text{{epoch}}) = \max\left(0, -\frac{1}{60} \cdot \text{{epoch}} + 1\right)
\end{equation}

The accuracy of this DNN on the training and test set is given in Fig. \ref{training_test_set_example_fashion_MNIST_con}.

\begin{figure}[h!]
\centering
\begin{subfigure}[b]{0.45\textwidth}
\includegraphics[width=\textwidth]{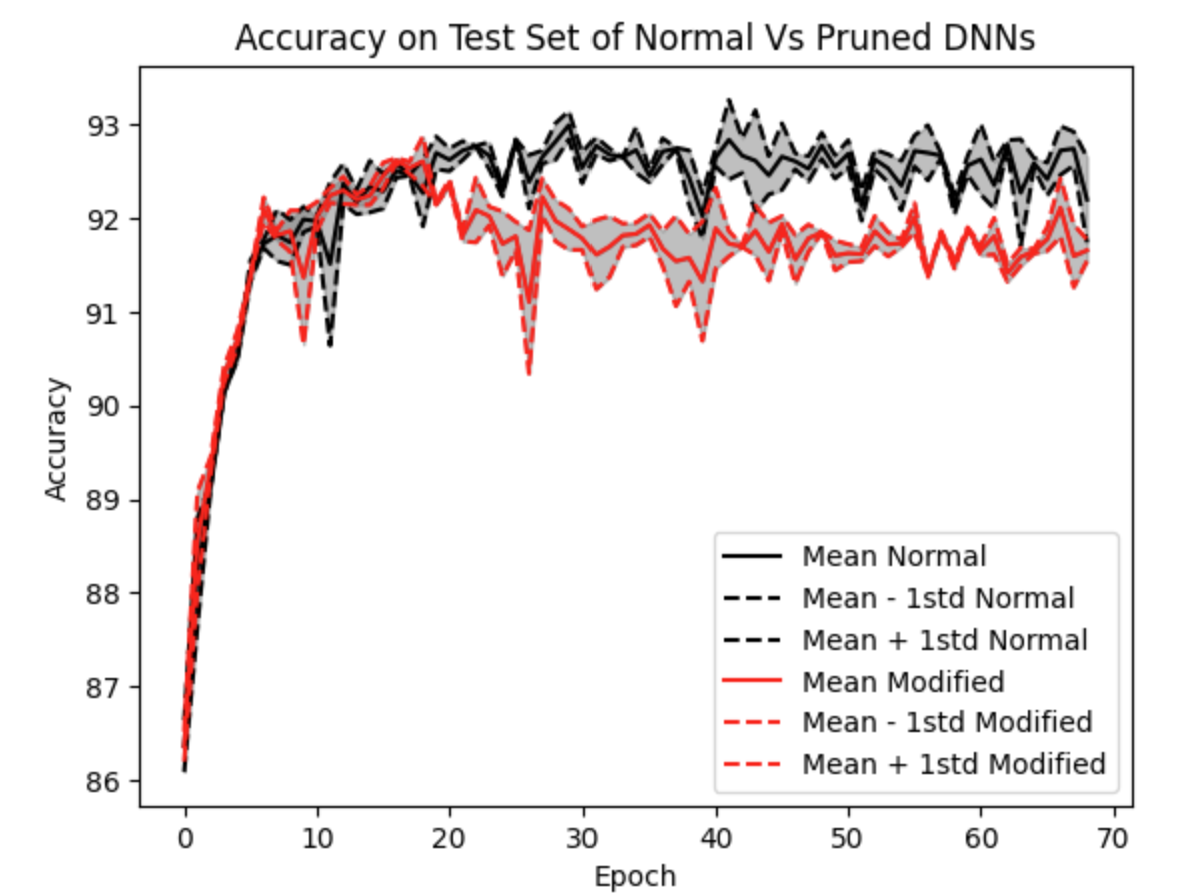}
\caption{}
\label{fig:ex1_con_fashion_MNIST_test}
\end{subfigure}
\begin{subfigure}[b]{0.465\textwidth}
\includegraphics[width=\textwidth]{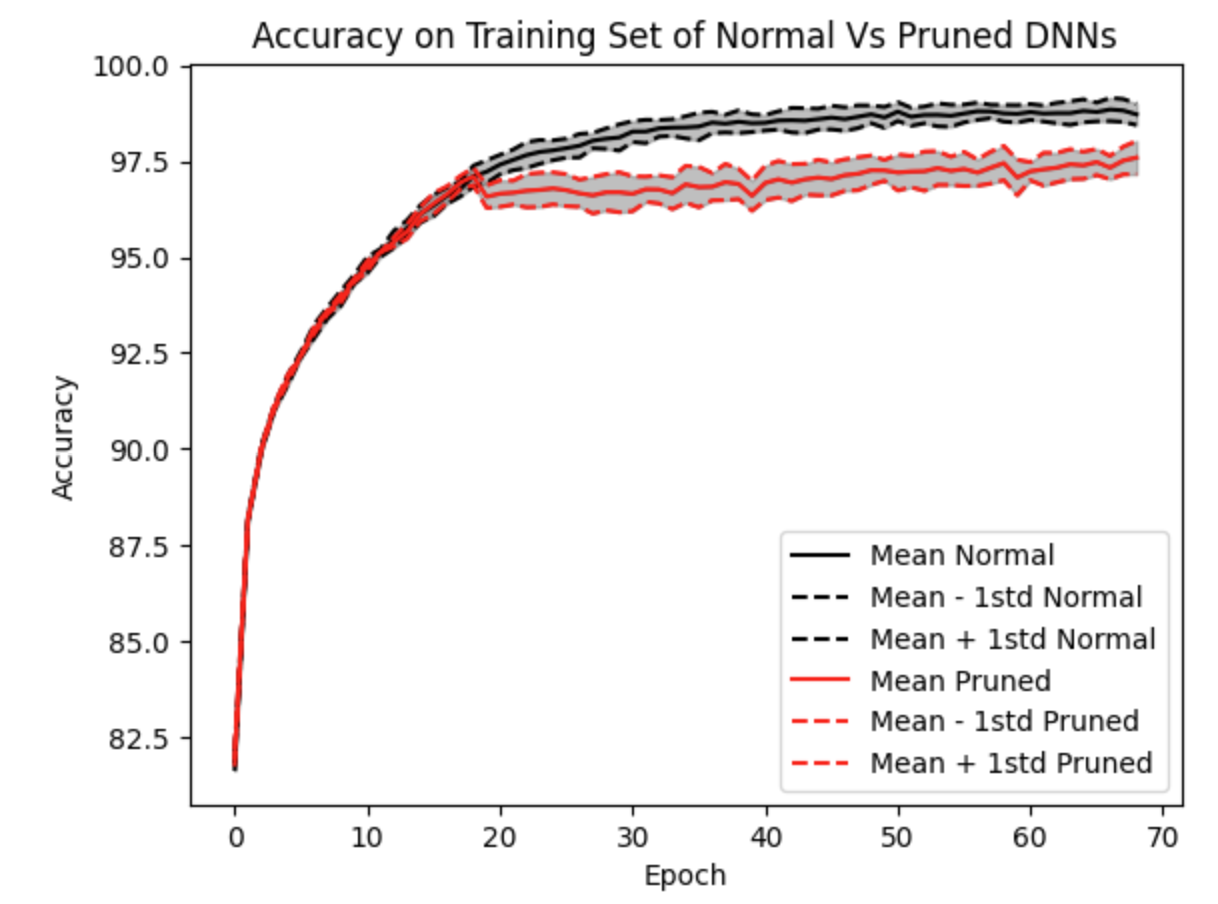}
\caption{}
\label{fig:ex1_con_fashion_MNIST_train}
\end{subfigure}
\caption{Comparison of Normal DNN, trained normally, and pruned DNN, trained using the RMT approach on the Fashion MNIST test and training sets.}
\label{training_test_set_example_fashion_MNIST_con}
\end{figure}

\end{ex}

In Example \ref{CNN_F_MNIST_example_1}, the pruned CNN exhibited a slightly lower accuracy in comparison to the conventionally trained CNN. Despite this marginal drop in performance, the pruned CNN maintained this level of accuracy with approximately half of the parameters used by the conventionally trained CNN. While the normally trained CNN had $491,381,774$ parameters, the pruned CNN utilized only $261,891,332$ parameters. 

In terms of performance on the training set, the conventionally trained DNN demonstrated an accuracy of approximately 99\%, suggesting potential overfitting to the training data. On the other hand, the pruned DNN displayed a lower accuracy on the training set. The smaller variance between the training and test set accuracies for the pruned DNN could again be interpreted as a sign of less overfitting. 

\paragraph{MP-based pruning with sparsification for CNN trained on Fashion MNIST}
\label{MP+sp_2}

We train the CNN found in Example \ref{CNN_F_MNIST_example_1} to achieve $\sim 92\%$ accuracy (on the Fashion MNIST test set) with the same MP-based pruning approach as in Subsection \ref{MNIST_Fashion_Con_Red}. At the end of the training, we employ sparsification by setting to zero all weights in the DNN smaller than some threshold $\xi$. 

\begin{figure}[ht]
    \centering
    \begin{subfigure}[b]{1.1\textwidth}
        \centering        \includegraphics[scale=1]{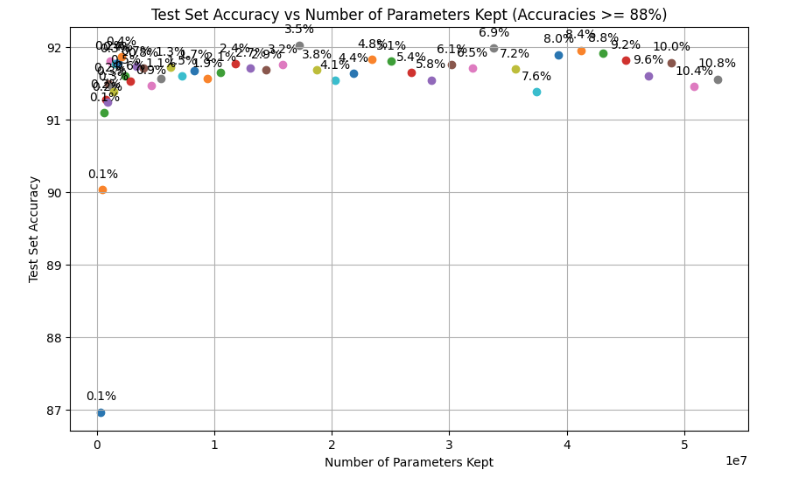}
        \caption{Sparsification with MP-based pruning. The percentage of parameters kept is shown above each point on the graph.}
        \label{fig:with_mp_pruning}
    \end{subfigure}
    \hfill
    \begin{subfigure}[b]{1\textwidth}
        \centering
        \includegraphics[scale=1.1]{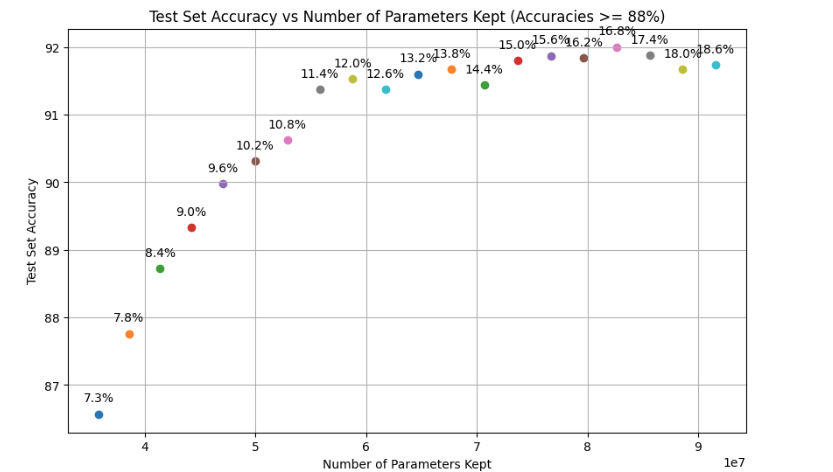}
        \caption{Sparsification without MP-based pruning. The percentage of parameters kept is shown above each point on the graph.}
        \label{fig:without_mp_pruning}
    \end{subfigure}
    \caption{Accuracy vs. number of parameters kept for CNNs trained on Fashion MNIST with and without MP-based pruning.}
    \label{fig:mp_pruning_comparison}
\end{figure}

As shown in Fig. \ref{fig:with_mp_pruning}, this additional sparsification leads to a large reduction in parameters, by over 99.5\%, without a significant drop in accuracy ($\sim .1\%$ drop). As mentioned, Lemma \ref{main_result_remove _R} provides an explanation for why removing the small weights does not affect accuracy, as these weights appear to correspond to the noise in the weight layers, and removing them should not change accuracy.

Finally, we applied the sparsification pruning method after training without MP-based pruning (during training). Fig. \ref{fig:without_mp_pruning} shows that the pruning threshold seems to affect the accuracy of the DNN in a much more significant manner. That is, even when pruning $95\%$ of the parameters, the accuracy drops by multiple percentage points. We see that a combination of MP-based pruning with sparsification is useful for pruning while also ensuring the DNN accuracy does not decrease much for CNNs trained on Fashion MNIST.

\subsubsection{Numerics for training DNNs on CIFAR-10: reducing parameters via MP-based pruning}
\label{CIFAR-10}

In this numerical simulation, we applied the RMT algorithm to prune a DNN trained on the CIFAR-10 dataset.  The CIFAR-10 dataset consists of 60,000 color images 
spanning 10 different classes. The dataset is split into a training set and a test set. 
The training set contains 50,000 images, while the test set comprises 10,000 images, which is standard.

Throughout the training process, we tracked the performance metrics of both the pruned and normally trained DNNs on both the test and training sets. Our analysis showed that the pruned network, despite having a reduced number of parameters, managed to achieve performance metrics comparable to those of the normally trained network. Additionally, the pruning process significantly reduced the number of parameters in the pruned DNN, resulting in a more efficient network with a lower computational footprint. The hyperparameters for the simulations in this subsection can be found in Subsection \ref{hyperparam_4}.

The lr is also modified every epoch to be:

\begin{equation}
    lr_n=lr_{n-1}*.96
\end{equation}

where $lr_k$ is the learning rate at epoch $k$.

\begin{ex}
\label{CIFAR10_example}

In this example, we trained a CNN on CIFAR10 for 350 epochs with a split frequency of 40. The convolutional layers are given by  $[3, 32,64, 128, 256, 512]$. We again apply a kernel for each layer of size $3 \times 3$.

The fully connected layers are given by $[8192, 500, 10]$. The GoF parameter for the fully connected layers is $.08$, while the GoF parameter for the convolutional layers is $.06$. 

In this simulation, the portion of singular values smaller than $\lambda_{+}$ that we retain is given by the linear function:

\begin{equation}
    f(\text{{epoch}}) = \max\left(0, -\frac{1}{200} \cdot \text{{epoch}} + 1\right)
\end{equation}

The accuracy of this DNN on the training and test set is given in Fig. \ref{training_test_set_example_CF10_con}.

\begin{figure}[h!]
\centering
\begin{subfigure}[b]{0.45\textwidth}
\includegraphics[width=\textwidth]{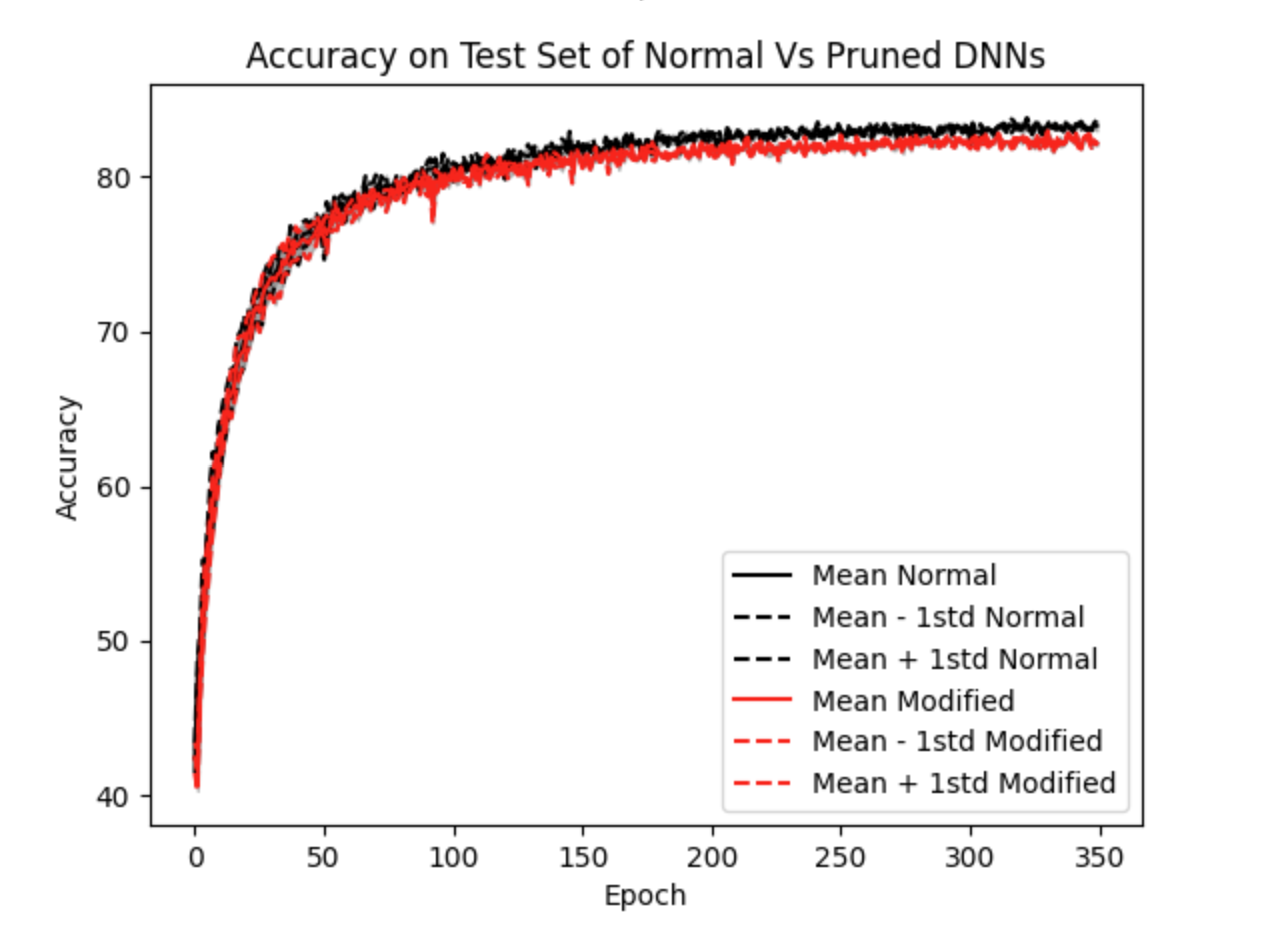}
\caption{}
\label{fig:ex1_con_CF10_test}
\end{subfigure}
\begin{subfigure}[b]{0.455\textwidth}
\includegraphics[width=\textwidth]{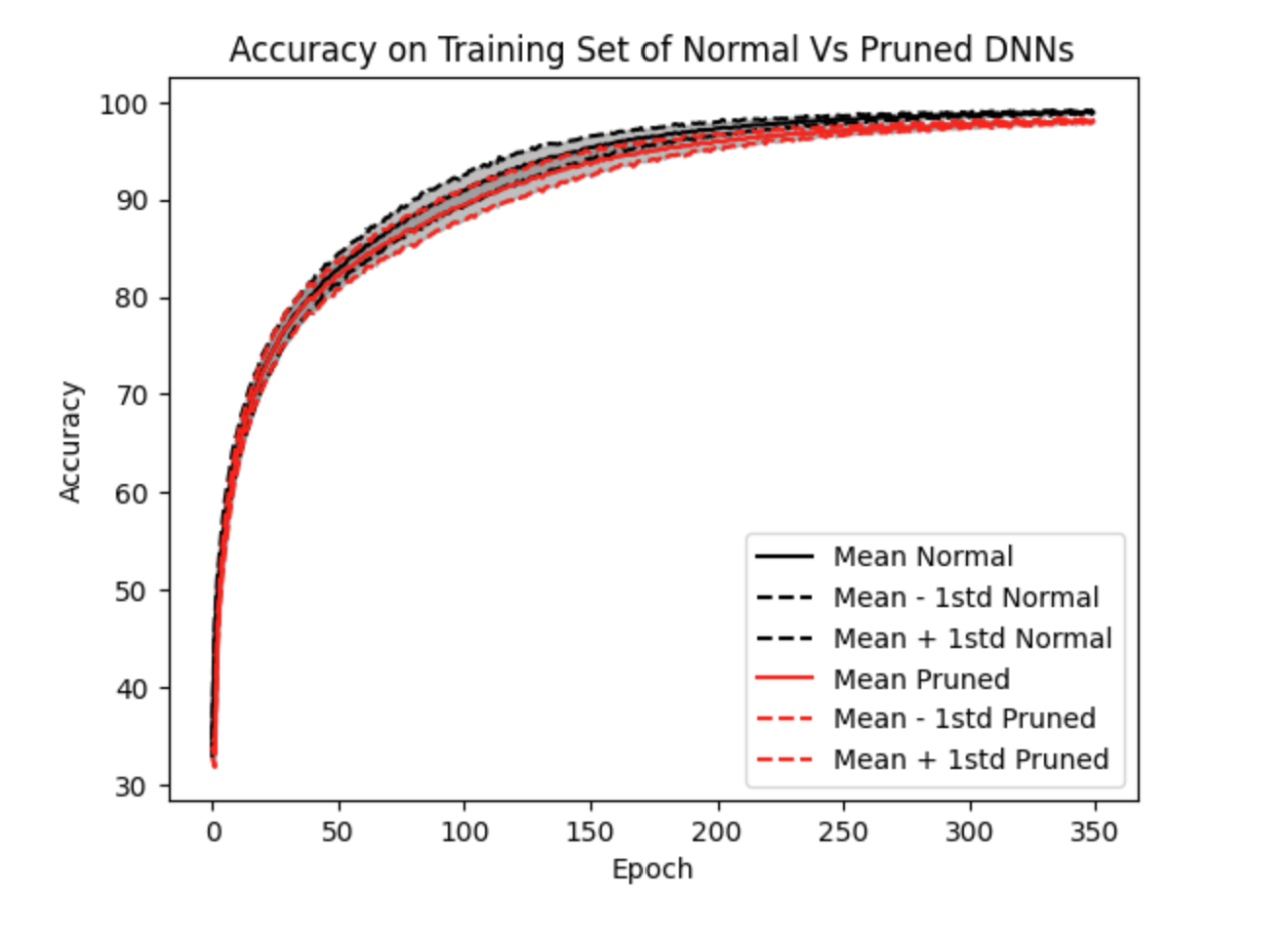}
\caption{}
\label{fig:ex1_con_CF10_train}
\end{subfigure}
\caption{Comparison of Normal DNN, trained normally, and pruned DNN, trained using the RMT approach on the test and training sets.}
\label{training_test_set_example_CF10_con}
\end{figure}

\end{ex} 

In Example \ref{CIFAR-10}, it was observed that the pruned DNN exhibited a slightly lower accuracy in comparison to the normally trained DNN. Again, despite this marginal drop in performance, the pruned DNN managed to maintain this level of accuracy with much fewer parameters than was used by the normally trained DNN. While the normally trained DNN has $5,673,090$ parameters, the pruned DNN has $3,949,078$ parameters.

\paragraph{MP-based pruning with sparsification for CNN trained on CIFAR10}
\label{MP+sp_3}

We train the CNN found in Example \ref{CIFAR10_example} on CIFAR10 to achieve $\sim 82\%$ accuracy (on the test set) with the same MP-based pruning approach as in Subsection \ref{CIFAR-10}. At the end of training, we sparsify the DNN by setting to zero weights in the DNN smaller than some threshold $\xi$. 

\begin{figure}[h!]
    \centering
    \begin{subfigure}[b]{0.6\textwidth}
        \centering        \includegraphics[scale=.7]{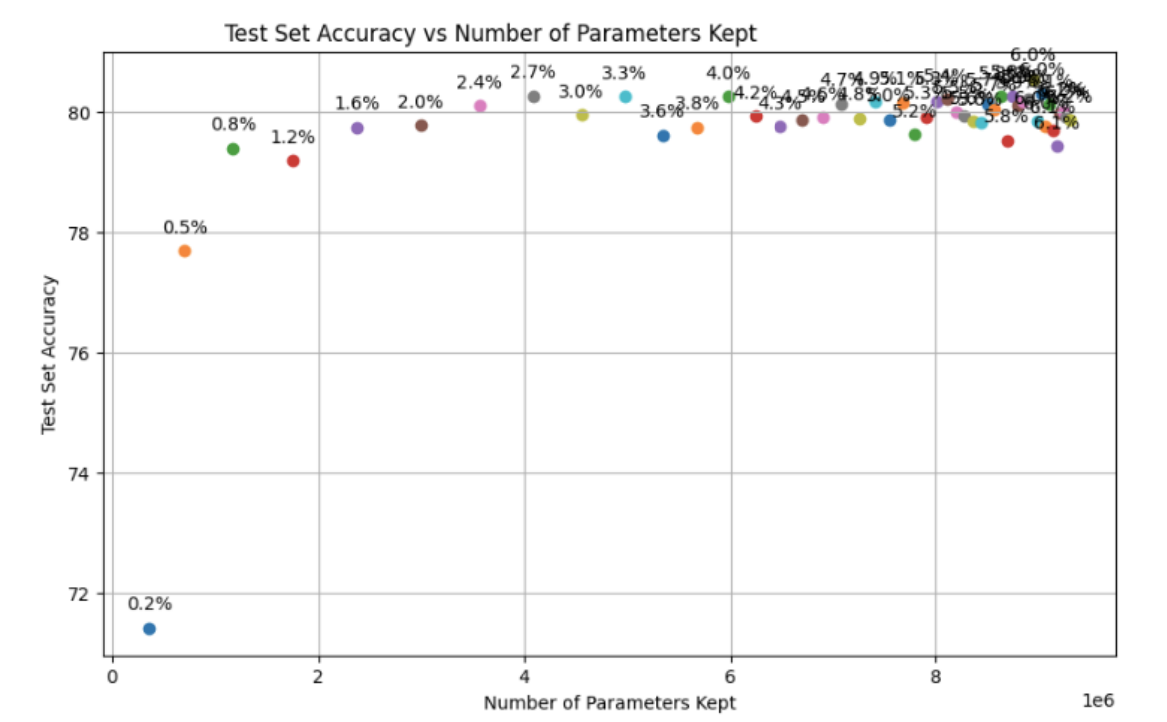}      \caption{Sparsification with MP-based pruning. The percentage of parameters kept is shown above each point on the graph.}
        \label{fig:with_mp_pruning_2}
    \end{subfigure}
    \hfill
    \begin{subfigure}[b]{1\textwidth}
        \centering
        \includegraphics[scale=0.7]{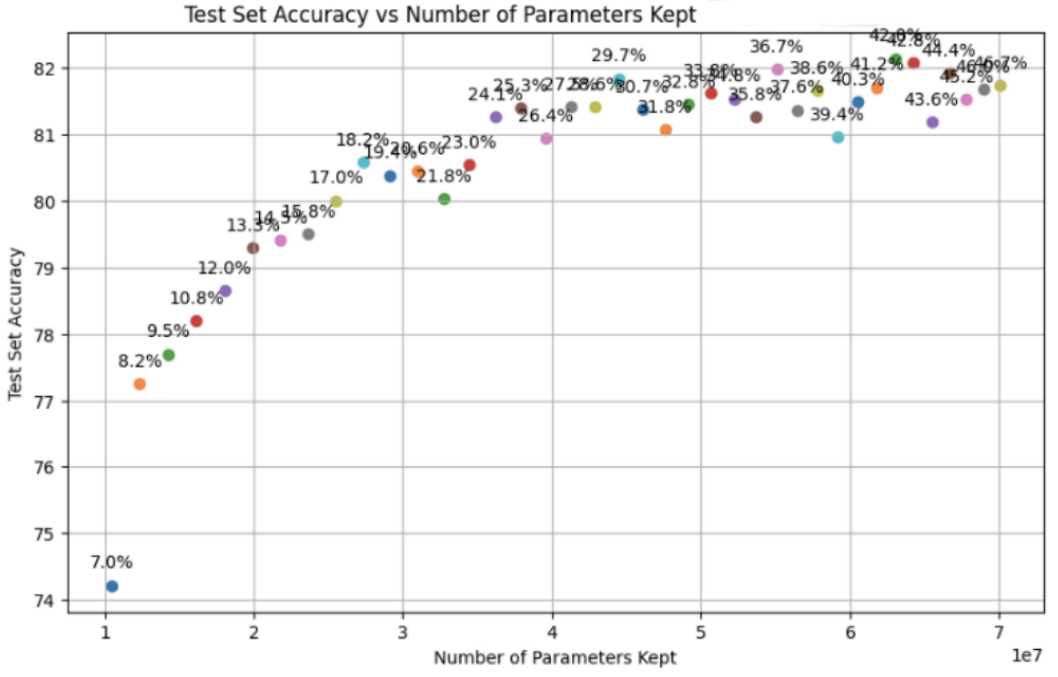}
        \caption{Sparsification without MP-based pruning. The percentage of parameters kept is shown above each point on the graph.}
        \label{fig:without_mp_pruning_2}
    \end{subfigure}
    \caption{Accuracy vs. number of parameters kept for CNNs trained on CIFAR-10 with and without MP-based pruning.}
    \label{fig:mp_pruning_comparison_2}
\end{figure}

As shown in Fig. \ref{fig:with_mp_pruning_2}, this additional sparsification leads to a large reduction in parameters, by over 97\%, without a significant drop in accuracy ($\sim 0$ drop). Finally, we tried the sparsification pruning method after training without MP-based pruning (during training). Fig. \ref{fig:without_mp_pruning_2} shows that the threshold pruning seems to affect the accuracy of the DNN in a much more significant manner. That is, even when pruning $80\%$ of the parameters, the accuracy drops by $\sim 2\%$. We see that a combination of MP-based pruning with sparsification is useful for pruning while also ensuring the DNN accuracy does not decrease much for CNNs trained on CIFAR-10.

\section{Mathematical underpinning of numerical results}
\label{theoretical_results}

In this section, we introduce the Pruning Theorem, which provides the relationship between the accuracy of a DNN before and after being pruned. First, we introduce an important tool for this analysis, the \textit{classification confidence} of a DNN.

\subsection{The classification confidence}
\label{Assumptions}

     We now introduce the classification confidence; see \cite{berlyand2021stability}. Take $X(s,\alpha)$ to be the output of the final layer in our DNN before softmax. The classification confidence is defined as follows:
  	
\begin{equation}\label{eq:defdeltaX}
\delta X(s,\alpha):=X_{i(s)}(s,\alpha)-\max_{j\neq i(s)} X_j(s,\alpha).
\end{equation}

  	In other words, 
  	\begin{itemize}
		\item $\delta X(s,\alpha(t))>0\Rightarrow s$ is well-classified by $\phi$.
		\item $\delta X(s,\alpha(t))< 0\Rightarrow s$ is misclassified by $\phi$
			\end{itemize}
			
			For $T'$ the test set we can now define the accuracy of the DNN on $T'$ using the classification confidence,
		\begin{equation}\label{2_class_prob}
		 \text{acc}_{\alpha}(t)=\frac{\#\left(\{s\in T':\delta X(s,\alpha(t))>0\}\right)}{\#T'}
		\end{equation}

\subsection{How pruning affects classification confidence (for deterministic weight layer matrices)}

 Now, we state a theoretical result that shows how, at least for simple DNN models, pruning the singular values of the weight layers of a DNN impacts the DNN accuracy. This result is not based on RMT but will help in understanding the results which follow. In the following lemma, we assume that we are given a threshold $\sqrt{\lambda_+}$, which we use to prune the singular values of the layers of the DNN. In general, this threshold is given by the MP distribution and numerically can be found using the BEMA algorithm, see Subsection \ref{finding_lambda}. For simplicity, for $W$ a matrix and $\beta$ a bias vector, we define $(W+\beta)s:=(Ws+\beta)$.

\begin{lem}
\label{main_result_new}
Let \( W_1, W_2, \cdots W_L \) and \( \beta_1, \beta_2, \cdots \beta_L \) be the weight matrices and bias vectors of a DNN with the absolute value activation function. Assume we prune a layer matrix \( W_b \) to obtain \( W'_b \) by removing singular values of \( W_b \) smaller than \( \sqrt{\lambda_+} \). For any input \( s \) (either from the training set $T$ or the test set $T'$), denote the change in classification confidence due to pruning as:
\begin{equation}
 \Delta(\delta X) = |\delta X(s,\alpha_{W_b}) - \delta X(s,\alpha_{W_b'})|. 
\end{equation}

Here $X(s,\alpha_{W_b})$ and $X(s,\alpha_{W_b'})$ are the outputs of the final layer before softmax of the DNN with weight layer matrices $W_b$ and $W_b'$ respectively.

Then 
\begin{equation}
    \Delta(\delta X) \leq  \sqrt{2\lambda_+} \|\lambda \circ (W_{b-1}+\beta_{b-1}) \circ \dots \circ \lambda \circ (W_{1}+\beta_1)s\|_2  \sigma_{\max}(W_{b+1}) \dots \sigma_{\max}(W_{L}),
\end{equation}

\end{lem}

See Subsection \ref{proof_main_result_1_new} for a proof of this Lemma. 

\begin{remark}
\label{remark_pruning}
For the simplified case when the bias vectors are zero, this lemma says that the change in classification confidence \( \delta X \) after pruning is bounded by:
\begin{equation}
    \label{eq_bound}
 \Delta(\delta X) \leq \sqrt{2\lambda_+} \|\lambda \circ W_{b-1} \circ \dots \circ \lambda \circ W_{1}s\|_2  
 \sigma_{\max}(W_{b+1}) \dots \sigma_{\max}(W_{L}).  
 \end{equation}
 This means that if elements were well classified before the pruning and $\sqrt{2\lambda_+}\|\lambda \circ W_{b-1} \circ \dots \circ \lambda \circ W_{1}s\|_2   \sigma_{\max}(W_{b+1}) \dots \sigma_{\max}(W_{L}) $ is small relative to $\delta X(s,\alpha_W)$, then after pruning $s$ will stay accurately classified.

 A crucial observation from the lemma is the product of the maximum singular values, denoted as \( \sigma_{\max}(W_{b+1}) \dots \sigma_{\max}(W_{L}) \). These singular values can be considerably large, implying that their product can amplify the magnitude of the bound \eqref{eq_bound}, thereby making it substantial.

At first glance, this might seem concerning as it suggests that pruning might lead to a large drop in the network's accuracy. However, this is not necessarily a grave issue. Subsequent sections will introduce two more theoretical results based on RMT, which will elucidate why, in practice, this potential drop in classification confidence due to pruning does not occur.

Furthermore, it is crucial to note that the lemma provides a worst-case scenario. In real-world scenarios, the actual impacts of pruning are expected to be much milder than what the lemma indicates. This is a common theme in theoretical computer science and machine learning: the worst case doesn't always reflect the average or common case.

Furthermore, the product $\sigma_{\max}(W_{b+1}) \dots \sigma_{\max}(W_{L})$ is used as a naive bound on the Lipschitz constant of the function $\ W_L \circ \lambda \circ \cdots \circ \lambda \circ W_{b+1}$. In practice, this value can be substantially smaller. There exist other methodologies for estimating the Lipschitz constant of this function which might yield a more conservative estimate. See \cite{fazlyab2019efficient} for more on the numerical estimation of the Lipschitz constant in deep learning. 

Another practical takeaway from this theorem is the preference to prune the final layers of the DNN rather than the earlier layers. The reasoning is simple: pruning the latter stages has a lesser effect on the overall accuracy, making it a safer bet in terms of maintaining the network's performance. However, this also depends on \( \|\lambda \circ (W_{b-1}+\beta_{b-1}) \circ \dots \circ \lambda \circ (W_{1}+\beta_1)s\|_2 \) which depends on the earlier layers in the network. 

In essence, while the theorem paints a potentially alarming picture of pruning's effects, practical simulations, and further theoretical results can assuage these concerns. The nuanced understanding provided by the theorem can guide efficient pruning strategies, ensuring minimal loss in accuracy.

\end{remark}

\begin{ex}
\label{result_deltax}
 The following example shows the histogram of $\delta X$ of a trained DNN for the problem given in Subsection \ref{Example_delta_X}. We train a DNN with one hidden layer. The weight layer matrices $W_1$, $W_2$ were initialized with components taken from i.i.ds, normally distribution with zero mean and variance $\frac{1}{N_l}$. We obtained a $98\%$ accuracy on the training set, which had $1000$ objects. $\delta X$ of the test set, after the $600$th epoch of training, is given in Fig. \ref{deltaX_final}.  

  \begin{figure}
		\centering	\includegraphics[scale=.7]{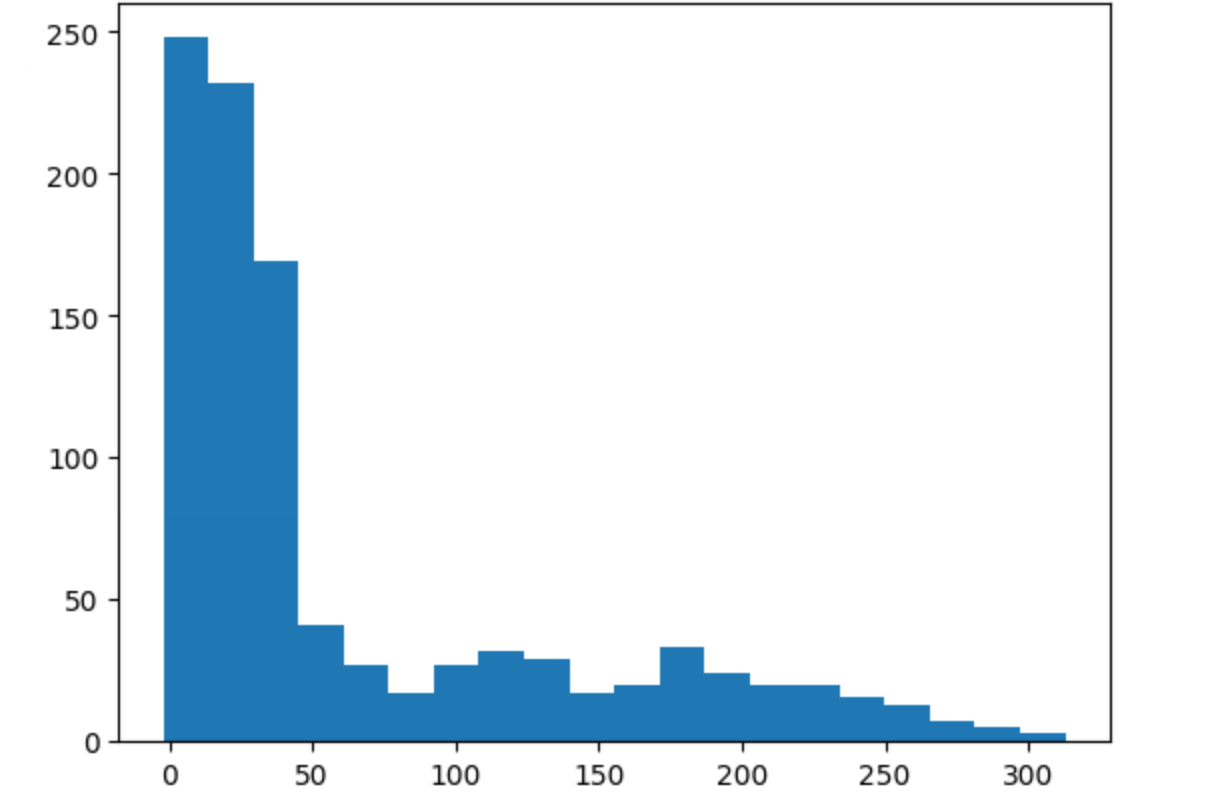}
			 \caption{Histogram of $\delta X$ of the test set for the final epoch of training. On the x-axis, we have the size of $\delta X$.}
			\label{deltaX_final}
		\end{figure}

 For the most part, we have $\|s\|_2 \leq \sqrt{2}$ and if we were to prune the first layer of the DNN, we would obtain $\sqrt{2\lambda_+}\sigma_{\max}(W_2)\|s\|_2\leq 6.5$.  However we see that for many objects $s$, $\delta X$ can be much larger than 
 $6.5$.  
  
\end{ex}

Next we would like to obtain a better result than Lemma \ref{main_result_new} using the properties of random matrices.

\subsection{Assumptions on the random matrix $R$ and the deterministic matrix $S$}
\label{assumptions_2}
  We considered a class of admissible matrices $W$, where $W=R+S$ and $W$, $R$ and $S$ satisfy the following three assumptions. The first  assumption is a condition on $R$: 

\textbf{Assumption 1}:
Assume \(R\) is a random \(N\times M\) matrix with entries taken from i.i.ds with zero mean and variance $\frac{1}{N}$. Further, as $N \to \infty$, we have that  $\sigma_{\max}(R) \to \sqrt{\lambda_+}$ a.s.    \\     


   We then assume the following for the deterministic matrix $S$:

   \textbf{Assumption 2:}

   Assume $S$ is a deterministic matrix with $S=\sum_{i=1}^r \sigma_i u_i v^T_i=U \Sigma V^T$, with $\sigma_i$ the singular values and $u_i$, $v^T_i$  column and row vectors of $U$ and $V$. Thus, $S$ has $r$ non-zero singular values corresponding to the diagonal entries of $\Sigma$, and all other singular values of $S$ are zero. We also assume that these $r$ singular values of $S$ have multiplicity $1$.\\

   Finally, we assume for $W:=R+S$:

   \textbf{Assumption 3}:  Take $\sigma_i$ to be the singular values of $S$, with corresponding left and right singular vectors $u_i$ and $v^T_i$ and $\sigma'_i$ to be the singular values of $W=R+S$, with corresponding left and right singular vectors $u'_i$ and $v'^T_i$. First we assume that $\frac{N}{M} \to c \in (0,+\infty)$ as $N \to \infty$. Second, assume also that we know explicit functions $g_{\sigma_i,R}$, $g_{v_i,R}$ and $g_{u_i,R}$ such that as $N \to \infty$:

\begin{equation}\label{singualr_value_cases_assumption}
\sigma'_i(W) \xrightarrow[\text{}]{a.s.} 
\begin{cases}
  g_{\sigma_i,R} & \sigma_i>\bar{\theta}(\lambda_+)\\
  \sqrt{\lambda_+} & \sigma_i<\bar{\theta}(\lambda_+),
\end{cases}
\end{equation}

\

\begin{equation}\label{left_singualr_vectors_cases}
|<u'_i,u_i >|^2 \xrightarrow[\text{}]{a.s.} 
\begin{cases}
  g_{u_i,R} & \sigma_i>\bar{\theta}(\lambda_+)\\
  0 & \sigma_i<\bar{\theta}(\lambda_+),
\end{cases}
\end{equation}

and

\begin{equation}\label{right_singualr_vectors_cases}
|\langle v'_i,v_i \rangle|^2 \xrightarrow[\text{}]{a.s.}
\begin{cases}
  g_{v_i,R} & \sigma_i>\bar{\theta}(\lambda_+)\\
  0 & \sigma_i<\bar{\theta}(\lambda_+).
\end{cases}
\end{equation}

Third, also assume that 
 for $i \neq j$:

   \begin{equation}
   |<v'_i,v_j >|^2 \xrightarrow[\text{}]{a.s.} 0 
   \end{equation}

   and 

     \begin{equation}  
   |<u'_i,u_j >|^2 \xrightarrow[\text{}]{a.s.} 0. 
   \end{equation}

Here we take $\bar{\theta}(\lambda_+)$ to be a known explicit function depending on $\lambda_+$, for example see \eqref{theta_bar}. In the Pruning Theorem (\ref{the_main_result_RMT}), we assume that a weight layer $W_b$ of the DNN satisfies Assumptions 1-3, that is $W_b=R_b+S_b$, with $R_b$ a random matrix, $S_b$ a low-rank deterministic matrix, and that the non-zero singular values of $S_b$ are bigger than some threshold $\bar{\theta}(\lambda_+)$. Empirically, it has been observed that these assumptions are reasonable for weight matrices of a DNN; see \cite{thamm2022random,staats2022boundary}.  There are various spiked models in which assumption $3$ holds, for more on the subject see \cite{baik2005phase, benaych2011eigenvalues,dharmawansa2022eigenvectors,couillet2022random, bao2021singular, o2018random, agterberg2022entrywise, chen2021asymmetry, bao2022eigenvector, leeb2021matrix, zhang2020tracy,dharmawansa2022eigenvectors,o2018matrices}. Also, a number of works in RMT addressed the connection between a random matrix $R$ and the singular values and singular vectors of the deformed matrix $W=R+S$, see \cite{benaych2011eigenvalues2, benaych2011eigenvalues}. 
For example, one can show that the following two simpler assumptions on the matrices $R$ and $S$ are sufficient to ensure that $R$ and $S$ satisfy the above assumptions $1-3$. Recall that a bi-unitary invariant random matrix $R$ is a matrix with components taken from i.i.ds such that for any two unitary matrices  $U$ and $V^T$, the components of the matrix $URV^T$ have the same distribution as the components of $R$. We then assume:

\textbf{Assumption 1'} (statistical isotropy):
Assume \(R\) to be a bi-unitary invariant random \(N\times M\) matrix with components taken from i.i.ds with zero mean and variance $\frac{1}{N}$.      


   We then assume the following for the deterministic matrix $S$:

   \textbf{Assumption 2'} (low rank of deterministic matrix): Assume $S$ is a deterministic matrix with $S=\sum_{i=1}^r \sigma_i u_i v^T_i=U \Sigma V^T$, with $\sigma_i$ the singular values and $u_i$, $v^T_i$  column and row vectors of $U$ and $V$. Thus, $S$ has $r$ non-zero singular values contained on the diagonal entries of $\Sigma$, and all other singular values are zero. We also assume that these $r$ singular values of $S$ have multiplicity $1$. Finally, we assume that $\frac{N}{M} \to c \in (0,+\infty)$ as $N \to \infty$. 

   An explicit relationship between assumptions 1'-2' and assumptions 1-3 can be found in \cite{benaych2011eigenvalues}. The assumption $1'$ is indeed strong, as it implies that the random matrix $R$ is random in every direction. In other words, for any unitary matrices $U$ and $V^T$, the matrix $URV^T$ has the same distribution as $R$. Random matrices with complex Gaussian entries, also known as Ginibre matrices, are a class of random matrices that are bi-unitary invariant  \cite{kosters2015limiting}. 

Assumptions $2$ and $2'$ are related to the low-rank property of the deterministic matrix, see \cite{thamm2022random,staats2022boundary} for how this assumption is related to DNNs. 
We consider the case where we initialize the weight layer of a DNN using a Gaussian random matrix (see Example \ref{simple_case}) and, after training, we obtain that $W_l = R_l + S_l$ with $R_l$ still a Gaussian random matrix and $S_l$ having low rank. The Pruning Theorem \ref{the_main_result_RMT} can be then employed to determine that removing the small singular values of $W_l$ will not affect much the accuracy of the DNN. This is because the deformed model $W_l = R_l + S_l$ satisfies Assumptions 1-3, see Example \ref{simple_case}.  This insight can be used to reduce the number of parameters in the DNN without sacrificing its performance, as will be further discussed in Subsection \ref{stable_acc}.\\

We now formulate theoretical results that provide a rigorous relationship between pruning and accuracy. Note that these results are applicable to DNNs with the following architecture: Consider a DNN, denoted by $\phi$, with weight layer matrices $W_1, \cdots, W_n$ and the absolute values activation function. We assume the layer maps of the DNN are compositions of linear maps and activation functions; however, the results can also be adapted to the case when the DNN is a composition of affine maps composed with activation functions, that is when we add basis. The central idea in these results can be described as follows. Suppose a weight layer $W_l$ of the DNN satisfies the above Assumptions 1, 2, and 3. Then, the removal of small singular values of $W_l$, smaller than the MP-based threshold $\sqrt{\lambda_+}$, does not change the classification confidence of an object $s \in \R^n$ by a "large amount" (see \eqref{pruning_equation_1} and \eqref{pruning_equation_2}). That is, the classification confidence before pruning and after pruning are essentially the same for sufficiently large matrix $W_l$.

In essence, these results suggest that it is possible to maintain the performance of the DNN while reducing the number of parameters by eliminating the small singular values, which are considered less influential in terms of the network's overall accuracy. This insight can be used to create more efficient DNN architectures, leading to reduced computational complexity and memory requirements without sacrificing model performance, see Appendix C.

\subsection{Key technical Lemma: removing random weights for DNN with arbitrary many layers does not affect classification confidence}

First, we introduce a result based on the assumption that we can directly know what parts of the weight layer matrices are deterministic and what parts are random. For a DNN $\phi$ with  weight layer matrices $W_1, \cdots, W_L$ we start by defining, 

\begin{equation}
\label{from_DNN_st}
    g_{\phi}(s,b):=\|\lambda \circ W_{b-1} \circ \dots \circ \lambda \circ W_{1}s\|_2 \sigma_{\max}(W_{b+1}) \dots \sigma_{\max}(W_{L}),
\end{equation}

and

\begin{equation}
\label{from_DNN_st_2}
    h_{\phi}(s,b):=\|\lambda \circ W_{b-1} \circ \dots \circ \lambda \circ W_{1}s\|_{1} \|W_{b+1}\|_1 \dots \|W_{L}\|_1,
\end{equation}
where $s$ is an element of the test or training set. The \(\ell_1\) norm of a matrix \(W\), denoted as \(\|W\|_1\), is defined as the maximum absolute column sum of the matrix. Formally, if \(W\) is an \(m \times n\) matrix with entries \(w_{ij}\), then
\[
\|W\|_1 = \max_{1 \leq j \leq n} \sum_{i=1}^m |w_{ij}|.
\]

Recall that $\sigma_{\max}(W_{b+1}) \dots \sigma_{\max}(W_{L})$ is a theoretical bound on the Lipschitz constant of $W_{b+1} \cdots W_{L}$, but numerically one might be able to obtain a better bound, see e.g.,   \cite{fazlyab2019efficient}. Here, we call the Lipschitz constant of a matrix $A$ the Lipschitz constant of the linear map corresponding to that matrix, and this definition extends to the product of matrices.  

The following lemma describes how the classification confidence changes when the weight layer matrix $W_b=R_b+S_b$ is changed with the weight layer matrix $S_b$-- the ultimate pruning.

\begin{lem}
\label{main_result_remove _R}
    Take $\phi$ to be a DNN with weight layer matrices $W_1, \cdots, W_L$ and absolute value activation function and fix object $s$ from the test set $T'$. Assume for some $b$ that $W_b=R_b+S_b$, with $R_b$ a $N \times M$ random matrix satisfying Assumption $1$ and matrix  $S_b$ a deterministic matrix satisfying Assumption $2$ given in Subsection \ref{assumptions_2}.

    Suppose we replace the weight layer matrix $W_b$ with the deterministic matrix $S_b$. Then we have that there $\exists D(N),a(N),b(N)$ such that for the classification confidence threshold of the non-pruned DNN:

   \begin{equation}
       E:=a(N) h_{\phi}(s,b) +b(N)
   \end{equation}
   we have the conditional probability
   \begin{equation}
   \label{pruning_equation_1}
   \mathbb{P}\bigg(\delta X(s,\alpha_{S_b})\geq 0  \mid \delta X(s,\alpha_{W_b}) \geq  E\bigg)\geq 1-D(N), 
   \end{equation}
 with $D(N), a(N) ,b(N)\to 0$ as $N \to \infty$ and $h_{\phi}(s,b)$ coming from \eqref{from_DNN_st_2}. Here, $\alpha_{S_b}$ are the parameters of the DNN, which has the weight matrix $S_b$ and  $\alpha_{W_b}$ are the parameters of the DNN with the weight layer matrix $W_b$.

\end{lem}

Here, when $R_b$ has components i.i.d. from $N(0,\frac{1}{N})$ then  $a(N)=\frac{2}{N^{\frac{1.5}{4}}}$, $b(N)=2\sqrt{\frac{2 \log N^2}{N}}$ and  we have that $D(N)=2 \exp \left( - \frac{N^{1/4}}{2}\right)$.  The proof for this lemma can be found in Subsection \ref{proof_main_result_remove_random}.

\begin{remark}
    Here, we take $s$ from the test set $T'$; however, the result also holds if we take $s$ from the training set $T$. Furthermore, using a proof similar to the one given in Subsection \ref{proof_main_result_remove_random}, one can show a more general result. That is, taking
    \begin{equation}
 \Delta(\delta X) := |\delta X(s,\alpha_{S_b}) - \delta X(s,\alpha_{W_b})|, 
\end{equation}
we have that 
\begin{equation}
    \mathbb{P}\bigg(\Delta(\delta X)\leq E\bigg)\geq 1-D(N).
\end{equation}

\end{remark}

\begin{remark}
\label{sparsification}

Lemma \ref{main_result_remove _R} addresses the removal of parameters while preserving accuracy in a more general context than MP-based pruning. In particular, it also explains the numerics of parameter removal via sparsification (see Section \ref{num_results}). Indeed, in Lemma \ref{main_result_remove _R}, the random matrix $R_b$ has entries taken from i.i.ds with zero mean and variance $\frac{1}{N}$. Therefore, as $N \to \infty$ the entries of $R_b$ are small w.r.t. sparsification threshold $\xi(N)$. If, in addition, we assume that the entries of $S_b$ are large  (c.f. assumption in Theorem \ref{the_main_result_RMT}), then large entries of $W_b=R_b+S_b$ are entries of $S_b$ with high probability. Therefore, sparsifying $W_b$ by removing the entries smaller than the $\xi(N)$ amounts to replacing the weight layer matrix $W_b$ with the deterministic matrix $S_b$. Therefore, Lemma \ref{main_result_remove _R} implies that sparsification preserves accuracy in the sense of \eqref{pruning_equation_1}.


\end{remark}

\begin{remark}


Imagine you are trying to predict the weather. Initially, your prediction is based on both the randomized patterns you have observed over time (noise in input layer) and your initial random weights of the DNN (matrix \( R \)) and the deterministic factors you are sure of (matrix \( S \)).  Now, if you decide to base your prediction just on the deterministic factors (that is, totally remove the random part of the weight layer matrix). Then how much would your confidence in the prediction change? This Lemma provides a bound on that change. 

The Lemma states that there exists a function $D(N) \to 0$ as $N \to \infty$  (the size of the matrix) increases. The magnitude of this change in classification confidence (how much our "confidence" drops when we remove the random part) is given by \( E \), which is related to the combined effects of all layers up to \( b \) and the maximum scaling factors (or singular values) of layers after \( b \). 

Most importantly, the conditional probability states that if our initial confidence (with the random matrix) was bounded away by $E>0$, then after removing the randomness, our confidence would most likely be at least \( 0 \). And as the size \( N \) of the matrix increases, the probability that our classification confidence would be bigger than $0$ becomes closer to $1$.

In essence, even if we remove the randomness from our prediction model (in this case, the DNN), we can still be quite confident about our predictions, especially as our layer widths grow.

\end{remark}

\begin{remark}
\label{remark_pruning_S}
We want to understand what happens to the classification confidence threshold $E$ in \eqref{pruning_equation_1} as $N \to \infty$. Assuming that \begin{equation}
    h_{\phi}(s,b)\leq C
\end{equation} 
for all $N$, then $E \to 0$ as $N \to \infty$. This is because  \( a(N) = \frac{2}{N^{\frac{1.5}{4}}} \), goes to zero as \( N \) increases. Consequently, the contribution from \( a(N) \) becomes negligible, implying that the effect of dropping the random matrix and only keeping the deterministic matrix becomes inconsequential. 

\end{remark}

\subsection{Pruning Theorem for DNN with arbitrary many layers: how pruning random weights using PM distribution affects the classification confidence}

\begin{thm}[The Pruning Theorem for a single object]
\label{the_main_result_RMT}
    Take $\phi$ to be a DNN with weight layer matrices $W_1, \cdots, W_n$ and absolute value activation functions and take some $s \in T'$, with $T'$ the test set. Assume for some $b$ that $W_b=R_b+S_b$, with $R_b$ a $N \times M$ random matrix satisfying assumption $1$, matrix  $S_b$ a deterministic matrix satisfying assumption $2$ and $R_b+S_b$ satisfying assumption 3, for the assumptions given in Subsection \ref{assumptions_2}. Further, assume that all the non-zero singular values of $S$ are bigger than $\bar{\theta}(\lambda_+)$  with $\lambda_+$ given by the MP distribution of the ESD of $R_b^TR_b$ as $N \to \infty$ and $\bar{\theta}(\lambda_+)$ given in \eqref{singualr_value_cases_assumption}. 

    Construct the truncated matrix $W'_b$ by pruning the singular values of $W_b$  smaller than $\sqrt{\lambda_+} +\epsilon$ for any $\epsilon$. Then 
   we have that there exists an explicit function $f_{W_b}>0$ such that $\forall \epsilon$, $\exists C_{\epsilon}(N)$ so that for the classification confidence threshold of the non-pruned DNN:
     
    \begin{equation} 
    \label{pruning_equation_2}
     E':=(1+\epsilon)(\sqrt{2}(1+\epsilon)\min\{f_{W_b},\sqrt{\lambda_+}\}g_{\phi}(s,b) +a(N)h_{\phi}(s,b)+b(N))
   \end{equation}
   we have the conditional probability
   \begin{equation}
   \mathbb{P}\bigg(\delta X(s,\alpha_{W_b'})\geq 0  \mid \delta X(s,\alpha_{W_b}) \geq E'\bigg)\geq 1-C_{\epsilon}(N), 
   \end{equation}
with $C_{\epsilon}(N), a(N),b(N) \to 0$ as $N \to \infty$, $g_{\phi}(s,b)$ coming from \eqref{from_DNN_st} and $h_{\phi}(s,b)$ coming from \eqref{from_DNN_st_2}.  $f_{W_b}$ is given in Lemma \ref{approx_lemma_example}. Also, $\alpha_{W_b}$ are the parameters of the DNN that has the weight matrix $W_b$ and  $\alpha_{W_b'}$ are the parameters of the DNN with the weight layer matrix $W_b'$.

\end{thm}

See Subsection \ref{proof_main_result_RMT} for a proof of this theorem. 

\begin{remark}
    Here we take $s$ from the test set $T'$; however, the result also holds if we take $s$ from the training set $T$. Furthermore, using the proof given in Subsection \ref{proof_main_result_RMT}, one can show a more general result. That is, taking
    \begin{equation}
 \Delta(\delta X) := |\delta X(s,\alpha_{W'_b}) - \delta X(s,\alpha_{W_b})|, 
\end{equation}
we have that 
\begin{equation}
    \mathbb{P}\bigg(\Delta(\delta X)\leq E'\bigg)\geq 1-C_{\epsilon}(N).
\end{equation}

\end{remark}

\begin{remark}

The truncation of the matrix \(W_b\) is done by pruning the singular values smaller than \(\sqrt{\lambda_+} + \epsilon\). This choice is made because, as \(N \to \infty\), the largest singular value \(\sigma_{\max}(R_b)\) of the random matrix \(R_b\) converges to \(\sqrt{\lambda_+}\) a.s. However, for finite \(N\), there can be fluctuations around \(\lambda_+\) due to the inherent randomness of the matrix. These fluctuations are described by the Tracy-Widom distribution. To account for these fluctuations and ensure robustness in the pruning process, we add a small positive \(\epsilon\) to \(\sqrt{\lambda_+}\).

In the numerical part of the paper, the value of \(\lambda_+\) was determined using the BEMA algorithm (see Subsection \ref{finding_lambda}.) This algorithm approximates \(\lambda_+\) by incorporating the Tracy-Widom distribution to account for the finite-size effects and the fluctuations of the largest eigenvalue of $R^T_bR_b$. By using this algorithm, we obtain a more accurate estimation of \(\lambda_+\) for practical, finite-dimensional settings, which is crucial for effectively applying the pruning theorem in real-world scenarios.
    
\end{remark}

    Pruning Theorem \ref{the_main_result_RMT} shows that if we replace the matrix $W_b$ with a truncated matrix $W'_b$, then for any given object $s \in T'$, we have that if the classification confidence $\delta X(s,\alpha)$ is positive enough for matrix $W_b$, it stays positive for the truncated matrix $W'_b$ with high probability. In other words, almost all well-classified objects remain well-classified after replacing $W_b$ with $W'_b$.   We also show numerically that it is easier to prevent overfitting using matrix $W'_b$ instead of the larger matrix $W_b$.  We verified that removing small singular values based on the MP-based threshold $\sqrt{\lambda_+}$ for the case when the weight matrices $W_b$ were initialized with $N(0,\frac{1}{N})$ does not reduce the accuracy of the DNN; see Example \ref{threshold_example}. Here, $f_W = \|W'-S\|$ and for a large class of RMT matrix models (see assumptions 1-3 in Subsection \ref{assumptions_2}), for the case $N \to \infty$, we obtain $f_W$ based on the singular values of $W$ only.

\begin{remark}
    The assumptions made in Theorem \ref{the_main_result_RMT} are quite natural and hold for a wide range of DNN architectures. Assumption $1$ focuses on the random matrix $R$. This assumption ensures that the random matrix $R$ captures the essential randomness in the weight layer while also satisfying the requirements given in Theorem \ref{RMT_MP_theorem}. 

Assumption $2$ pertains to the deterministic matrix $S$, which is assumed to have a specific structure, with $r$ non-zero singular values and all other singular values being zero. Moreover, these $r$ singular values have multiplicity 1, which is a reasonable expectation for a deterministic matrix that contributes to the information content in the layer $W_b$. Assumption $3$ holds for many spiked models and has been studied in much detail.

The assumption that the singular values of the deterministic matrix $S$ are larger than some $\bar{\theta}(\lambda_+)$ is also quite natural, see \cite{thamm2022random,staats2022boundary}. This is because the deterministic matrix $S$ represents the information contained in the weight layer, and its singular values are expected to be large, reflecting the importance of these components in the overall performance of the DNN. On the other hand, the random matrix $R$ captures the inherent randomness in the weight layer, and with high probability depending on $N$, its singular values should be smaller than the MP-based threshold. This means that there is a clear boundary between the information and noise in the layer $W_b$, which is also natural, see \cite{staats2022boundary}. 


This distinction between the singular values of $S$ and $R$ highlights the separation between the information and noise in the weight layer, allowing us to effectively remove the small singular values without impacting the accuracy of the DNN. The assumption thus provides a solid basis for studying the behavior of DNNs with weight layers modeled as spiked models. It contributes to our understanding of the effects of removing small singular values based on the random matrix theory MP-based threshold $\sqrt{\lambda_+}$.
\end{remark}

\begin{remark}
    
In our work, we leverage the Marchenko-Pastur distribution to select significant singular values for the low-rank approximation of our weight layers $W_l$. Other low-rank approximation techniques, such as the bootstrapping technique proposed in \cite{naumov2019bootstrap}, could potentially be integrated with the Marchenko-Pastur distribution to further refine the low-rank approximation of $W_l$.

\end{remark}

\subsection{Simple example of DNN with one hidden layer}
\label{Corollary}

   The following is a simple example of the Pruning Theorem:

\begin{ex}

\label{example_main_result_1}
    Take $\phi$ to be a DNN with three weight layer matrices $W_1,W_2,W_3$ and the absolute value activation function and take $s \in T'$. This is: \begin{equation}
			\vspace{-.2cm}
			\phi(s,\alpha)=\rho \circ \lambda \circ W_3  \circ \lambda \circ W_2 \circ \lambda \circ W_1 s, \hspace{.4cm} s\in \R^n. 
		\end{equation}

		Assume $W_2=S_2+R_2$ satisfies Assumptions 1-3 and $W_1$, $W_3$ are arbitrary. More specifically, assume $R_2$ to be a random matrix with i.d.ds taken from the distribution $N(0,\frac{1}{N})$ and $S_2$ to be a $N \times N$ deterministic matrix with non-zero singular values bigger than $1$.         
    
   Take $W'$ to be the same as $W$ but with all the singular values of $W$ smaller than $2+\epsilon$, for any $\epsilon$, set to zero. Then for $f_W$ the positive function given in \eqref{f_w_simple_case}
   we have that $\forall \epsilon>0$:

   \begin{equation}
   \mathbb{P}\bigg(\delta X(s,\alpha_{W'})\geq 0 \mid  \delta X(s,\alpha_W)\geq  (1+\epsilon)\sqrt{2}(a(N)\|W_1s\|_{1}\|W_3\|_1+f_{W_2}\|W_1s\|_2\sigma_{\max}(W_3)+b(N))\bigg)\geq 1-C_{\epsilon}(N),
   \end{equation}
    with $C_{\epsilon}(N),a(N),b(N) \to 0$ as $N \to \infty$.

\label{f_W_for_example}
Here we have that for $\sigma_r$ the smallest non-zero singular value of $S_2$: 
  \begin{equation}
  \label{f_w_simple_case}
\begin{split}
f_W = \max_{1\leq i\leq r} \left\{ \sqrt{ \left(\sigma^2_i+\left(\frac{1+\sigma^2_i}{\sigma_i}\right)^2 \left( 1-\frac{1}{\sigma^2_i}\right)\right) - 2(1+\sigma^2_i) \left(1-\frac{1}{\sigma^2_i}\right)} \right\}.
\end{split}
\end{equation}

Note that as $\sigma_r \to \infty$ we have $f_W \to 1$. In fact, we show numerically that already for $\sigma_r\geq 5$ we have $|f_W-1|\leq .03$, see Fig \ref{f_W}.

\begin{figure}[h!]
\centering
\includegraphics[width=.5\linewidth]{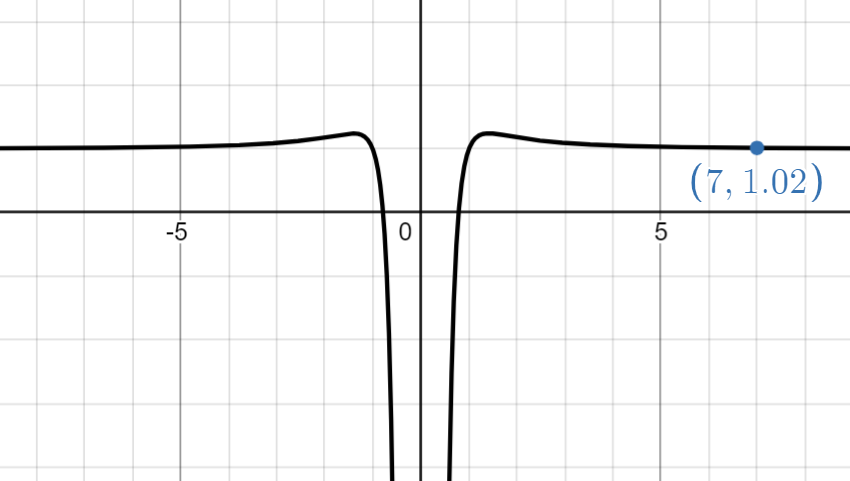}
\caption{
Graph of $f_W$. On the x-axis, we have $\sigma_i$. 
}
\label{f_W}
\end{figure}

Note that this estimate is given in terms of the singular values of $S$ (which are $\sigma_i$). However, the singular values of $S$ might not be known. Nevertheless, by Theorem \ref{simple_case}, we have that as $N \to \infty$ the singular values of $S$ can be obtained directly from the singular values of $W$ via $\sigma'_i=\frac{1+\sigma^2_i}{\sigma_1}$. Thus, as $N \to \infty$ this estimate can be obtained in terms of singular values of $W$ only, which is why we use the notation $f_W$ and not $f_S$. For simplicity, we keep using the current notation. 

We numerically checked that for R a $3000 \times 3000$ random matrix initialized with the above Gaussian distribution, and for $S$ a diagonal matrix with $5$ non-zero singular values given by $30, 40, 50, 60, 70$, we have $\|S-W'\|_2 \approx f_W \approx 1$, see Fig. \ref{fig:S-W'}. Thus, in this example $f_W < \sqrt{\lambda_+}$, given that $\sqrt{\lambda_+}=2$, and so in \eqref{main_equation} we would have $\min\{f_{W_b},\sqrt{\lambda_+}\} \approx 1$. Thus, Theorem \ref{the_main_result_RMT} provides a better result than Lemma \ref{main_result_new}. For more on this example, see Subsection \ref{gass_example_1}.

\begin{figure}[ht]
\centering
\includegraphics[width=0.5\linewidth]{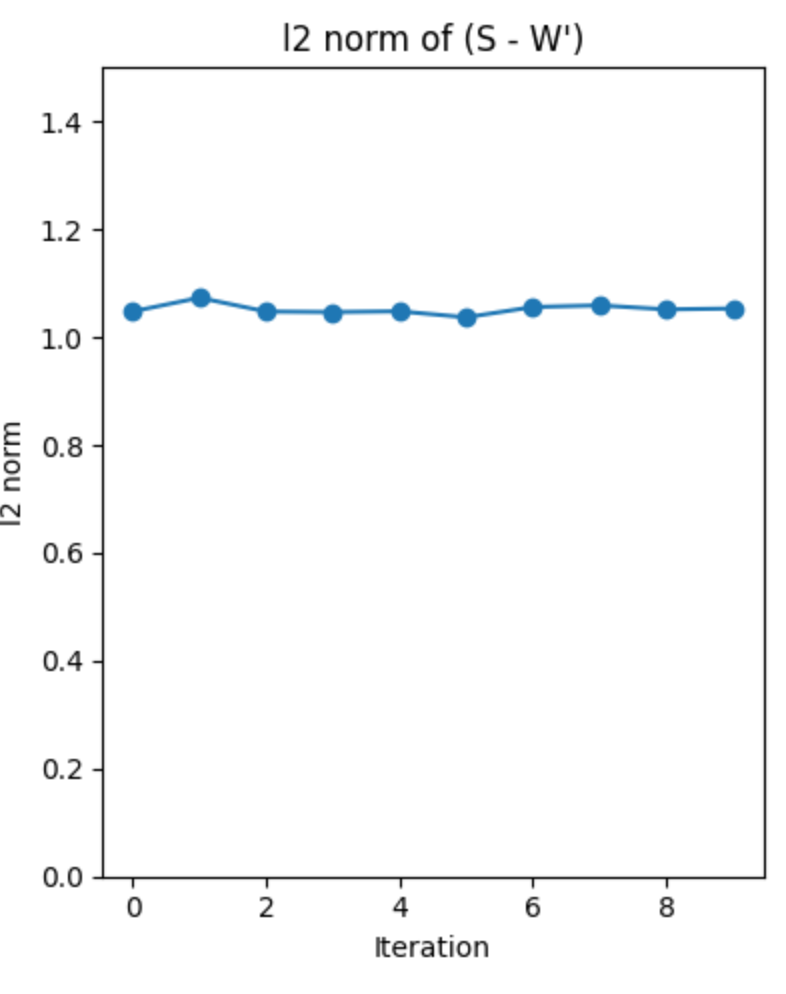}
\caption{
The norm $\|S-W'\|_2$ is shown for $10$ random matrices with elements i.i.ds taken from $N(0,\frac{1}{N})$ and $S$ a diagonal matrix with elements on the diagonal given by $30, 40, 50, 60, 70$. We see that the norm is very close to $f_W\approx 1$. 
}
\label{fig:S-W'}
\end{figure}

\textbf{It is an important question:} under what conditions of $R$ and $S$ would we have that $f_W = c$ such that $c < \sqrt{\lambda_+}$.

\end{ex}

\begin{figure}[ht]
\centering
\includegraphics[width=0.5\linewidth]{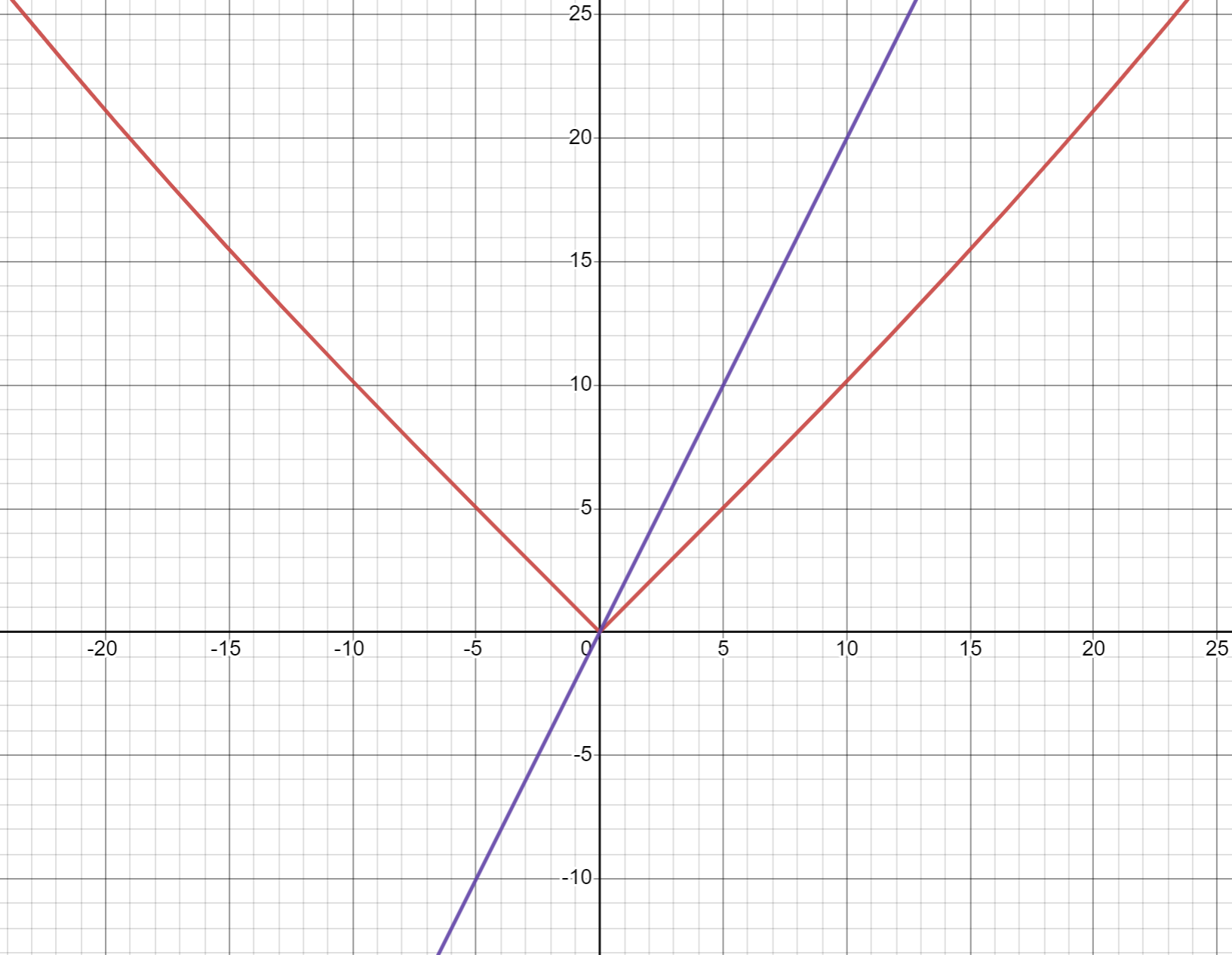}
\caption{
A graphical comparison between $f_W$ and $2\sigma$ as functions of $\sigma$. The x-axis represents the variable $\sigma$ ranging from -25 to 25. The y-axis provides the computed values for both $f_W$ and $2\sigma$. We see that $f_W\approx|
\sigma|$, which is smaller than $2\sigma$ for positive $\sigma$. Note, to obtain this numerical result, we must ensure that $\sigma_r \geq \sigma$. 
}
\label{fig:comparison_fw_2sigma}
\end{figure}

\begin{ex}
\label{Example_2_main}

    Consider \( R \) to be an \( n \times n \) symmetric (or Hermitian) matrix with independent, zero mean, normally distributed entries. The variance of the entries is \( \frac{\sigma^2}{n} \) on the diagonal and \( \frac{\sigma^2}{2n} \) on the off-diagonal.

In the setting where $S = \sum_{i=1}^r\sigma_i u_i u_i^T $, let \( u_i' \) be the unit eigenvectors of \( W = R + S \) associated with its r largest eigenvalues. Assuming that for all $1\leq i\leq r$ we have $\sigma_i>\sigma$, then:

\begin{equation}
  \label{f_w_simple_case_2}
\begin{aligned}
f_W &= \max_{1\leq i \leq r} \sqrt{\left(\sigma_i^{2}+\left(1-\frac{\sigma^{2}}{\sigma_i^{2}}\right)\left(\sigma_i+\frac{\sigma^{2}}{\sigma_i}\right)^{2}\right)-2\sigma_i\left(\left(\sigma_i+\frac{\sigma^{2}}{\sigma_i}\right)\left(1-\frac{\sigma^{2}}{\sigma_i^{2}}\right)\right)}.
\end{aligned}
\end{equation}

In this example, $\sqrt{\lambda_+}=2\sigma$. Thus, in Fig. \ref{fig:comparison_fw_2sigma} we compare $f_W$ vs $2\sigma$. As mentioned, it would be interesting to try and find a probability distribution which, if $R$ is initialized with would result in a very small $f_W$ for reasonable assumptions on the singular values of $S$.

See Subsection \ref{Example_2_main_num} for more information. 

\end{ex}

\subsection{Pruning Theorem for accuracy: how pruning affects accuracy }

In this subsection, we present a version of the Pruning Theorem for accuracy, which describes how the accuracy of a DNN is affected by pruning. We present this Theorem for DNNs with one hidden layer. However, it can be generalized for DNNs with more layers.

We first recall the notion of the \textit{good set} of a DNN, introduced in \cite{berlyand2021stability}. The good set is a subset of the test set $T'$ defined as follows:
	for $\eta\geq 0$, the \emph{good set} of margin $\eta$ at time $t$ is
	\begin{equation}\label{eq:good_set}
G_{\eta(t),\alpha}:=\{s\in T':\delta X(s,\alpha(t))>\eta\}.
	\end{equation}

Basically, the good set consists of positively classified objects whose classification confidence is bounded below by $\eta$. Next, we formulate the Pruning Theorem for accuracy. Loosely speaking, it says that for some threshold $E_{acc}$ (see \eqref{E_acc}), we have that the accuracy of the DNN after pruning is bounded from below by the number
\begin{equation}   \frac{|G_{E_{acc},\alpha}|}{|T'|},
\end{equation}
where for a finite set $A$, we have $|A|$, which is the number of elements in that set.

 \begin{thm}[Pruning Theorem for accuracy]
 \label{Cor_1}
 
		Let $\phi$ be a DNN with weight layer matrices $W_1,W_2,W_3$, and $\lambda$  the absolute value activation function: \vspace{-.3cm}
		
		\begin{equation}
			\vspace{-.2cm}
			\phi(s,\alpha)=\lambda \circ W_3  \circ \lambda \circ W_2 \circ \lambda \circ W_1 s, \hspace{.4cm} s\in \R^n. 
		\end{equation}

		Assume $W_2=S_2+R_2$ satisfies Assumptions 1-3 in Subsection \ref{assumptions_2}, and $W_1$, $W_3$ are arbitrary matrices.      
		
		 Construct the truncated matrix $W'_2$ by pruning singular values of $W_2$ smaller than $\sqrt{\lambda_+}+\epsilon$, for any $\epsilon$. For every $\epsilon$, introduce the classification confidence threshold for the non-pruned DNN as the smallest number $E_{acc}\geq 0$ for which we satisfy:
    \begin{equation}
    \label{E_acc}
    E_{acc}=(1+\epsilon)(\sqrt{2}(f_{W_2}\sigma_{\max}(W_3)+a(N)\|W_3\|_1))\max_{s\in G_{E_{acc},\alpha}} \|W_1s\|_1 +b(N)\quad (\text{positive}),
    \end{equation}
   with  $f_{W_2}>0$ an explicit rational function of $W_2$. Then we have $\forall \epsilon$:
   
   \begin{equation}
   \label{main_eq_2}
    \mathbb{P}\bigg(acc_{\alpha'}(t)\geq \frac{|G_{E_{acc},\alpha}|}{|T'|}
\bigg)\geq (1-C_{\epsilon}(N))^{|G_{E_{acc}\alpha}|}
    \end{equation}
with $C_{\epsilon}(N),a(N),b(N) \to 0$ as $N \to \infty$. Here  $\alpha,\alpha'$ are the parameters of the non-pruned and pruned DNNs, respectively and $acc_{\alpha'}(t)$ is given in \eqref{2_class_prob}.
\end{thm}

See Subsection \ref{RMT_corollary_proof} for a proof of this Theorem. 

\begin{remark}
    
Theorem \ref{Cor_1} applies to the entire training and test set.  It is important to note that if the sizes of the training and test sets depend on the matrix dimension \(N\), then the properties of the good set, which is derived from these sets, will inherently depend on \(N\) as well. The specifics of this dependency remain undefined within the current scope of our analysis.

\end{remark}

\begin{remark}
  One can always find a $E_{acc} \geq 0$ which satisfies  \eqref{E_acc}. 
This is because $0\leq (1+\epsilon)(\sqrt{2}(f_{W_2}\sigma_{\max}(W_3)+a(N)\|W_3\|_1))\max_{s\in G_{0,\alpha}} \|W_1s\|_1$. Finally, for large enough $\eta$, we have $(1+\epsilon)(\sqrt{2}(f_{W_2}\sigma_{\max}(W_3)+a(N)\|W_3\|_1))\max_{s\in G_{\eta,\alpha}} \|W_1s\|_1 =0$.
\end{remark}
	\bibliographystyle{alpha}

\bibliography{DNNstability_biblio}

  \section{Appendix A: Some known results on perturbation of matrices}

       Matrix perturbation theory is concerned with understanding how small changes in a matrix can affect its properties, such as eigenvalues, eigenvectors, and singular values. In this section, we state a couple of known results from matrix perturbation theory.
       \subsection{Asymptotics of singular values and singular vectors of deformation matrix}
       \label{bi_unitary_case}
The results in this subsection are taken from \cite{benaych2011eigenvalues}. Given the assumptions 1'-2' on $R$ and $S$ described in Section \ref{assumptions_2}, the authors were able to show that the largest eigenvalues and corresponding eigenvectors of $W=S+R$  are well approximated by the largest eigenvalues and eigenvectors of $S$.

We start by defining the following function:  \begin{equation}
D_{\mu_R}(z) = \left[ 
\int \frac{z}{z^2 - t^2} d\mu_R(t)
\right] \times \left[ c \int \frac{z}{z^2-t^2} d\mu_R(t)  + \frac{1-c}{z}\right]
\end{equation}

for $z>\sqrt{\lambda_+}$, with $\lambda_+$ given by the MP distribution of $R^TR$. Take $D^{-1}_{\mu_R}(\cdot)$ to be its functional inverse.

Set \begin{equation} 
\label{theta_bar}
\bar{\theta}= D_{\mu_R}(\sqrt{\lambda_+})^{-\frac{1}{2}}
\end{equation}

\begin{thm}\label{phase_transition_2}{Theorem for large singular values [Benaych-Georges and Nadakuditi (2012)]}
Take $W=R+S$, with $W, R$ and $S$ all  $N \times M$ matrices satisfying assumptions $1'-2'$. The $r$ largest singular values of $W$, denoted as $\sigma'_i(W)$ for $1 \leq i \leq r$,  exhibit the following behaviour as $N \to \infty$:

\begin{equation}\label{singualr_value_cases}
\sigma'_i(W) \xrightarrow[\text{}]{a.s.} 
\begin{cases}
  D^{-1}_{\mu_R}(\frac{1}{(\sigma_i)^2}) & \sigma_i>\bar{\theta}\\
  \sqrt{\lambda_+} & \sigma_i<\bar{\theta}
\end{cases}
\end{equation}
 
\end{thm}

\begin{thm}{Norm of projection of largest singular vectors [Benaych-Georges and Nadakuditi (2012)]}\label{singualr_vector_cases}

Take indices $i_0 \in \{1, . ..,r\}$  such that $\sigma_{i_0} > \Bar{\theta}$. Take $\sigma'_{i_0} =\sigma'_{i_0}(W)$ and let $u',v'$ be left and right unit singular vectors of $W$ associated with the singular value $\sigma'_{i_0}$ and $u,v$ be the corresponding singular vectors of $S$.  Then we have, as $N \to \infty$:

\begin{equation}\label{left_singualr_vectors_cases_2}
|<u',\mathrm{Span}\{u_i \ s.t. \ \sigma_i=\sigma_{i_0}\}>|^2 \xrightarrow[\text{}]{a.s.} \frac{-2\phi_{\mu_R}(\rho)}{\sigma^2_{i_0}D'_{\mu_R}(\rho)}
\end{equation}

and 

\begin{equation}\label{right_singualr_vectors_cases_2}
|<v',\mathrm{Span}\{v_i \ s.t. \ \sigma_i=\sigma_{i_0}\}>|^2 \xrightarrow[\text{}]{a.s.} \frac{-2\phi_{\mu_R}(\rho)}{\sigma^2_{i_0}D'_{\Tilde{\mu_R}}(\rho)}
\end{equation}

Here $\rho=D^{-1}_{\mu_R}(\frac{1}{(\sigma_{i_0})^2})$ and $\Tilde{\mu_R}=c\mu_R+(1+c)\delta_0$.

Further, 

\begin{equation}\label{left_singualr_vectors_cases_3}
|<u',\mathrm{Span}\{u_i \ s.t. \ \sigma_i\neq \sigma_{i_0}\}>|^2 \xrightarrow[\text{}]{a.s.} 0
\end{equation}

\begin{equation}\label{right_singualr_vectors_cases_3}
|<v',\mathrm{Span}\{v_i \ s.t. \ \sigma_i\neq \sigma_{i_0}\}>|^2 \xrightarrow[\text{}]{a.s.} 0
\end{equation}

\end{thm}

\begin{ex}\label{simple_case}
Take $S=\sum_{i=1}^r\sigma_i u_i v_i^T$ to be a $N \times N$ deterministic matrix, with $\sigma_i$ the singular values and $v_i$ and $u_i$ the singular vectors of $S$. Take $R$ to be a $N \times N$ random matrix with real i.i.d components taken from the normal distribution $N(0,\frac{1}{N})$. For $W=R+S$ we have:

\begin{thm}(Theorem for large singular values for Example \ref{simple_case})
\label{singualr_values_simple_case}
The $r$ largest singular values of $W$, denoted  $\sigma'_i(W)$ for $1 \leq i \leq r$,  exhibit the following behaviour as $N \to \infty$:
\[
      \sigma'_i(W) \xrightarrow[\text{}]{a.s.} 
\begin{cases}
  \frac{1+\sigma_i^2}{\sigma_i} & \sigma_i>1\\
  2 & \sigma_i<1
\end{cases}
\]
\end{thm}

\begin{thm}(Theorem for large singular vectors  for Example \ref{simple_case})
\label{singualr_vectors_simple_case}
Assuming that the $r$ largest singular values of $W$ have multiplicity $1$, then the right and left singular vectors $u'_i, v'_i$ of $W$ corresponding with the $r$ largest singular values $\sigma'_i(W)$  exhibits the following behaviour as $N \to \infty$:
 \[
      |<v_i,v'_i>|^2, |<u_i,u'_i>|^2  \xrightarrow[\text{}]{a.s.}
\begin{cases}
  (1-\frac{1}{\sigma^2_i}) & \sigma_i>1\\
  0 & \sigma_i<1
\end{cases}
\]  
\end{thm}

\end{ex}

\begin{remark}
\label{a.s_CIL}
   We say that $X_n \to X$ in probability if $\forall \epsilon$ $\mathbb{P}(|X_n-X|>\epsilon) \to 0$ as $n \to \infty$. One can show that $X_n \to X$ a.s. implies that $X_n \to X$ in law. Thus for Theorems \ref{phase_transition_2} and \ref{singualr_vector_cases} we have that if the $r$ large singular values  of $S$ are bigger than $1$ and have multiplicity $1$ then there exists some constant $B_N(\epsilon)$ with $B_N(\epsilon) \to 0$ as $N \to \infty$ such that for the $r$ largest singular values of $W$ and their corresponding singular vectors and $\forall \epsilon$ we have:

   \begin{equation}
   \label{singualr_value_approx}
   \mathbb{P}(|\sigma'_i(W) -\frac{1+\sigma_i^2}{\sigma_i}|>\epsilon)<B_N(\epsilon),
   \end{equation}

   \begin{equation}
      \mathbb{P}(||<v_i,v'_i>|^2 - (1-\frac{1}{\sigma^2_i})|>\epsilon)<B_N(\epsilon)
   \end{equation}

   and 

    \begin{equation}
      \mathbb{P}(||<u_i,u'_i>|^2 - (1-\frac{1}{\sigma^2_i})|>\epsilon)<B_N(\epsilon). 
   \end{equation}

   Finally, we also have that:

   \begin{equation}
   \label{cross_D_2}
      \mathbb{P}(|<u_i,u'_j>|>\epsilon)<B_N(\epsilon). 
   \end{equation}

   and 

    \begin{equation}
    \label{cross_D_1}
      \mathbb{P}(|<v_i,v'_j>|>\epsilon)<B_N(\epsilon). 
   \end{equation}

 for $i \neq j$. 
\end{remark}

\subsection{Gershgorin's Circle Theorem}

Finally, we state Gershgorin's Circle Theorem:

\begin{thm}
[Gershgorin's Circle Theorem]
\label{Gershgorin's Circle Theorem}
Let $B = [b_{ij}]$ be an $n \times n$ complex matrix. Define the Gershgorin discs $D_i$ for $1 \leq i \leq n$ as
\begin{equation} 
\label{GCT}
D_i = \left\{z \in \mathbb{C} : |z - b_{ii}| \leq \sum_{j \neq i} |b_{ij}|\right\}.
\end{equation}
Then, every eigenvalue $\lambda$ of the matrix $B$ lies within at least one of the Gershgorin discs $D_i$.
\end{thm}

\begin{remark}
We apply Theorem \ref{Gershgorin's Circle Theorem} for almost diagonal matrices when \eqref{GCT} estimates how close the eigenvalues are to the diagonal elements. This closeness is estimated in terms of the magnitude of the non-diagonal elements.     
\end{remark}

\section{Appendix B: An Approximation Lemma -- pruned matrix $W'$ approximates the deterministic matrix $S$}

Assume we are given a deterministic matrix $S$ and we add to it a random matrix $R$, for the $R$ and $S$ given in Example \ref{simple_case}. Suppose we take $W=S+R$; it is well known that one can find a rank $k$ approximation of $W$.  This is done by taking the SVD of $W=U\Sigma V^T$ and setting all but the top $k$ singular values in $\Sigma$ to zero. The following is a known theorem of this result:

\begin{thm}
    Given the singular value decomposition (SVD) of $W=U \Sigma V^T$, where $U$ and $V$ are unitary matrices and $\Sigma$ is a diagonal matrix containing the singular values of $W$, the rank $k$ approximation of $W$ is given by:

\begin{equation}
\tilde{W}_{k} = U_k \Sigma_k V_k^T
\end{equation}

where $U_k$ and $V_k$ are the matrices obtained by retaining only the first $k$ columns of $U$ and $V$, respectively, and $\Sigma_k$ is the matrix obtained by retaining only the first $k$ diagonal entries of $\Sigma$. Here we use $\tilde{W}_{k}$ to distinguish it from the weight layer matrix $W_k$. This approximation represents the best rank $k$ approximation to $W$ in the following sense:
\begin{equation}
    \tilde{W}_{k} = \operatorname{min}_{\text{rank}(X)=k} \| W - X \|_F
    \end{equation}

where $X$ is an arbitrary matrix of rank $k$ and $\| \cdot \|_F$ is the Frobenius norm.
\end{thm}

See \cite{demmel1986matrix} for more on this result. In this section, we wish to obtain slightly different results in a similar direction. We wish to show that for the $R$ and $S$ given in Example \ref{simple_case}, $\tilde{W}_{r}$ is a good approximation of $S$ (recall $S$ has $r$ non-zero singular values). That means that $W_r$ is a good approximation of the deterministic part of $W$. We, therefore, state a lemma that shows that: 
\begin{equation}
    \|(\tilde{W}_{r}-S)z\|<f_W\|z\|.
\end{equation}

Rather than state this lemma in terms of $\tilde{W}_{r}$, we state them in terms of $W'$, which is defined as follows:
\begin{defn}
\label{W'}
Take $W=R+S$, with $R$ and $S$ given in assumptions $1-3$ from Section \ref{assumptions_2}. Take $W=U\Sigma V^T$ to be the SVD of $W$ and take, for any $\epsilon$, \begin{equation}
\Sigma_{i,j}'= \begin{cases} 
\Sigma_{i,j} & \Sigma_{i,j}> \sqrt{\lambda_+}+\epsilon \\ 0 & \Sigma_{i,j}\leq \sqrt{\lambda_+}+\epsilon 
\end{cases}
\end{equation}

Then we obtain the \textit{truncated matrix} $W'$ by taking $W'=U\Sigma' V^T$.
    
\end{defn}

Next, we define a padded unitary matrix, which is helpful when proving the approximation theorem.

\begin{defn}[Padded Unitary Matrix]
\label{padded}
A padded-unitary matrix \( Q \) of size \( (n+m) \times (n+m) \) is defined as a matrix where the first \( n \times n \) submatrix is a unitary matrix, and the remaining entries are zeros. Formally, if \( U \) is an \( n \times n \) unitary matrix, then \( Q \) is constructed as
\[
Q = \begin{pmatrix}
U & \mathbf{0} \\
\mathbf{0} & \mathbf{0}
\end{pmatrix},
\]
where \( \mathbf{0} \) represents the appropriate-sized matrix with all zero entries.
\end{defn}

Next, we present an approximation lemma that gives a bound for $\|(W'-S)z\|_2$, where $z$ is any vector. This lemma describes how well $S$ is being approximated by $W'$.

\begin{aplem}
\label{approx_lemma_example}
Assume that $W=R+S$, with $R$ a $N \times M$ random matrix and $S$ a deterministic matrix satisfying assumptions $1-3$, for the assumptions in Section \ref{assumptions_2}. Assume that the singular values of $S$ are bigger than $\bar{\theta}$, given in assumption 3, and all singular values have multiplicity one. Take  $z$ to be any vector in $\R^n$. Then for $W'$ given in Def. \ref{W'} we have that $\exists f_W>0, \exists B^*_N(\epsilon)$ such that $\forall \epsilon>0$:

\begin{equation}
\label{approx_equation}
    \mathbb{P}\bigg(\|(W'-S)z\|_2 \geq (1+\epsilon) f_W \|z\|_2\bigg)<B^*_N(\epsilon),
\end{equation}
with $B^*_N(\epsilon) \to 0$ as $N \to \infty$.

Moreover, assuming that $\sigma_r$ is the smallest singular values of $S$, we have:  

 \begin{equation}  
  f_W= \max_{1\leq i \leq r} \sqrt{(g_{v_i}g_{\sigma_i}^2+\sigma^2_i)-2g_{\sigma_i}\sigma_i\sqrt{g_{v_i}}\sqrt{g_{u_i}}},
\end{equation}  
with $g_{\sigma_i}$, $g_{u_i}$ and $g_{v_i}$ given in assumption $3$.
\end{aplem}

\begin{remark}
    Here $f_W$ depends on the distribution of the eigenvalues of $R$ and on the eigenvalues of $S$.  
\end{remark}

\begin{ex}
    Assume that $W$ is the matrix given in Example \ref{simple_case}. Then 
    
    \begin{equation}
        f_W = \max_{1\leq i\leq r} \left\{ \sqrt{ \left(\sigma^2_i+\left(\frac{1+\sigma^2_i}{\sigma_i}\right)^2 \left( 1-\frac{1}{\sigma^2_i}\right)\right) - 2(1+\sigma^2_i) \left(1-\frac{1}{\sigma^2_i}\right)} \right\}.
         \end{equation}
\end{ex}

\begin{ex}

Assume that $W=R+S$, with $R$ a $N \times M$ random matrix satisfying assumptions $1'$ and $S$ a deterministic matrix satisfying assumption $2'$, for the assumptions in Section \ref{assumptions_2}. Assume that the singular values of $S$ are bigger than $\bar{\theta}$, given in \eqref{theta_bar}, and all singular values have multiplicity one.  Then for $W'$ given in Def. \ref{W'} we have that:

    \begin{align*}  
  f_W = \max_{1\leq i \leq r} \sqrt{(\frac{-2\phi_{\mu_R}(\rho)}{\sigma^2_{r}D'_{\Tilde{\mu_R}}(\rho)}D^{-1}_{\mu_R}(\frac{1}{(\sigma_i)^2})^2+\sigma^2_i)-2\sigma_i D^{-1}_{\mu_R}(\frac{1}{(\sigma_i)^2})\sqrt{\frac{-2\phi_{\mu_R}(\rho)}{\sigma^2_{r}D'_{\Tilde{\mu_R}}(\rho)}}\sqrt{\frac{-2\phi_{\mu_R}(\rho)}{\sigma^2_{r}D'_{\mu_R}(\rho)}}}.
\end{align*}

\end{ex}

\begin{proof}

  We prove this for the simple case when $S=\sum_{i=1}^r \sigma_i u_i v^T_i$ and for the example given in \ref{simple_case}, when $W$, $R$ and $S$ are $N \times N$ matrices. The proof for the more general case is the same.  

    Take $W'=U_W\Sigma'_WV_W^T$ and $S=U_S\Sigma'_SV^T_S$ to be the SVD of $W'$ and $S$, with $\Sigma'_S$ a $N \times N$ matrix with $r$ non-zero singular values  $\sigma_i$ on its diagonal and all other elements zero, and assume the smallest singular value of $S$, which is $\sigma_r$, is bigger than $1$. Furthermore, take $U_W, V_W$ to be $N \times N$ unitary matrices, and $U_S, V_S$ to be $N \times N$ padded unitary matrices, as defined in Def \ref{padded}, with the "singular vectors" corresponding to the zero singular values $\sigma_i$ being the zero vector. If we look at the SVD in terms of its summation representation, changing the singular vectors corresponding to the zero singular values to zero vectors (which are not unit vectors) will not change the reconstructed components of the matrix $S$. Let's clarify this:

The SVD of a matrix \( S \) can be written as:

\[ S = \sum_{i=1}^{r} \sigma_i u_i v_i^T \]
where \( \sigma_i \) are the singular values, \( u_i \) are the left singular vectors, and \( v_i \) are the right singular vectors, with \( r \) being the rank of \( S \). This sum runs over all singular values, including the zero singular values.

In this sum, each term \( \sigma_i u_i v_i^T \) contributes to the matrix \( S \). However, for terms where \( \sigma_i = 0 \), the entire term \( \sigma_i u_i v_i^T \) becomes a zero matrix, regardless of what \( u_i \) and \( v_i \) are. This is because multiplying by zero annihilates any contribution from these vectors.

Therefore, if you change the singular vectors corresponding to the zero singular values (let's say, replacing them with zero vectors), these terms in the sum still contribute nothing to \( S \) because they are multiplied by zero. The non-zero singular values and their corresponding vectors still determine the matrix \( S \), and the contributions from the terms with zero singular values remain zero. 

In the adapted form of the SVD of $S$ being considered, the matrices \( U \) and \( V \) do not strictly adhere to the traditional definition of unitary matrices. Instead, they can be described as padded unitary matrices.  For the purposes of the proof in question, this "almost" unitary nature of \( U \) and \( V \) is sufficient.

Then we have 
\begin{equation}
\begin{split}
\|(W'-S)z\|_2 & = \|(U_W\Sigma'_WV_W^T-U_S\Sigma'_SV^T_S)z\| \\
& \leq  \sqrt{\lambda_{\max}((U_W\Sigma'_WV_W^T-U_S\Sigma'_SV^T_S)^T(U_W\Sigma'_WV_W^T-U_S\Sigma'_SV^T_S))}\|z\|_2,
\end{split}
\end{equation}
with $\lambda_{\max}(A)$ the largest eigenvalue of $A$. Thus we obtain

\begin{equation}
  \|(W'-S)z\|_2\leq \sqrt{\lambda_{\max}((V_W\Sigma_W'^2V^T_W+V_S\Sigma_S'^2V^T_S-V_W\Sigma'_WU^T_WU_S\Sigma'_SV^T_S-V_S\Sigma'_SU^T_SU_W\Sigma'_WV^T_W))}\|z\|_2.  
\end{equation}

We can multiply the right and left of the matrix $(V_W\Sigma_W'^2V^T_W+V_S\Sigma_S'^2V^T_S-V_W\Sigma'_WU^T_WU_S\Sigma'_SV^T_S-V_S\Sigma'_SU^T_SU_W\Sigma'_WV^T_W)$ by the unitary matrices $V_W$ and $V^T_W$ respectively, without increasing the absolute value of the max eigenvalue, to obtain:  

\begin{equation}
  \|(W'-S)z\|_2\leq \sqrt{\lambda_{\max}(\Sigma_W'^2+V^T_W V_S\Sigma_S'^2V^T_SV_W-\Sigma'_WU^T_WU_S\Sigma'_SV^T_SV_W-V^T_WV_S\Sigma'_SU^T_SU_W\Sigma'_W)}\|z\|_2.  
\end{equation}

 By \eqref{cross_D_1}, \eqref{cross_D_2} and Theorem \ref{Gershgorin's Circle Theorem}, since the sum of the off diagonal elements of 

 \begin{equation}
     G:=(\Sigma_W'^2+V^T_W V_S\Sigma_S'^2V^T_SV_W-\Sigma'_WU^T_WU_S\Sigma'_SV^T_SV_W-V^T_WV_S\Sigma'_SU^T_SU_W\Sigma'_W)
 \end{equation}
can be made arbitrarily small with high probability as $N \to \infty$, $\exists B^{*}_N(\epsilon)$ so that for large enough $N$ we have: \begin{equation}
\begin{split}
\mathbb{P}\bigg(|\lambda_{\max}(G)|-\lambda_{\max}((\Sigma_W'^2+D^2_2\Sigma_S'^2)-2\Sigma'_W\Sigma'_S((D_2)D_1))|>\epsilon\bigg)  <  B^{*}_N(\epsilon),
\end{split}
\end{equation}
where $D_1$ is the diagonal matrix containing $<u_i,\tilde{u}_i>$ on its diagonal and all other elements zero and $D_2$ the diagonal matrix containing $<v_i,\tilde{v}_i>$  on its diagonal and all other elements zero. In fact, because $S$ only has $r$ non-zero singular values, we obtain less than $r \times r$ non-zero off-diagonal elements for the matrices $U^T_WU_S$, $U^T_SU_W$, $V^T_WV_S$, and $V^T_SV_W$. Thus, because $r$ is fixed  we have 
by Theorems \ref{singualr_values_simple_case}, \ref{singualr_vectors_simple_case} and Remark \ref{a.s_CIL}, that $\exists B^*_N(\epsilon)$ such that for large enough $N$:
\begin{align}
\mathbb{P}\bigg(&\left\|\,(W'-S)z\,\right\|_2 \geq (1+\epsilon)\max_{1\leq i\leq r} \bigg\{\sqrt{\begin{aligned}[t]
&\left(\sigma^2_i+\left(\frac{1+\sigma^2_i}{\sigma_i}\right)^2 \left( 1-\frac{1}{\sigma^2_i}\right)\right) \\
&- 2(1+\sigma^2_i) \left(1-\frac{1}{\sigma^2_i}\right)
\end{aligned}}\|z\|_2 \bigg\} \bigg) \leq B^{*}_N(\epsilon).
\end{align}

In fact, using the above argument, we can show that as $N \to \infty$:

\begin{equation}
    \|(W'-S)\| \xrightarrow[\text{}]{a.s.} \max_{1\leq i\leq r} \left\{ \sqrt{ \left(\sigma^2_i+\left(\frac{1+\sigma^2_i}{\sigma_i}\right)^2 \left( 1-\frac{1}{\sigma^2_i}\right)\right) - 2(1+\sigma^2_i) \left(1-\frac{1}{\sigma^2_i}\right)} \right\}.
\end{equation}

This completes the proof.

\end{proof}

\subsection{Numerics for Example \ref{example_main_result_1}}
\label{gass_example_1}

\begin{figure}[ht]
    
    \includegraphics[width=1.15\textwidth]{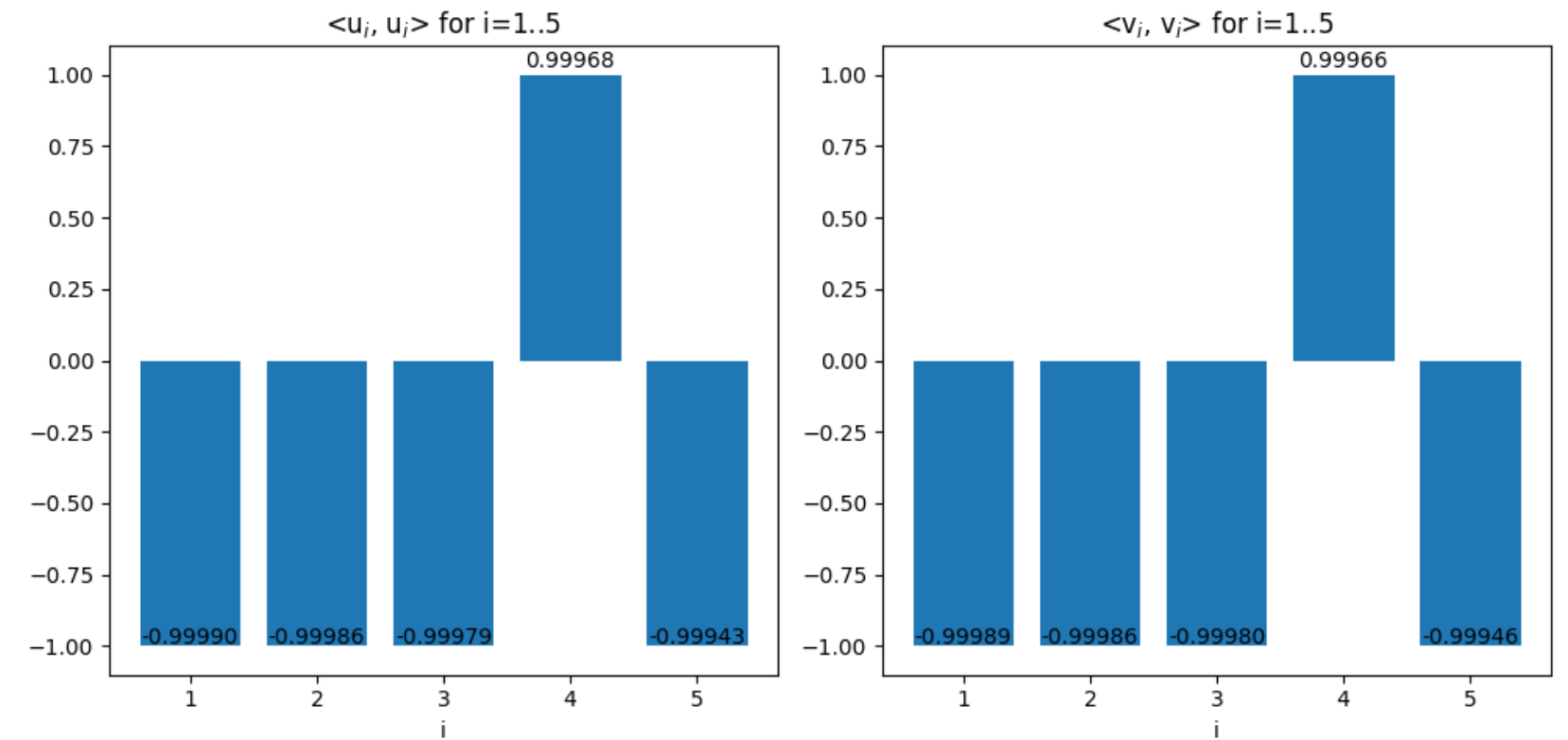}
    \caption{Left: Dot product of the left singular vectors. Right: Dot product of the right singular vectors. }
    \label{fig:singular_plots}
\end{figure}

In the following subsection, we provide a figure of the dot products of the $5$ left and right singular values for the matrices $W=R+S$ and $S$ described in Example \ref{example_main_result_1}, see Fig \ref{fig:singular_plots}. As mentioned earlier, the $5$ singular values of $S$ were $30, 40, 50, 60, 70$. We see that the dot product of the left and right singular vectors of $W$ and $S$ can be approximated almost perfectly by the equation $\sqrt{1-\frac{1}{\sigma^2_i}}$, see \eqref{singualr_value_approx}. That is, for 
$\sigma_5=30$, we have $\sqrt{1-\frac{1}{\sigma^2_i}}\approx 0.99944$ and indeed $ <u_5,u'_5> \approx 0.99943$ and similarly for the other dot products.

\subsection{Details for Example \ref{Example_2_main}}
\label{Example_2_main_num}

The details provided in the subsection were taken from \cite{benaych2011eigenvalues2}. Under the setting given in Example \ref{Example_2_main}, let \( u_i' \) be a unit-norm eigenvector of \( R + S \) associated with its r largest eigenvalues. We have  for \( 1 \leq i \leq r \),

\[
\lambda_i(R + S) \overset{\text{a.s.}}{\longrightarrow}
\begin{cases} 
\sigma_i + \frac{\sigma^2}{\sigma_i} & \text{if } \sigma_i > \sigma, \\
2\sigma & \text{otherwise},
\end{cases}
\]
as \( n \rightarrow \infty \). 

We also have
\[
|\langle u, u' \rangle|^2 \overset{\text{a.s.}}{\longrightarrow}
\begin{cases} 
1 - \frac{\sigma^2}{\sigma_i^2} & \text{if } \sigma_i > \sigma, \\
0 & \text{otherwise}.
\end{cases}
\]

\section{Appendix C: Proof for Pruning Theorem}

The proof of the Pruning Theorem \ref{the_main_result_RMT} consists of two steps. First, we show how classification confidence changes when the weight matrix $W_b$ is replaced with its deterministic part $S_b$. Second, we approximate the deterministic matrix $S_b$ with the pruned matrix $W'_b$, keeping track of the corresponding change in the classification confidence. The key idea in the second step is to use asymptotics of the spectrum of deformed matrices $W=R+S$ with a finite number of singular values of $S$, see \cite{benaych2011eigenvalues}, in combination with GCT.  

\subsection{Proof for key technical Lemma \ref{main_result_remove _R}}
\label{proof_main_result_remove_random}

\begin{proof}

We first provide a proof for the case when our DNN has only one layer matrix $W$.

We proceed by showing that if $\|s\|_2$ is independent of $N$ then as $N \to \infty$, $\delta X(s,\alpha_W) - \delta X(s,\alpha_S)=0$ in probability, with $\alpha_W$ the parameters of the DNN with weight layer $W$ and $\alpha_S$ the parameters of the same DNN but with weight layer $S$. This means that the random matrix $R$ does not improve the accuracy of the DNN as $N \to \infty$.

More generally, we also show that:

\begin{equation}     
 \mathbb{P}\bigg(\delta X(s,\alpha_S) \geq 0 \mid \delta X(s,\alpha_W) \geq  a(N))\|s\|_2\bigg) \geq 1- \frac {1}{N^\frac{1}{4}},
   \end{equation}
with $a(N)=\frac{2}{N^{\frac{1.5}{4}}}$.

We start by approximating the components of $Rs$ and showing that they are small and go to $0$ as $N \to \infty$.  When $R$ is a random matrix with components taken from i.i.d with mean $0$ variance $\frac{1}{N}$, we have that $(Rs)_m$ is a random variable taken from a  distribution with mean $0$ and variance $\sigma_{(Rs)_m}^2=\sum_{i=1}^M s^2_i \frac{1}{N}=\|s\|_2^2\frac{1}{N}$. Thus, $(Rs)_m$ is also a random variable with $0$ mean and variance $\|s\|_2^2\frac{1}{N}$. 

Then, using Chebyshev's inequality (see Theorem \ref{Chebyshev's inequality}) and taking $k=N^{\frac{.5}{4}}$ we obtain 

\begin{equation}
\label{Chebyshev's inequality_equation}
    \Pr\bigg(|(Rs)_m|\geq \frac{1}{N^{\frac{1.5}{4}}} \|s\|_2\bigg) \leq \frac {1}{N^\frac{1}{4}}.
\end{equation}

Thus, given that:
\begin{equation}
\begin{aligned}
\delta X(s,\alpha_W) -\delta X(s,\alpha_S) &= |(R+S)s_{i(s)}|-|\max_{j\neq i(s)}((S+R)s)_j|- |(S)s_{i(s)}|+|\max_{j\neq i(s)}(Ss)_j| \\
&\leq |(R)s_{i(s)}|-|\max_{j\neq i(s)}((S+R)s)_j|+|\max_{j\neq i(s)}(Ss)_j| \\
\end{aligned}
\end{equation}

By \eqref{Chebyshev's inequality_equation}, we have  

\begin{equation}   
        \label{Eq_used}
\mathbb{P}\bigg((\delta X(s,\alpha_W) -\delta X(s,\alpha_S))\leq  \frac{1}{N^{\frac{1.5}{4}}}\|s\|_2-|\max_{j\neq i(s)}((S+R)s)_j| +|\max_{j\neq i(s)}(Ss)_j|\bigg)\geq 1- \frac {1}{N^\frac{1}{4}}.
\end{equation}

Suppose $|\max_{j\neq i(s)}(Ss)_j|$ is satisfied for the component $k^*(N)$, meaning that:

\begin{equation}
    |\max_{j\neq i(s)}(Ss)_j| =|(Ss)_{k^*(N)}|.
\end{equation}

Then, again by \eqref{Chebyshev's inequality_equation} we have: 

   \begin{equation} 
   \label{CI_2}\mathbb{P}\bigg(||(S+R)s)_{k^*}|-|(Ss)_{k^*}|| \leq \frac{1}{N^{\frac{1.5}{4}}}\|s\|_2\bigg) \geq 1- \frac{1}{N^{\frac{1}{4}}},
\end{equation}
given that $S$ and $R$ are independent from each other and $S$ is deterministic. Given that $\max_{j\neq i(s)}(R+S)s_j\geq (R+S)s_{k^*}$, from \eqref{Eq_used} 
 we obtain 
    
            \begin{equation}   
\mathbb{P}\bigg((\delta X(s,\alpha_W) -\delta X(s,\alpha_S))\leq \frac{1}{N^{\frac{1.5}{4}}}\|s\|_2-|(S+R)s)_{k^*}|+|(Ss)_{k^*}|\bigg)\geq 1- \frac {1}{N^\frac{1}{4}}.
\end{equation}

    \textbf{Final Result from \eqref{CI_2}}:
    \begin{equation}
        \mathbb{P}\bigg(
        \delta X(s,\alpha_{W}) - \delta X(s,\alpha_{S})\leq \frac{2}{N^{\frac{1.5}{4}}}\|s\|_2\bigg) \geq 1 - \frac{1}{N^{\frac{1}{4}}}
    \end{equation}
with $a(N)=\frac{2}{N^{\frac{1.5}{4}}}$.

When the DNN has more than one layer, we continue this proof as follows:

\begin{align}
Z=\mathbb{P} \left( \left| \lambda \circ W_4 \circ \lambda \circ W_3 \circ \lambda \circ (R + S) \circ \lambda \circ W_1 s - \lambda \circ W_4 \circ \lambda \circ W_3 \circ \lambda \circ S \circ \lambda \circ W_1 s \right|_i > t \right)
\end{align}

By the triangle inequality, we have

\begin{align}
Z\leq\mathbb{P} \left( \left|  W_4 \circ \lambda \circ W_3 \circ \lambda \circ (R + S) \circ \lambda \circ W_1 s -  W_4 \circ \lambda \circ W_3 \circ \lambda \circ S \circ \lambda \circ W_1 s \right|_i > t \right)
\end{align}

We factor out the common term \( W_4 \)  and use the inequality $\|Av\|_{\infty}\leq \|A\|_1\|v\|_{\infty} $ to obtain:

\begin{align}
Z\leq \mathbb{P} \left( \left\|  W_4   \right\|_{1} \cdot \left\|   W_3 \circ \lambda \circ (R + S) \circ \lambda \circ W_1 s - W_3 \circ \lambda \circ S \circ \lambda \circ W_1 s \right\|_{\infty} > t \right)
\end{align}

Next, we apply the inequality  $\|Av\|_{\infty}\leq \|A\|_1\|v\|_{\infty} $ on \( W_3 \) and $\lambda \circ (R + S) \circ \lambda \circ W_1 s -  \lambda \circ S \circ \lambda \circ W_1 s$ together with the triangle inequality, we get:

\begin{align}
Z\leq \mathbb{P} \left( \left\|  W_4   \right\|_1 \cdot \left( \left\| W_3 \right\|_1 \cdot \left\| R  W_1 s  \right\|_\infty \right)  > t \right),
\end{align}
given that  $\|\lambda \circ (R + S) - \lambda \circ S||\leq \|R\|$. Thus we have,

\begin{align}
Z\leq \mathbb{P} \left( \left\|  W_4   \right\|_1 \cdot \left( \left\| W_3 \right\|_1 \cdot  \max_{i,j}|R_{i,j} |\left\|   W_1 s  \right\|_1 \right)  > t \right),
\end{align}

If \( R \) is a random matrix with i.i.d entries taken from the normal distribution \( N(0, 1/N) \), then using the concentration inequality in Theorem \ref{Borell-TIS Inequality}, we get:

If we choose \( t = \frac{\left\|  W_4 \right\|_1 \left\| W_3 \right\|_1 \left\| W_1 s \right\|_{1}}{N^{1.5/4}} \):

\[
Z\leq \mathbb{P} \left(  \max_{i,j}|R_{i,j} |> \frac{1}{N^{1.5/4}} +\sqrt{\frac{2 \log N^2}{N}}\right) \leq 2 \exp \left( - \frac{N^{1/4}}{2} \right)
\].

Thus, we have:

\begin{align}
\mathbb{P}  \bigg(\delta X(s, \alpha_{S_2}) \geq 0 \mid \delta X(s, \alpha_{W_2}) > 2\frac{\left\|  W_4 \right\|_1 \left\| W_3 \right\|_1 \left\| W_1 s \right\|_{1}}{N^{1.5/4}} + 2\sqrt{\frac{2 \log N^2}{N}}\bigg)\leq 2 \exp \left( - \frac{N^{1/4}}{2}\right).
\end{align}

For a DNN with more layers and a different $R$, the steps in this proof would be the same and would use similar concentration inequalities to bound $\max_{i,j}|R_{i,j} |$.

\end{proof}

\textbf{Proof for the Pruning Theorem \ref{the_main_result_RMT}:}
\label{proof_main_result_RMT}

\begin{proof}
We first provide a proof for the case when our DNN has only one layer matrix $W$.

By Lemma \ref{main_result_remove _R} we have that there $\exists D(N)$ such that for

   \begin{equation}
       E:=a(N) g_{\phi}(s,b) 
   \end{equation}
   we have the conditional probability
   \begin{equation}
   \mathbb{P}(\delta X(s,\alpha_{S})\geq 0  \mid \delta X(s,\alpha_{W}) \geq  E)\geq 1-D(N), 
   \end{equation}
 with $D(N), a(N) \to 0$ as $N \to \infty$ and $g_{\phi}(s,b)$ coming from \eqref{from_DNN_st}. Again, $\alpha_{S}$ are the parameters of the DNN with the weight matrix $S$ and  $\alpha_{W}$ are the parameters of the DNN with the weight layer matrix $W$. 
       
We then use Approximation Lemma \ref{approx_lemma_example} to obtain that $\exists f_W, \exists B^*_N(\epsilon)$ such that $\forall \epsilon$:

\begin{equation}
\label{approx_equation_used}
    \mathbb{P}(\|(W'-S)z\|_2 > (1+\epsilon) f_W \|z\|_2)<B^*_N(\epsilon),
\end{equation}
with $B^*_N(\epsilon) \to 0$ as $N \to \infty$. 
 
We then follow an argument similar to that given for Lemma \ref{proof_main_result_1_new}. That is, the change in classification confidence due to pruning is:
\[ \Delta(\delta X) = |\delta X(s,\alpha_S) - \delta X(s,\alpha_{W'})| \]

For a particular component \( i \), the change \( \Delta X_i \) due to pruning is given by:
\[ \Delta X_i = |X_i(s,\alpha_S) - X_i(s,\alpha_{W'})| \]

We have that $ X(s,\alpha_S)=\lambda \circ (W)s$ and $ X(s,\alpha_{W'})=\lambda \circ (W')s$.

Given that $\lambda$ is the absolute value activation function, for any scalar values $x$ and $y$, we have:

\begin{equation}
|\lambda(x) - \lambda(y)| \leq |x - y|.
\end{equation}
Thus, from \eqref{approx_equation_used} we have $\| X(s,\alpha_S)- X(s,\alpha_{W'})\| \leq \| (S-W')s \| \leq (1+\epsilon)f_W\|s\|_2$, with probability $1-B^*_N(\epsilon)$. Furthermore, 

\begin{equation}
\label{change_together_2}
    \Delta X_{\text{max}}+\Delta X_{\text{max-1}} \leq \sqrt{2}(1+\epsilon)f_W\|s\|_2,
\end{equation}
with probability $1-B^*_N(\epsilon)$.
Here $\Delta X_{\text{max-1}}$ is the change in the component of $X$ which has the second to biggest change, and  $\Delta X_{\text{max}}$ is the change in the change in the component of $X$ which had the biggest change.   

Then, using the same steps given in Lemma \ref{proof_main_result_1_new}, we obtain:

\begin{equation}
     \Delta(\delta X) \leq     \Delta X_{\text{max}}+\Delta X_{\text{max-1}}  \leq  \sqrt{2}(1+\epsilon)f_W\|s\|_2,
\end{equation}
with probability $1-B^*_N(\epsilon)$.

Thus, given that: \begin{equation}
   \mathbb{P}(\delta X(s,\alpha_{S})\geq 0  \mid \delta X(s,\alpha_{W}) \geq  E)\geq 1-D(N), 
   \end{equation}
   we have that there exists an explicit function $f_{W}>0$ such that $\forall \epsilon>0$   there $\exists C_\epsilon(N)$ such that for 
     
    \begin{equation} 
    \label{main_equation}
E':=(\sqrt{2}(1+\epsilon)\min\{f_{W},\sqrt{\lambda_+}\}+a(N)) g_{\phi}(s,b)
   \end{equation}
   we have the conditional probability
   \begin{equation}
   \mathbb{P}(\delta X(s,\alpha_{W'})\geq 0 \mid \delta X(s,\alpha_{W}) \geq  E')\geq 1-C_\epsilon(N), 
   \end{equation}
with $C_\epsilon(N), a(N) \to 0$ as $N \to \infty$  and $g_{\phi}(s,b)$ coming from \eqref{from_DNN_st}.  $f_{W}$ is given in Lemma \ref{approx_lemma_example}.

   When the DNN has more than one layer, we continue this proof with the same steps found in Subsection \ref{proof_main_result_1_new}.
\end{proof}  

\begin{thm}[Chebyshev's inequality]
\label{Chebyshev's inequality}
    Let $X$  be a random variable with mean $\mu$ and finite non-zero variance $\sigma^2$. Then for any real number $k > 0$,

\begin{equation}
\Pr(|X-\mu |\geq k\sigma )\leq {\frac {1}{k^{2}}}.
\end{equation}
\end{thm}

\subsection*{Borell-TIS Inequality for i.i.d. Gaussian Variables}

\begin{thm}[Borell-TIS Inequality]
\label{Borell-TIS Inequality}
Let \( X_1, X_2, \ldots, X_n \) be i.i.d. centered Gaussian random variables with \( X_i \sim N(0, \sigma^2) \). Set \( s_X^2 := \max_{i=1,\ldots,n} \mathbb{E}(X_i^2) = \sigma^2 \). Then for each \( t > 0 \):

\[
P\left( \max_{i=1, \ldots, n} |X_i| - \mathbb{E}\left[ \max_{i=1, \ldots, n} X_i \right] > t \right) \leq \exp\left( -\frac{t^2}{2s_X^2} \right).
\]

For the absolute value bound:

\[
P\left( |\max_{i=1, \ldots, n} |X_i| - \mathbb{E}\left[ \max_{i=1, \ldots, n} X_i \right]| > t \right) \leq 2 \exp\left( -\frac{t^2}{2s_X^2} \right).
\]

In particular, if \( X_i \sim N(0, \frac{1}{N}) \):

\[
s_X^2 = \frac{1}{N},
\]

and for any \( t > 0 \):

\[
P\left( \max_{i=1, \ldots, n} |X_i| > \sqrt{\frac{2 \log n}{N}} + t \right) \leq 2 \exp\left( -\frac{N t^2}{2} \right).
\]
\end{thm}

\subsection{Proof of Pruning Theorem for accuracy}
\label{RMT_corollary_proof}

This Theorem follows from the Pruning Theorem \ref{the_main_result_RMT}. Under the assumptions of this theorem and from the Pruning Theorem, and assuming that $R_2$ is i.i.d. from $N(0,1/N)$ we have that $\forall \epsilon>0, \forall s \in T'$:

   \begin{equation}
   \label{CC_used_in_co_1}\mathbb{P}\bigg(\delta X(s,\alpha_{W'})\geq 0 \mid  \delta X(s,\alpha_W)\geq  (1+\epsilon)(\sqrt{2}(f_{W_2}\sigma_{\max}(W_3)+2N^{-\frac{1.5}{4}}\|W_3\|_1))\max_{s\in G_{E_{acc},\alpha}} \|W_1s\|_1+b(N)\bigg)\geq 1-C_{\epsilon}(N),
   \end{equation}
    with $C_{\epsilon}(N),b(N) \to 0$ as $N \to \infty$ and $f_{W_2}$ given by the Approximation Lemma \ref{approx_lemma_example}. In this case, $E'$ from the Pruning Theorem given in \eqref{pruning_equation_2} is equal to $(1+\epsilon)(\sqrt{2}(f_{W_2}\sigma_{\max}(W_3)+2N^{-\frac{1.5}{4}}\|W_3\|_1))\max_{s\in G_{E_{acc},\alpha}} \|W_1s\|_1+b(N)$.  By taking
\begin{equation}
     E_{acc}=(1+\epsilon)(\sqrt{2}(f_{W_2}\sigma_{\max}(W_3)+2N^{-\frac{1.5}{4}}\|W_3\|_1))\max_{s\in G_{E_{acc},\alpha}} \|W_1s\|_1 +b(N),
\end{equation}
we make the classification confidence threshold of the non-pruned DNN independent on $s$. We then obtain, from \eqref{CC_used_in_co_1}, that for all $s$:

   \begin{equation}
   \label{CC_used_in_co_2}\mathbb{P}\bigg(\delta X(s,\alpha_{W'})\geq 0 \mid  \delta X(s,\alpha_W)\geq  E_{acc}\bigg)\geq 1-C_{\epsilon}(N).
   \end{equation}
   
Given that this is independent on $s$, we obtain the final result: 

   \begin{equation}
   \label{main_eq_1}
    \mathbb{P}\bigg( G_{E_{acc},\alpha} \subset G_{0,\alpha'}   \bigg)\geq (1-C_{\epsilon}(N))^{|G_{E_{acc},\alpha}|}.
    \end{equation}
The theorem then follows from the fact that $acc_{\alpha'}(t)=\frac{|G_{0,\alpha'}|}{|T'|}$.

\section{Appendix D: Other algorithms required for implementing RMT-SVD based pruning of DNN}
\label{numerics}

\subsection{BEMA algorithm for finding $\lambda_+$}
\label{finding_lambda}

The following is the BEMA algorithm for finding the best fit $\lambda_+$ of $\frac{1}{N}R^TR$ based on the ESD of $X=R+S$.  It is used in the analysis of matrices with the information plus noise structure (i.e., deformed matrices), where one wants to determine the rightmost edge of the compact support of the MP distribution. In this context, the Tracy-Widom distribution provides the limiting distribution of the largest eigenvalue \(\lambda_+\) of large random matrices, allowing us to compute a confidence interval for \(\lambda_+\) in the presence of the Marčenko-Pastur distribution, see \cite{ke2021estimation}.
 The BEMA algorithm is computationally efficient and has been shown to provide accurate results for matrices with the information plus noise structure.  More details on the algorithm and its relationship with the MP and Tracy-Widom distributions can be found in \cite{ke2021estimation}. Here, we present a simplified version of the algorithm for $R$ a $N \times N$ matrix: 

\begin{algorithm}
\caption{Computation of $\lambda_+$ using MP and Tracy-Widom Distributions}
\begin{algorithmic}[1]
\State Choose parameters $\alpha \in (0, 1/2), \beta \in (0,1)$.
\For{each $\alpha N \leq k \leq (1-\alpha)N$}
    \State Obtain $q_k$, the $(k/N)$ upper-quantile of the MP distribution with $\sigma^2=1$ and $c = 1$.
    \Comment{Each $q_k$ is a solution to $\int\limits_{0}^{q_k} \frac{1}{2\pi}\frac{\sqrt{(4-\lambda)\lambda}}{\lambda} = k/N$}
\EndFor
\State Compute $\hat{\sigma}^2 = \frac{\sum_{\alpha N \leq k \leq (1-\alpha)N}q_k \lambda_k}{\sum_{\alpha N \leq k \leq (1-\alpha)N}q_k^2}$.
\Comment{where $\lambda_k$ is the $k^{th}$ smallest eigenvalue of $X$}
\State Obtain $t_{1-\beta}$, the $(1-\beta)$ quantile of Tracy-Widom distribution.
\State Return $\lambda_+ = \hat{\sigma}^2[4+2^{4/3}t_{1-\beta}\cdot N^{-2/3}]$.
\end{algorithmic}
\end{algorithm}

    \begin{remark}
   
The algorithm depends on parameters $\alpha \in (0,1/2), \beta \in (0,1)$. We show this by varying $\alpha$ and $\beta$ for the case found in Example \ref{deterministic_random}. See Fig. \ref{dep_alpha} and \ref{dep_beta}. The red  line is $\lambda_+ = 4 $, which is the correct $\lambda_+$ of $\frac{1}{N}R^TR$. In this example, while dependence on $\alpha$ is insignificant for sufficiently large values, dependence on $\beta$ allows us to control the confidence that the eigenvalues of the random matrix $R$ will be smaller than the estimator for $\lambda_+$ of the MP distribution. In all the numerical simulations given in Section \ref{num_results}, we took $\alpha=.1$ and $\beta=.1$. It would be interesting to try and see what happens when we take a larger $\beta$, as it would prevent the algorithm from pruning too many parameters but might lead to even higher accuracy.  

\begin{figure}
     \begin{subfigure}{0.4\textwidth}
  \includegraphics[width=\textwidth]{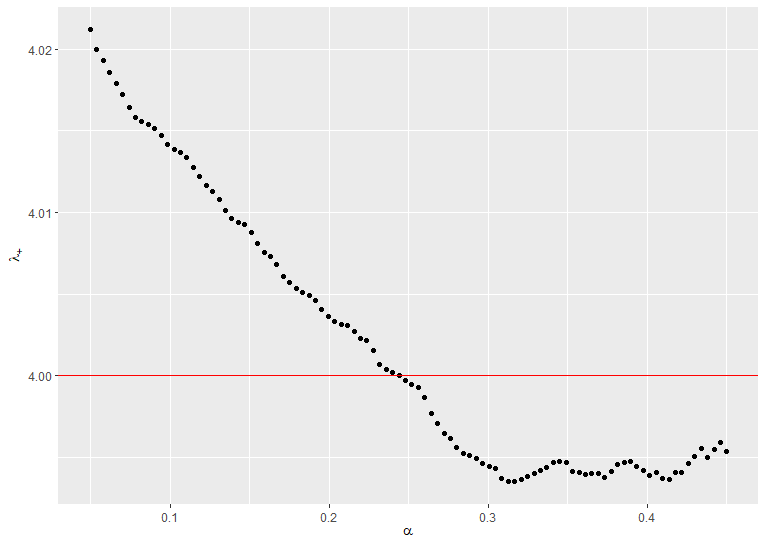}
    \caption{Dependence of algorithm the choice of $\alpha$, $\beta = 0.5$. In this example, the rank of the deterministic matrix $S$ is fairly low.}
    \label{dep_alpha}
     \end{subfigure}
     \hfill
     \begin{subfigure}{0.4\textwidth}
  \includegraphics[width=\textwidth]{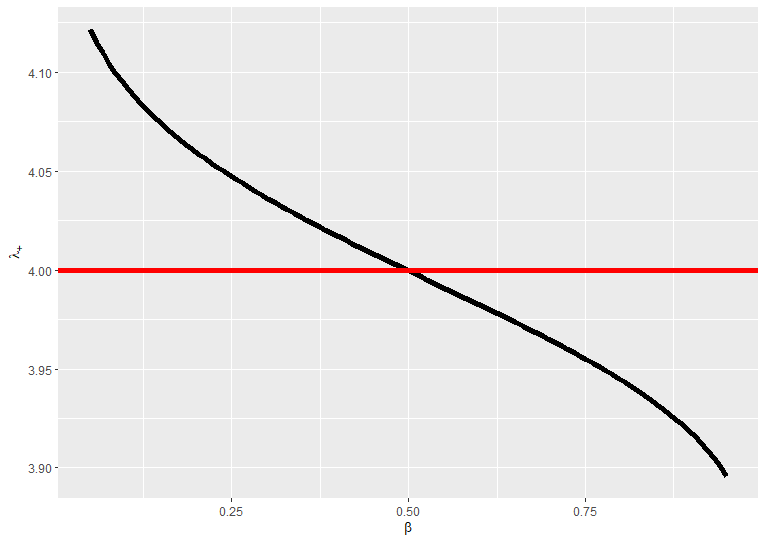}
    \caption{Dependence of algorithm on the choice of $\beta$, $\alpha = 0.25$.}
    \label{dep_beta}
     \end{subfigure}
     \end{figure}

    \end{remark}

    In the numeric portion of the paper, we always divide $R^TR$ by $\frac{1}{N}$ when obtaining the ESD of $X$ regardless of the original distribution of the initial random matrix $R(0)$. If $R(o)$ is distributed using $N(0,\frac{1}{N})$, dividing $R^TR$ by $N$ does not seem to change the fact that the ESD of $X$ is given by the MP distribution.

\begin{ex}\label{deterministic_random}
  In this example, we create a random $N\times N$ matrix $R$  with components taken from i.i.ds using the normal distribution of zero mean and unit variance ($\sigma^2=1$).  We take $S$ to be a  $N\times N$ deterministic matrix  with components given by
  
  \begin{equation}
  S[i,j]=\tan(\frac{\pi}{2}+\frac{1}{j+1})+\cos(i)\cdot\log(i+j+1)+\sin(j)\cdot\cos(\frac{i}{j}),
  \end{equation}
      $W=R+S$ and $X = \frac{1}{N}W^T W$. The BEMA algorithm is used to find the $\lambda_+$ of the ESD of $X$, as described in Subsection \ref{finding_lambda}. $R$ is a random matrix satisfying the conditions of Theorem \ref{RMT_MP_theorem}, and so the ESD of $\frac{1}{N}R^TR$ converges to the Marchenko-Pastur distribution as $N \to \infty$ and has a $\lambda_+$ that determines the rightmost edge of its compact support. We can imagine a situation in which $R$ is not directly known, and the goal is to find an estimator of $\lambda_+$ from the ESD of $X$. See Fig. \ref{fig:ESD_X&MP} for the result of the ESD of $X$ with the Marchenko-Pastur distribution that best fits the ESD shown in red. 
 \begin{figure}[h!]
	\centering	\includegraphics[width=.8\textwidth]{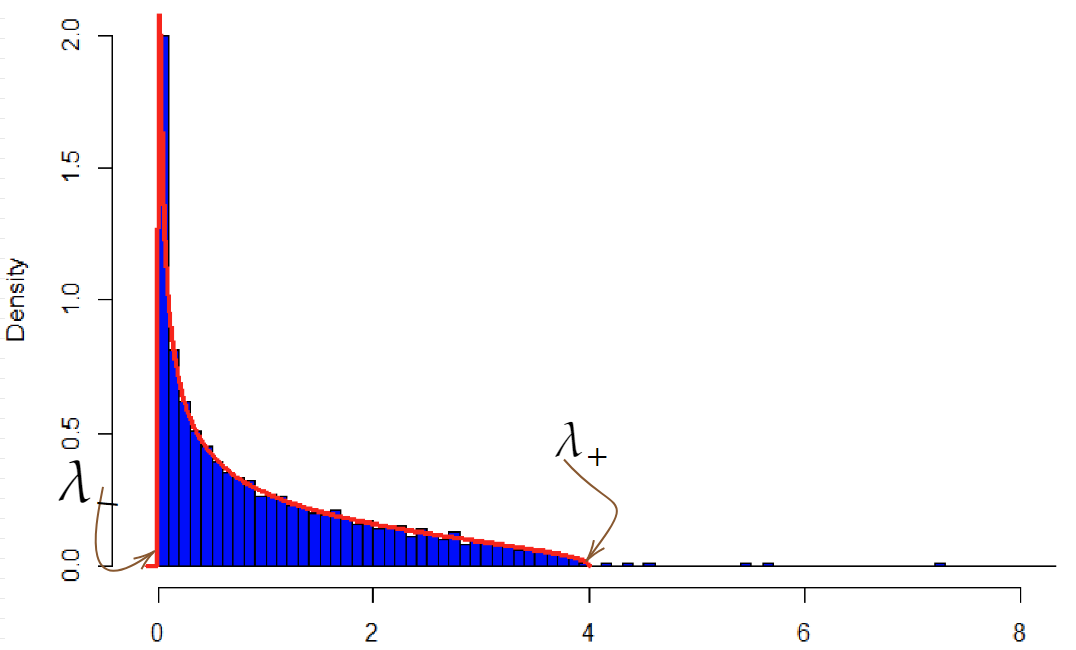}
			
		\caption{In blue, we have the ESD of X; in red, the Marchenko-Pastur distribution, which best fits the ESD based on the BEMA algorithm.}
		\label{fig:ESD_X&MP}
\end{figure}

 The bulk of the eigenvalues are well-fit by the MP distribution, but some eigenvalues bleed out to the right of $\lambda_+$. These eigenvalues correspond to the singular values of $S$. The direct calculation of the $\lambda_+$ of the MP distribution corresponding to $\frac{1}{N}R^TR$ gives $\lambda_+=\sigma^2\cdot(1+1)^2=4$, and the $\lambda_+$ obtained to fit the bulk of the ESD of $X$ and the $\lambda_+$ of $\frac{1}{N}R^TR$ are approximately the same.

\end{ex}

The BEMA algorithm will be employed to estimate $\lambda_+$ from the ESD of $X_l(t)$. As the DNN training progresses, it is expected that the majority of the eigenvalues of $X_l(t)$ will conform to the MP distribution. Nonetheless, some eigenvalues may extend beyond the bulk of the MP distribution and be associated with the singular values of $S_l(t)$. The purpose of the BEMA algorithm is to identify the furthest edge of the MP distribution, which helps determine the value of $\lambda_+$. Understanding $\lambda_+$ is crucial as it offers insights into the DNN's behavior during training and its capacity to generalize to new data.

In combination with the SVD, the BEMA algorithm can be applied to decide which singular values of the DNN's weight matrices $W_l$ should be eliminated during training. The SVD breaks down the weight matrix into its singular values and singular vectors, allowing for an RMT-based analysis of their distribution. Utilizing the BEMA algorithm, one can pinpoint the eigenvalues associated with the singular values of $S_l$ and differentiate them from the eigenvalues related to the singular values of $R_l$. By removing the eigenvalues corresponding to $R_l$, the DNN's training process can be made more effective and efficient.

\subsection{The role of singular value decomposition in deep learning}

Consider a $N \times M$ matrix $A$. A singular value decomposition (SVD) of $A$ consists of a factorization $A = U\Sigma V^T$, where:

\begin{itemize}
\item $U$ is an $N \times N$ orthogonal matrix.
\item $V$ is an $M \times M$ orthogonal matrix.
\item $\Sigma$ is an $N \times M$ matrix with the $i$th diagonal entry equal to the $i$th singular value $\sigma_i$ and all other entries of $\Sigma$ being zero.
\end{itemize}

For $\lambda_i$, the eigenvalues of a matrix $X = W^T W$, the singular values of $W$ are given by $\sigma_i = \sqrt{\lambda_i}$. Consequently, singular values are connected to eigenvalues of the symmetrization of a matrix $W$.

For a DNN's $W_l$, it has been demonstrated that discarding small singular values of $W_l$ through its SVD during the DNN's training can decrease the number of parameters while improving accuracy, as shown in \cite{yang2020learning,xue2013restructuring,cai2014fast,anhao2016svd}. In the remainder of this work, we illustrate how RMT can aid in identifying the singular values to be removed from a DNN layer without compromising the DNN's accuracy.

In particular, the BEMA algorithm can be combined with the SVD of $W_l$ to ascertain which singular values should be removed during the DNN's training. To achieve this, one first computes the SVD of $W_l$ and then calculates the eigenvalues of the symmetrized matrix $X_l = \frac{1}{N}W_l^T W_l$. The eigenvalues derived from the symmetrization can be linked to the singular values of $W_l$ through $N\lambda_i = \sigma_i^2$. Employing the BEMA algorithm to estimate the value of $\lambda_+$ allows for the determination of a threshold for the singular values of $W_l$. Singular values below the threshold can be removed without impacting the DNN's accuracy, as they are likely less crucial for the DNN's performance. This process can be carried out iteratively during the DNN's training since the threshold can be updated as training advances.

	 \subsection{Eliminating singular values while preserving accuracy}
\label{stable_acc}

\begin{figure}%
    \centering
    \subfloat[\centering Full Empirical Density]{{\includegraphics[width=6.5cm]{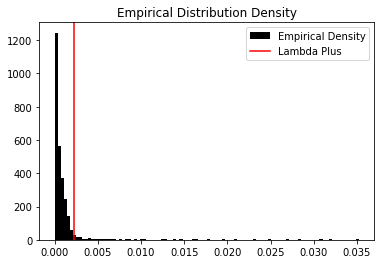} }}%
    \qquad
    \subfloat[\centering Zoomed Density ]{{\includegraphics[width=6.5cm]{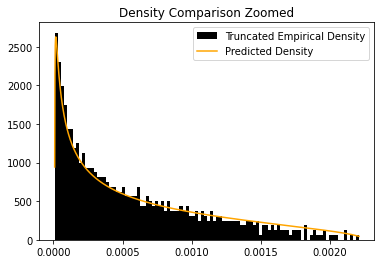} }}
    \caption{The ESD of $X_l$ and its best fit MP distribution}
    \label{MPdis1}%
\end{figure}

In this subsection, we demonstrate how SVD can be employed to remove the random components of $W_l$ without compromising accuracy. This could potentially lead to a significant reduction in the number of parameters in the DNN, resulting in faster training.

\begin{algorithm}
\caption{Pruning a Weight Matrix from a Trained DNN}
\begin{algorithmic}[1]
\State Acquire a weight matrix $W_l$ from a trained DNN.
\State Perform SVD on $W_l$: $W_l = U \Sigma V^T$.
\State Calculate the eigenvalues $\lambda_i$ of the square matrix $\frac{1}{N} W_l^T W_l$.
\State Apply the BEMA algorithm from Subsection \ref{finding_lambda} to find the best fit MP distribution for the ESD of $X = \frac{1}{N} W_l^T W_l$ and its corresponding $\lambda_+$.
\Comment{See Fig. \ref{MPdis1}}
\State Determine whether the ESD of $X$ fits the MP distribution using the algorithm in Subsection \ref{Conformance_Assessment}.
\Comment{Ensures $W_l=R+S$ assumption is valid}
\State Replace a portion, e.g., 0.1, of the singular values less than $\sqrt{\lambda_+N}$ with zeros to form a new diagonal matrix $\Sigma'$ and the truncated matrix $W'_l$.
\State Use $\Sigma'$ to obtain $W'_{1,l}=U\sqrt{\Sigma'}$ and $W'_{2,l}=\sqrt{\Sigma'}V^T$.
\end{algorithmic}
\end{algorithm}

\begin{ex}
    Consider an original DNN with two hidden layers, each consisting of 10 nodes. The total number of parameters in this case would be 100. By employing SVD and removing $8$ small singular values in the weight layer matrix of this DNN, we can split the hidden layer into two, resulting in a new DNN with three hidden layers. The first layer will have 10 nodes, the second layer will have 2 nodes, and the third layer will have 10 nodes. By keeping only two singular values in the SVD, we now have only 20 parameters; see Figure \ref{fig:dnn_split}. In practice, we don't actually split the layer. 

  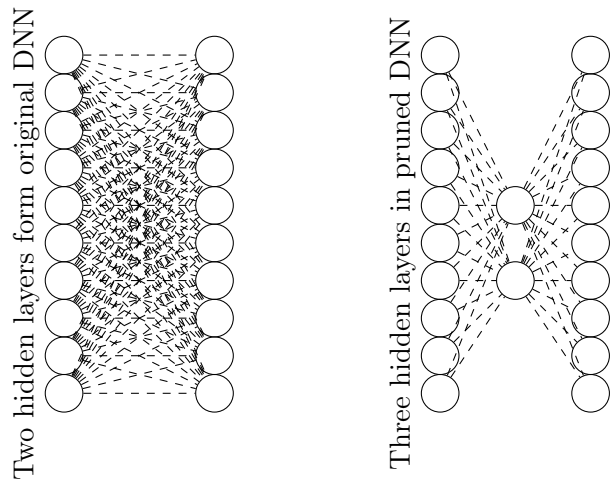
\begin{figure}[h]
\centering
\begin{tikzpicture}
\foreach \i in {1,2,3,4,5,6,7,8,9,10} {
\node[draw, circle, minimum size=0.5cm] (A\i) at (1, 0.5*\i) {};
\node[draw, circle, minimum size=0.5cm] (B\i) at (3, 0.5*\i) {};
}
\foreach \i in {1,2,3,4,5,6,7,8,9,10} {
\foreach \j in {1,2,3,4,5,6,7,8,9,10} {
\draw[dashed] (A\i) -- (B\j);
}
}
\node[rotate=90] at (0.5, 2.5) {Two hidden layers form original DNN};
\begin{scope}[xshift=5cm]
    \foreach \i in {1,2,3,4,5,6,7,8,9,10} {
        \node[draw, circle, minimum size=0.5cm] (C\i) at (1, 0.5*\i) {};
        \node[draw, circle, minimum size=0.5cm] (E\i) at (3, 0.5*\i) {};
    }
    \node[draw, circle, minimum size=0.5cm] (D1) at (2, 2) {};
    \node[draw, circle, minimum size=0.5cm] (D2) at (2, 3) {};
    
    \foreach \i in {1,2,3,4,5,6,7,8,9,10} {
        \draw[dashed] (C\i) -- (D1);
        \draw[dashed] (C\i) -- (D2);
        \draw[dashed] (D1) -- (E\i);
        \draw[dashed] (D2) -- (E\i);
    }
    \node[rotate=90] at (0.5, 2.5) {Three hidden layers in pruned DNN};
\end{scope}
\end{tikzpicture}
\caption{Two hidden layers from the original, in the right figure, have 10 nodes (total 100 parameters). Layers are transformed into three hidden layers, in the left figure, in pruned DNN. The first layer has 10 nodes, the second layer has 2 nodes (keeping only two singular values in the SVD), and the third layer has 10 nodes, resulting in a total of 20 parameters.}
\label{fig:dnn_split}
\end{figure}

\end{ex}

\begin{ex}
\label{threshold_example}
     We used the above approach for a DNN trained on MNIST. In this example, the DNN has two layers, the first with a $784 \times 1000$ matrix $W_1$ and the second with a $1000 \times 10$ matrix $W_2$. The activation function was ReLU. We trained the DNN for $10$ epocs and achieved a $98$\% accuracy on the test set.

 We perform an SVD on $W_1$, in this case $\Sigma$ is a $784 \times 1000$ matrix. Even if we only keep the biggest $60$ $\sigma_i$ of $W_1$ and transform the first layer into two layers $W_{1,1}$ and $W_{2,1}$ the accuracy is still $98$\%. $W_1$ had $784,000$ parameters, while $W_{1,1}$ and $W_{2,1}$ have $784(60)+1,000(60)=107,040$ parameters (not including the bias vector parameters). This is a reduction of over $85\%$. In Fig. \ref{pvalue} we show how the accuracy of the DNN depends on the number of singular values that we keep. The red line corresponds to the threshold given by the MP distribution (via $\lambda_+$) for how many of the large singular values should be kept. As the figure shows, this threshold is highly accurate. This example also numerically confirms  Theorem \ref{the_main_result_RMT} and shows that the threshold given in the theorems (for which singular values to keep) is highly useful and accurate. 
       
       \begin{figure}[h!]
	\centering		\includegraphics[width=.5\textwidth]{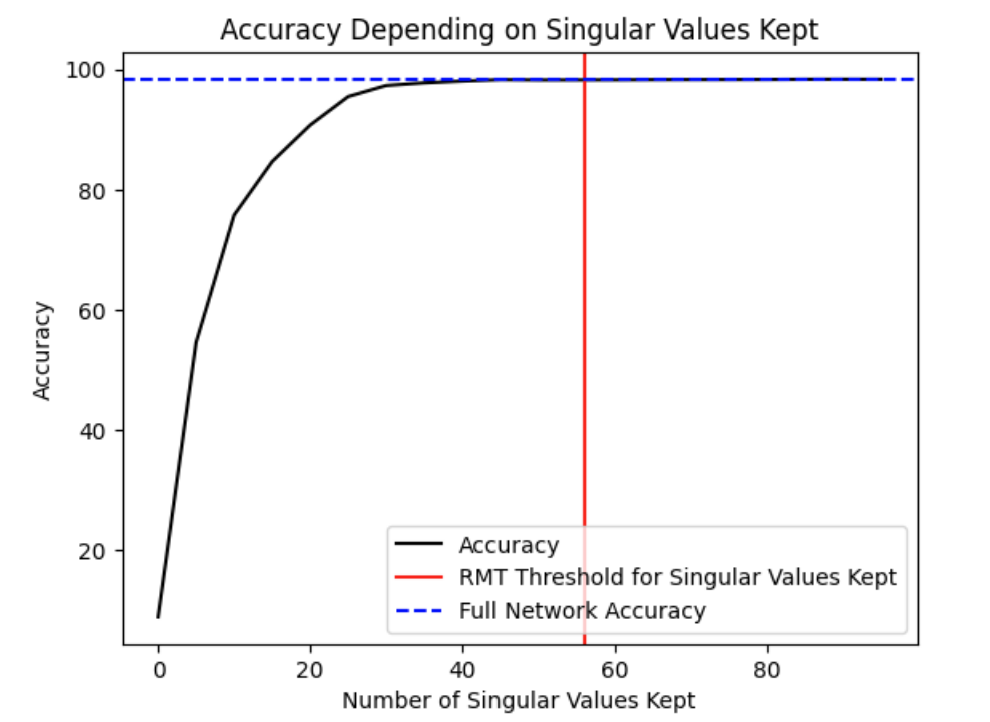}
		\caption{Number of eigenvalues kept is shown on the $x$-axis while the accuracy is shown on the $y$-axis.}
		\label{pvalue}
		\end{figure}

\end{ex}

\subsection{MP fit Criteria: Checking if the ESD of $X$ fits a MP distribution}
\label{Conformance_Assessment}
This subsection details a procedure to evaluate whether the ESD of a square matrix \(X\) is possibly drawn from a specific MP distribution (with a possibility of spiked eigenvalues). The initial step of this procedure relies on the BEMA method to identify the most fitting MP distribution. This optimal fitting distribution provides a theoretical cumulative distribution function (CDF), while the empirical cumulative spectral distribution related to \(X\) can be computed. A comparison of these two distributions allows us to dismiss the hypothesis that \(X\) follows the suggested MP distribution if the difference between the distributions is substantial. Let's formalize these concepts, starting with the concept of an empirical cumulative spectral distribution.

\begin{defn} \label{ECS}
    Assume \(G\) is an \(N \times M\) matrix and its ESD \(\mu_{G_M}\) is defined as in Definition \ref{ESD_Definition}. The empirical cumulative spectral distribution of \(G\), symbolized as \(F_G: \R \to \R\), is defined as: 
    \begin{equation}
        F_G(a) = \mu_{G_m}((-\infty, a])
    \end{equation}
\end{defn}

Interestingly, the cumulative distribution functions for the MP distribution are known and can be expressed in a closed form. With these equations, we can comprehensively describe our procedure. We set a tuning parameter \(\gamma \in (0,1)\) corresponding to the sensitivity of our test.
\begin{algorithm}
\caption{Assessing Conformance to the MP Distribution}
\begin{algorithmic}[1]
\State Accept \(X = \frac{1}{N} W^T W\) as input, where \(W\) is an \(N \times M\) matrix.
\State Calculate the spectrum of \(X = \{\sigma_1, \dots, \sigma_M\}\).
\State Compute the empirical cumulative spectral distribution of \(X\), denoted \(F_X\).
\State Execute the BEMA method with parameters \(\alpha\) and \(\beta\) to determine \(\hat\sigma^2\), the anticipated variance of each coordinate of $W$.
\State Calculate \(0 \leq i_{\text{low}} < i_{\text{high}} \leq M\) such that \(i_{\text{low}}\) is the smallest integer with \(\frac{i_{\text{low}}}{M} \geq \alpha\) and \(i_{\text{high}}\) is the largest integer with \(\frac{i_{\text{high}}}{M} \leq 1 - \alpha\).
\State Define \(F_X'\) as the theoretical cumulative distribution function for the MP distribution with parameters \(\hat \sigma^2\) and \(\lambda = N/M\).
\State Evaluate \(s = \max_{i \in [i_{\text{low}}, i_{\text{high}}]} \left |F_X(i) - F_X'(i) \right |\).

\If{\(s > \gamma\)}
    \State Dismiss the hypothesis that \(X\) follows the proposed distribution.
\Else
    \State Do not reject this hypothesis.
\EndIf
\end{algorithmic}
\end{algorithm}

This procedure computes the maximum difference between the expected and empirical cumulative distribution functions by sampling at each point in the empirical distribution. Since this is intended for the unique case of testing for spiked MP distributions, we can utilize this information to enhance our test over simply calculating the \(L^\infty\) difference between the expected and empirical distributions.

This enhancement is reflected in the step which calculates \(i_{\text{low}}\) and \(i_{\text{high}}\). As BEMA only uses data in the quantile between \((\alpha, 1 - \alpha)\) to find the best fit, it is logical to only examine for fit within the same range. In this context, we would expect a spiked MP distribution to be poorly approximated by its generative MP distribution around the highest eigenvalues (i.e., the spiked values), and hence it makes sense to only test the bulk values for goodness of fit.

\section{Appendix E: Some of the proofs and numerics}

\subsection{Proof of Lemma \ref{main_result_new}}
\label{proof_main_result_1_new}
\begin{proof}
The classification confidence before pruning is:
\[ \delta X(s,\alpha_{W_b}) = X_{i(s)}(s,\alpha_{W_b}) - \max_{j \neq i(s)} X_j(s,\alpha_{W_b}) \]

After pruning, it is:
\[ \delta X(s,\alpha_{W_b'}) = X_{i(s)}(s,\alpha_{W_b'}) - \max_{j \neq i(s)} X_j(s,\alpha_{W_b'}) \]

For simplicity, we will start by proving the theorem for the case of a DNN with only one layer matrix $W$ and a bias vector $\beta$. Thus, we take $ X(s,\alpha_{W_b})= X(s,\alpha_{W})$ and  $X(s,\alpha_{W'_b})= X(s,\alpha_{W'})$. 

Then, the change in classification confidence due to pruning is:
\[ \Delta(\delta X) = |\delta X(s,\alpha_W) - \delta X(s,\alpha_{W'})| \]

For a particular component \( i \), the change \( \Delta X_i \) due to pruning is given by:
\[ \Delta X_i = |X_i(s,\alpha_W) - X_i(s,\alpha_{W'})| \]

We have that $ X(s,\alpha_W)=\lambda \circ (W+\beta)s$ and $ X(s,\alpha_{W'})=\lambda \circ (W' +\beta)s$.

Given that $\lambda$ is the absolute value activation function, for any scalar values $x$ and $y$, we have:

\begin{equation}
|\lambda(x) - \lambda(y)| \leq |x - y|.
\end{equation}
Thus $\| X(s,\alpha_W)- X(s,\alpha_{W'})\| \leq \| (W-W')s \| \leq \sqrt{\lambda_+}\|s\|_2$. If the change in the norm of the entire output vector due to pruning is at most \( \sqrt{\lambda_+} \|s\|_2 \), then the maximum change in any individual component must also be bounded by that amount. That is:
\[ \Delta X_{\text{max}} \leq \sqrt{\lambda_+} \|s\|_2. \]

Furthermore, 

\begin{equation}
\label{change_together_1}
    \Delta X_{\text{max}}+\Delta X_{\text{max-1}} \leq \sqrt{2\lambda_+} \|s\|_2,
\end{equation}
with $\Delta X_{\text{max-1}}$ the change in the component of $X$ which has the second to biggest change and  $\Delta X_{\text{max}}$ the change component of $X$ which had the biggest change.   

Now assume $\max_{j \neq i(s)} X_j(s,\alpha_{W'}) \leq \max_{j \neq i(s)} X_j(s,\alpha_W) $, then given that $|X_{i(s)}(s,\alpha_W) - X_{i(s)}(s,\alpha_{W'})| \leq \sqrt{\lambda_+}\|s\|_2$, we must have $ \delta X(s,\alpha_{W'})= X_{i(s)}(s,\alpha_{W'}) - \max_{j \neq i(s)} X_j(s,\alpha_{W'}) \leq \sqrt{\lambda_+}\|s\|_2$.

Next, assume $\max_{j \neq i(s)} X_j(s,\alpha_{W'}) \geq \max_{j \neq i(s)} X_j(s,\alpha_W) $. Let's expand the change in \( \delta X \) due to pruning:
\begin{align*}
\Delta(\delta X) &= |(X_{i(s)}(s,\alpha_W) - \max_{j \neq i(s)} X_j(s,\alpha_W)) - (X_{i(s)}(s,\alpha_{W'}) - \max_{j \neq i(s)} X_j(s,\alpha_{W'}))| \\
&\leq |X_{i(s)}(s,\alpha_W) - X_{i(s)}(s,\alpha_{W'})| + |\max_{j \neq i(s)} X_j(s,\alpha_W) - \max_{j \neq i(s)} X_j(s,\alpha_{W'})|.
\end{align*}

Take $k^*$ to be an integer such that $\max_{j \neq i(s)} X_j(s,\alpha_{W'})= X_{k^*}(s,\alpha_{W'})$, we have, given that $\max_{j \neq i(s)} X_j(s,\alpha_{W'}) \geq \max_{j \neq i(s)} X_j(s,\alpha_W) $, that:

\begin{equation}
\begin{split}
    |X_{i(s)}(s,\alpha_W) - X_{i(s)}(s,\alpha_{W'})| + |\max_{j \neq i(s)} X_j(s,\alpha_W) - \max_{j \neq i(s)} X_j(s,\alpha_{W'})| \\
    \leq |X_{i(s)}(s,\alpha_W) - X_{i(s)}(s,\alpha_{W'})| +| X_{k^*}(s,\alpha_{W'}) -  X_{k^*}(s,\alpha_{W})|. 
\end{split}
\end{equation}

Consider the worst-case scenario where the change is maximally concentrated in the components $i(s)$ and $k^*$. Thus, given \eqref{change_together_1} we have:
\begin{equation}
     \Delta(\delta X) \leq     \Delta X_{\text{max}}+\Delta X_{\text{max-1}}  \leq \sqrt{2\lambda_+} \|s\|_2.
\end{equation}

Next, we consider the case where the DNN has multiple layers given by the matrices $W_1, \cdots, W_L$, and bias vectors $\beta_1,  \cdots \beta_L$ and we prune the last matrix layer of the DNN, i.e. we prune layer matrix $W_L$ to obtain $W'_L$. In this case, we can reduce this problem to that of a DNN with a single layer matrix $W_L$ and a single bias vector $\beta_L$ and an input vector $z=\lambda \circ W_{L-1} \circ \dots \circ \lambda \circ W_{1}s$. Then, by the above argument, we have:

\begin{equation}
   \Delta(\delta X)\leq  \sqrt{2\lambda_+} \|\lambda \circ (W_{L-1}+\beta_{L-1}) \circ \dots \circ \lambda \circ (W_{1}+\beta_1)s\|_2. 
\end{equation}

Finally, we assume that we prune a layer $W_b$ to obtain $W'_{b}$. By a similar argument to what we have above, we see that the max change of a component of the output vector $X(s,\alpha_{W_b})$ after pruning is: 

\[ \Delta X_{\text{max}} \leq \sqrt{\lambda_+} \|\lambda \circ (W_{b-1}+\beta_{b-1}) \circ \dots \circ \lambda \circ (W_{1}+\beta_1)s\|_2\sigma_{\max}(W_{b+1}) \dots \sigma_{\max}(W_{L}),\]
given that $\|\lambda\circ (Az+\beta) - \lambda\circ (Az'+\beta)\|_2 \leq \sigma_{\max}(A)\|z'-z\|_2$.  

Again, we obtain, 

\begin{equation}
    \Delta X_{\text{max}}+\Delta X_{\text{max-1}}\leq \sqrt{2\lambda_+} \|\lambda \circ (W_{b-1}+\beta_{b-1}) \circ \dots \circ \lambda \circ (W_{1}+\beta_1)s\|_2\sigma_{\max}(W_{b+1}) \dots \sigma_{\max}(W_{L}). 
\end{equation}
Thus, this completes the proof of the lemma.

\end{proof}

\subsection{Effectiveness of MP-based Pruning for Different Initialization Methods}
\label{Other_initilization}

In this subsection, we investigated the performance of various initialization methods, including He (see \cite{he2015delving}) and Xavier (see \cite{glorot2010understanding}) initialization. It is important to note that both the He and Xavier initializations align with the principles of the MP theorem given in Theorem \ref{RMT_MP_theorem}. That is, the weight layer components are i.i.ds with mean zero and bounded variance. In practice, for the weight layer matrices $W$ initialized based on those distributions, we have that the ESD of $W^TW$ fits the MP distribution well (with an error, see Subsection \ref{Conformance_Assessment}, of $\sim .001$ for both). Thus, the Pruning Theorem \ref{the_main_result_RMT} would hold for both of these initializations.

\begin{figure}[h]
    \centering
    \begin{subfigure}[b]{0.8\textwidth}
        \centering
        \includegraphics[width=\textwidth]{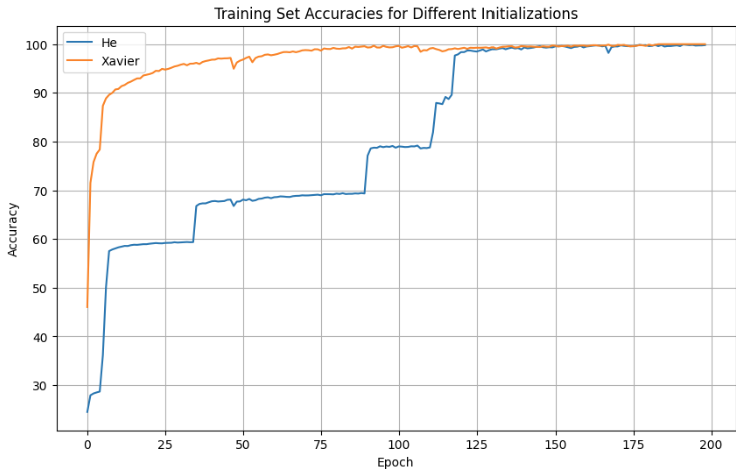}
        \caption{Training Set Accuracy vs Epoch for He and Xavier Initializations}
        \label{fig:train_init_comparison}
    \end{subfigure}
    \hfill
    \begin{subfigure}[b]{0.8\textwidth}
        \centering
        \includegraphics[width=\textwidth]{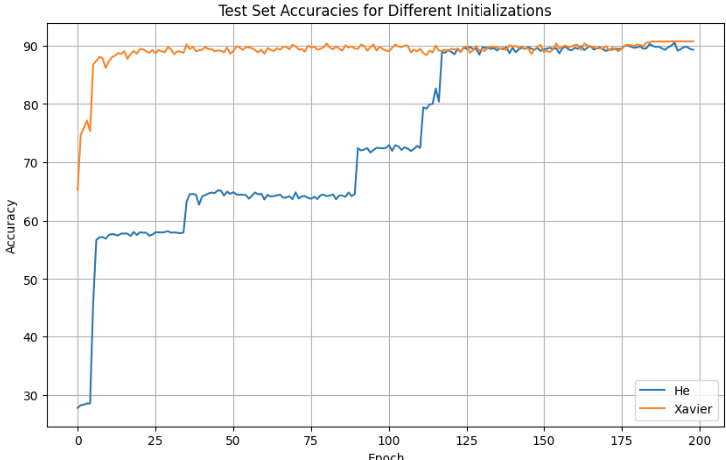}
        \caption{Test Set Accuracy vs Epoch for He and Xavier Initializations}
        \label{fig:test_init_comparison}
    \end{subfigure}
    \caption{Accuracy vs Epoch for He and Xavier Initializations}
    \label{fig:init_comparison}
\end{figure}

The DNNs were fully connected, and their architecture was given by $ [784, 3000, 3000, 3000, 3000, 500, 10]$. When pruning with MP-based pruning, both initializations achieved test accuracies above 90\%, which was comparable to the performance of networks initialized with a normal distribution (which was $90.74\%$), see Fig. \ref{fig:init_comparison}. Without MP-based pruning, the DNNs achieved accuracy on the test set of $\sim 89\%$.   All the DNNs were trained using a combination of $L2$ and $L1$ regularization (see \eqref{loss_noise_det}), which is why the DNNs achieved an accuracy above $90\%$. The using of both $L1$ and $L2$ regularization together with MP-pruning is a critical factor in achieving these high accuracies, consistently above 90\%. The interplay between MP-based pruning and regularization, which contributes to this performance, will be discussed in detail in another paper. We used the hyperparameters $\mu_1=.0000005$ and $\mu_2= .0000001$. The other hyperparameters are the same as those given in the numerical simulations from Subsection \ref{full_Fash_MNIST}.    

\begin{equation}
\label{loss_noise_det}
L(\alpha(a))=-\frac{1}{|T|}\sum_{s\in T}\log\left(\phi_{i(s)}(s,\alpha(a))\right) + \mu_1 \sum_{i=1}^L \|W_i(a)\|_1,
+ \mu_2 \sum_{i=1}^L \|W_i(a)\|_F^2
\end{equation}

Additionally, we conducted a similar simulation where the initial weights were drawn from a normal distribution $N(0,\frac{1}{N})$, but $90\%$ of the parameters were randomly initially sparsified afterward and set to zero (though we still used them during training). The resulting DNN's test accuracy plateaued at $ \sim 80\%$ and failed to improve beyond this point, even when using MP-based pruning. \textbf{It is important to note that the MP theorem does not hold for weight layers initialized in such a manner.} Similar results were found when we used He and Xavier's initializations but initially sparsified them so that $90\%$ of the parameters started out as zero (but were used during training).

Finally, we also ran the numerical simulations for the above DNN architecture and hyperparameters, but setting $\mu_1=0$ (that is only using $L2$ regularization, which is what was also done in Subsection \ref{full_Fash_MNIST}). For the He and Xavier initialization, the DNNs, with MP-based pruning, obtained an accuracy of $\sim 90\%$, comparable to the simulations found in Subsection \ref{full_Fash_MNIST}, while for the sparse initialization, their accuracy never appreciated higher than $80\%$ even when using MP-based pruning. Without MP-based pruning, the He and Xavier initialization accuracies were $\sim 88\%$.

\subsection{A Regression Problem: MP-based pruning in regression}
\label{MP_reg}
\begin{figure}[h!]
    \centering
    \includegraphics[width=.5\textwidth]{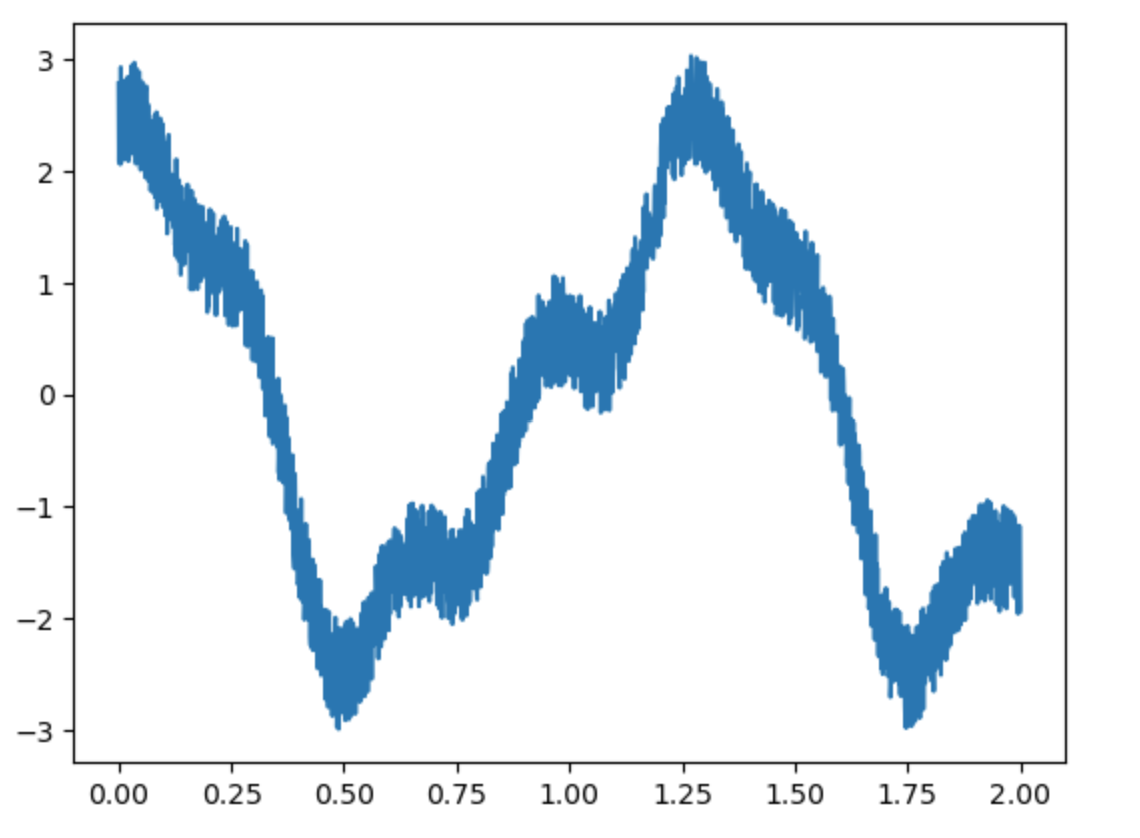}
    \caption{Illustration of the regression problem.}
    \label{fig:RMT_regression_Problem}
\end{figure}

We consider the task of finding a DNN that approximates a function best fitting the given data in terms of Mean Squared Error (MSE); see Fig. \ref{fig:RMT_regression_Problem}.

\paragraph*{Training Data}
The training data consists of 2000 equally spaced points between 0 and 2:
\[
x_{\text{train}} = \left\{ x_i \mid x_i = \frac{2i}{1999}, \, i = 0, 1, 2, \ldots, 1999 \right\}.
\]

Noise is added to the training labels. The noise is uniformly distributed between -1 and 1 with a scaling factor of $\sigma = 0.5$:
\[
\text{noise}_{\text{train}} = \left\{ \epsilon_i \mid \epsilon_i \sim \mathcal{U}(-1, 1) \right\}.
\]

The training labels are generated using the following function:
\[
y_{\text{train}} = 0.5 \cos(20x_{\text{train}}) + 2 \cos(5x_{\text{train}}) + 0.5 \sin(10x_{\text{train}}) + \sigma \cdot \text{noise}_{\text{train}}.
\]

\paragraph*{Testing Data}
The testing data consists of 500 equally spaced points between 0 and 2:
\[
x_{\text{test}} = \left\{ x_i \mid x_i = \frac{2i}{499}, \, i = 0, 1, 2, \ldots, 499 \right\}.
\]

The testing labels are generated using the following function without noise:
\[
y_{\text{test}} = 0.5 \cos(20x_{\text{test}}) + 2 \cos(5x_{\text{test}}) + 0.5 \sin(10x_{\text{test}}).
\]

\paragraph{Defining Output and Target}
The DNN used for this 2D regression has a simple structure:
\begin{itemize}
    \item The first and last layers each consist of a single neuron.
    \item The network outputs a single value and is trained using MSE together with $L2$ regularization as the loss function.
\end{itemize}

\paragraph{DNN Topology}
We use a fully connected DNN for the regression problem given in Fig. \ref{fig:RMT_regression_Problem}, that is, to find the curve that best fits the data. The fully connected DNN used in this regression problem has the following topology: $[1, 1000, 1000, 1000, 1000, 1]$, which represents the number of nodes in each layer. We don't use an activation function after the final layer of the DNN. 

\paragraph{Training Process}
Both an unpruned DNN and a DNN using RMT pruning were trained to minimize MSE loss. The hyperparameters for training can be found in Table \ref{Reg_hyper}. We decreased the lr by $.997$ every epoch.

\paragraph{Comparison of Unpruned vs Pruned DNN}
\begin{figure}[h!]
    \centering
    \includegraphics[width=0.49\textwidth]{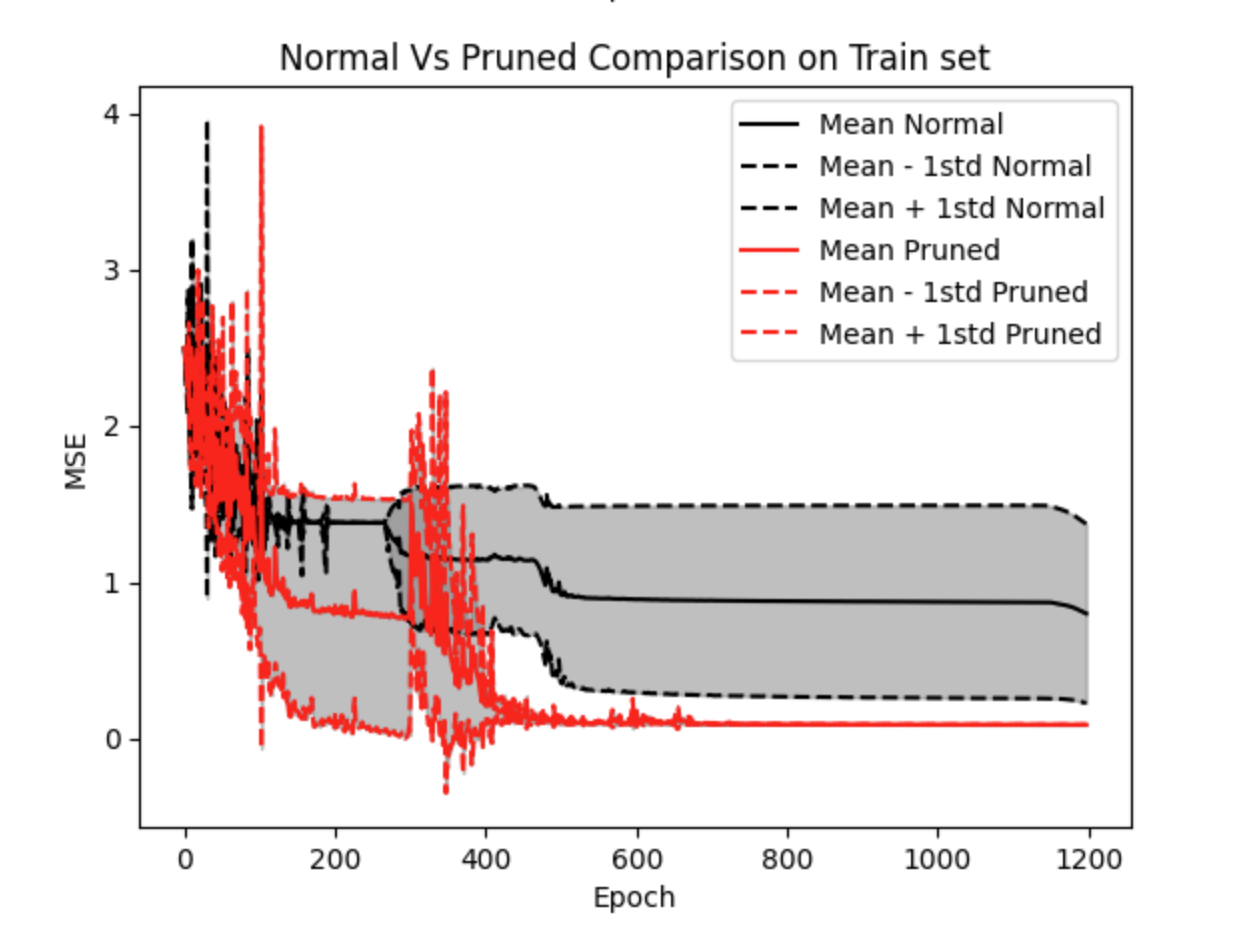}
    \caption{Training loss comparison between normal and pruned DNN.}
    \label{fig:training_loss}
\end{figure}

\begin{figure}[h!]
    \centering
    \includegraphics[width=0.5\textwidth]{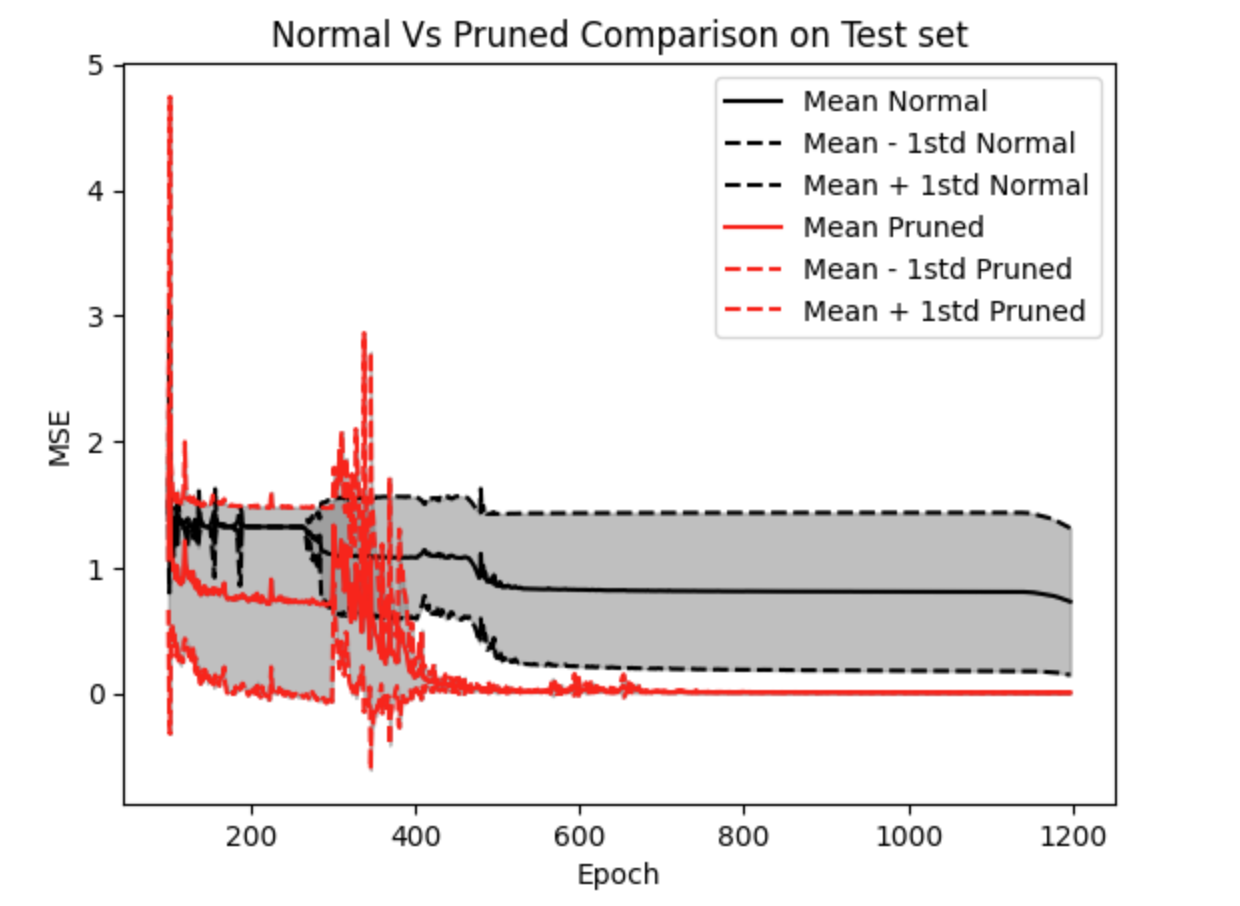}
    \caption{Test loss comparison between normal and pruned DNN.}
    \label{fig:test_loss}
\end{figure}

The pruned DNN shows a significantly smaller loss on both training and test sets, with a $50\%$ reduction in the number of parameters, see Figs. \ref{fig:training_loss} and \ref{fig:test_loss}. It is important to note that a smaller DNN might perform better on this task when MP-based pruning is not employed. However, for larger fully connected DNNs it is clear that MP-based pruning helps the DNN achieve much lower loss for this task.   

\begin{table}[h]
    \centering
    \begin{tabular}{ll}
        \hline
        \textbf{Hyperparameter} & \textbf{Value} \\
        \hline
        Number of epochs & 1198 \\
        Number of seeds & 5 \\
        Learning rate (lr) & 0.025 \\
        Momentum & 0.95 \\
        Batch size & 128 \\
        $L2$ regularization on loss & 0.001 \\
        \hline
    \end{tabular}
    \caption{Training hyperparameters for the DNN.}
    \label{Reg_hyper}
\end{table}

\subsection{Numerical example used to calculate $\delta X$.}
\label{Example_delta_X}

\begin{figure}[h!]
\centering
\includegraphics[width=.5\textwidth]
{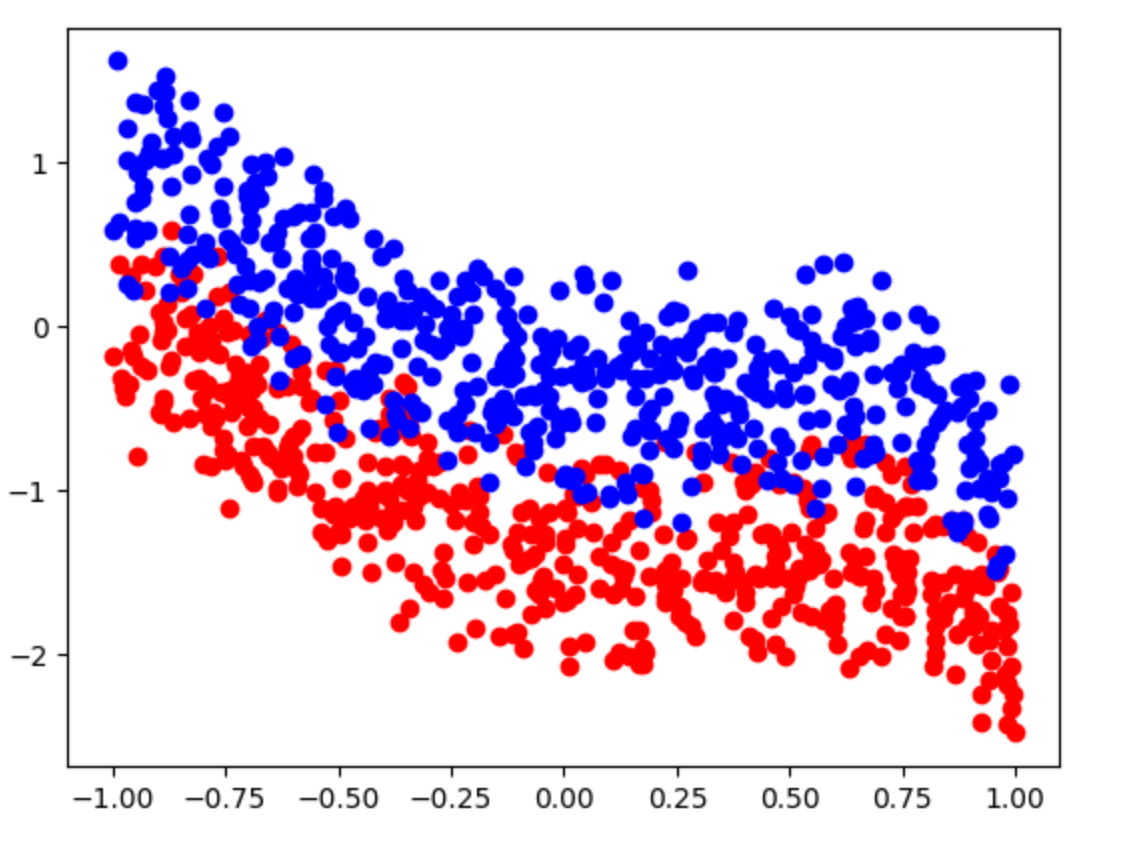}
\caption{Illustration of a decision boundary for a binary classification challenge created using a random polynomial function, supplemented with noise. The distinct blue and red points signify the two classes.}
\label{fig:toy_example_1}
\end{figure}

In this demonstration, we have a set of two-dimensional data points, and our objective is to identify the boundary that divides them into two categories. The dataset originates from a randomly constructed polynomial function of a designated degree. We then sample data points uniformly across a spectrum of x-values. For every x-value, the polynomial function gives a y-value. These points are then slightly offset either upwards or downwards, forming two distinguishable point clusters labeled as red and blue. Points positioned above the polynomial curve get a blue label, and those below are tagged red. We also add Gaussian noise to slightly modify the y-values, causing a few red data points to appear below the boundary and some blue ones above. This scenario is depicted in Fig. \ref{fig:toy_example_1}.

The goal is to harness a DNN to capture the decision boundary demarcating the two clusters. This DNN is designed to process a two-dimensional data point, producing a binary output indicating whether the point is red or blue. Given that the dataset is artificially curated, the actual decision boundary is known, allowing us to gauge the efficacy of our DNN.

For this task, our neural network model had one hidden layer with 500 neurons. We also used the ReLU activation function. Training is executed using the cross-entropy loss complemented by the SGD optimizer and momentum.

\subsection{Hyperparameters for Subsection \ref{fully_connected_MNIST}}
\label{hyperparam_1}

\begin{itemize}
    
\item Number of epochs: 40
\item Number of seeds: 10
\item Learning rate (lr): 0.02
\item Momentum: 0.9 
\item Batch size: 128 
\item $L2$ regularization on loss: $0.0005$. (The regularization term is applied for both DNNs, the normal and pruned versions). 

\end{itemize}

See \cite{lewkowycz2020large}, \cite{sutskever2013importance}, \cite{sutskever2013importance} and \cite{hinton2012improving} for information on learning rate, momentum, batch size, and regularization, respectively.

\subsection{Hyperparameters for Subsection \ref{full_Fash_MNIST}}
\label{hyperparam_2}

\begin{itemize}
\item Number of epochs: 70
\item Number of seeds: 10
\item Learning rate (lr): 0.02.
\item Momentum: 0.9
\item Batch size: 128
\item $L2$ regularization on loss: $0.0005$. 

\end{itemize}

\subsection{Hyperparamters for Subsection \ref{MNIST_Fashion_Con_Red}}
\label{hyperparam_3}

\textbf{Training hyperparameters:}

\begin{itemize}
\item Number of epochs: Depends on the example.
\item Number of seeds: 5
\item Learning rate (lr): 0.02.
\item Momentum: 0.9
\item Batch size: 128
\item Split frequency: Depends on the example. 
\item $L^2$ regularization on loss:  .001. 
\item goodness of fit (GoF): changes for every simulation
\end{itemize}

\subsection{CNN Architecture Description}
\label{CNN_arc}

The CNN model in this study consists of multiple convolutional and fully connected layers, incorporating batch normalization and dropout for regularization. The architecture is designed as follows:

\paragraph{Convolutional Layers:}
\begin{itemize}
    \item The network contains several convolutional layers with kernels of size 3. Each convolutional layer is followed by batch normalization.
    \item Activation functions (ReLU) are applied conditionally based on predefined configurations.
    \item Max-pooling layers are introduced after every second convolutional layer, except the first one. Specifically, max-pooling with a pool size of 2 is used to reduce the spatial dimensions of the feature maps.
    \item Dropout is applied after each convolutional layer to prevent overfitting, with a dropout rate of 0.35.
\end{itemize}

\paragraph{Fully Connected Layers:}
\begin{itemize}
    \item After the convolutional layers, the output is flattened and passed through a series of fully connected layers.
    \item Each fully connected layer is followed by batch normalization and, conditionally, by a ReLU activation function based on the configuration.
    \item Dropout is also applied to the output of the first fully connected layer to further prevent overfitting.
\end{itemize}

\paragraph{Output Layer:}
\begin{itemize}
    \item The final layer applies a log-softmax function to produce the output probabilities for classification.
\end{itemize}

\subsubsection{Pooling and Regularization Details}

\paragraph{Pooling Layers:}
Max-pooling layers are strategically placed to downsample the feature maps, specifically after every second convolutional layer. This pooling strategy helps in reducing the computational complexity and in extracting invariant features.

\paragraph{Batch Normalization:}
Batch normalization is applied after each convolutional and fully connected layer to stabilize and accelerate the training process by normalizing the input to each layer.

\paragraph{Dropout:}
Dropout is utilized after each convolutional layer and the first fully connected layer with rates of 0.35 and variable values, respectively, to reduce overfitting by randomly setting a fraction of activations to zero during training.

\subsection{The hyperparamters for Subsection \ref{CIFAR-10}}
\label{hyperparam_4}
\textbf{Training hyperparameters:}

\begin{itemize}
\item Number of epochs: 300.
\item Number of seeds: 5
\item Learning rate (lr): 0.001.
\item Momentum: 0.9
\item Batch size: 128
\item Split frequency (every how many epochs we split the pruned DNN and remove small singular values): 40. 
\item $L2$ regularization on loss:  .001. 
\item goodness of fit (GoF): 0.08 for fully connected layers, 0.06 for convolutional layers.
\end{itemize}

\end{document}